\documentclass{article}

\usepackage{csquotes}
\usepackage{microtype}
\usepackage{graphicx}
\usepackage{placeins}
\usepackage{booktabs} 

\usepackage{hyperref}
\usepackage{url}

\usepackage[accepted]{Styles/icml2025}

\usepackage{amsmath}
\usepackage{amssymb}
\usepackage{mathtools}
\usepackage{amsthm}
\usepackage{multirow}

\usepackage{subcaption}

\usepackage[capitalize,noabbrev]{cleveref}

\newcommand{\mistral}{Mixtral-8x22B}
\newcommand{\llama}{Llama-3.1-70B}
\newcommand{\aya}{Aya-23-35B}
\newcommand{\gemma}{Gemma-2-27b}

\usepackage[encapsulated]{CJK}
\newcommand{\md}[1]{\begin{CJK}{UTF8}{gbsn}#1\end{CJK}}

\usepackage{array}
\usepackage{longtable}

\icmltitlerunning{Do Multilingual LLMs Think In English?}

\begin{document}

\twocolumn[
\icmltitle{Do Multilingual LLMs Think In English?}

\icmlsetsymbol{equal}{*}

\begin{icmlauthorlist}
\icmlauthor{Lisa Schut}{oat}
\icmlauthor{Yarin Gal}{oat}
\icmlauthor{Sebastian Farquhar}{gdm}
\end{icmlauthorlist}

\icmlaffiliation{oat}{OATML, Department of Computer Science, University of Oxford.}
\icmlaffiliation{gdm}{Google DeepMind}

\icmlcorrespondingauthor{Lisa Schut}{schut@robots.ox.ac.uk}

\icmlkeywords{Machine Learning, ICML}

\vskip 0.3in
]

\printAffiliationsAndNotice{} 

\begin{abstract}
Large language models (LLMs) have multilingual capabilities and can solve tasks across various languages. 
However, we show that current LLMs make key decisions in a representation space closest to English, regardless of their input and output languages. 
Exploring the internal representations with a logit lens for sentences in French, German, Dutch, and Mandarin, we show that the LLM first emits representations close to English for semantically-loaded words before translating them into the target language. 
We further show that activation steering in these LLMs is more effective when the steering vectors are computed in English rather than in the language of the inputs and outputs. 
This suggests that multilingual LLMs perform key reasoning steps in a representation that is heavily shaped by English in a way that is not transparent to system users.
\end{abstract}

\section{Introduction}

\begin{figure}[t]    
    \begin{minipage}{0.5\textwidth}
    \centering
    \includegraphics[width=\textwidth]{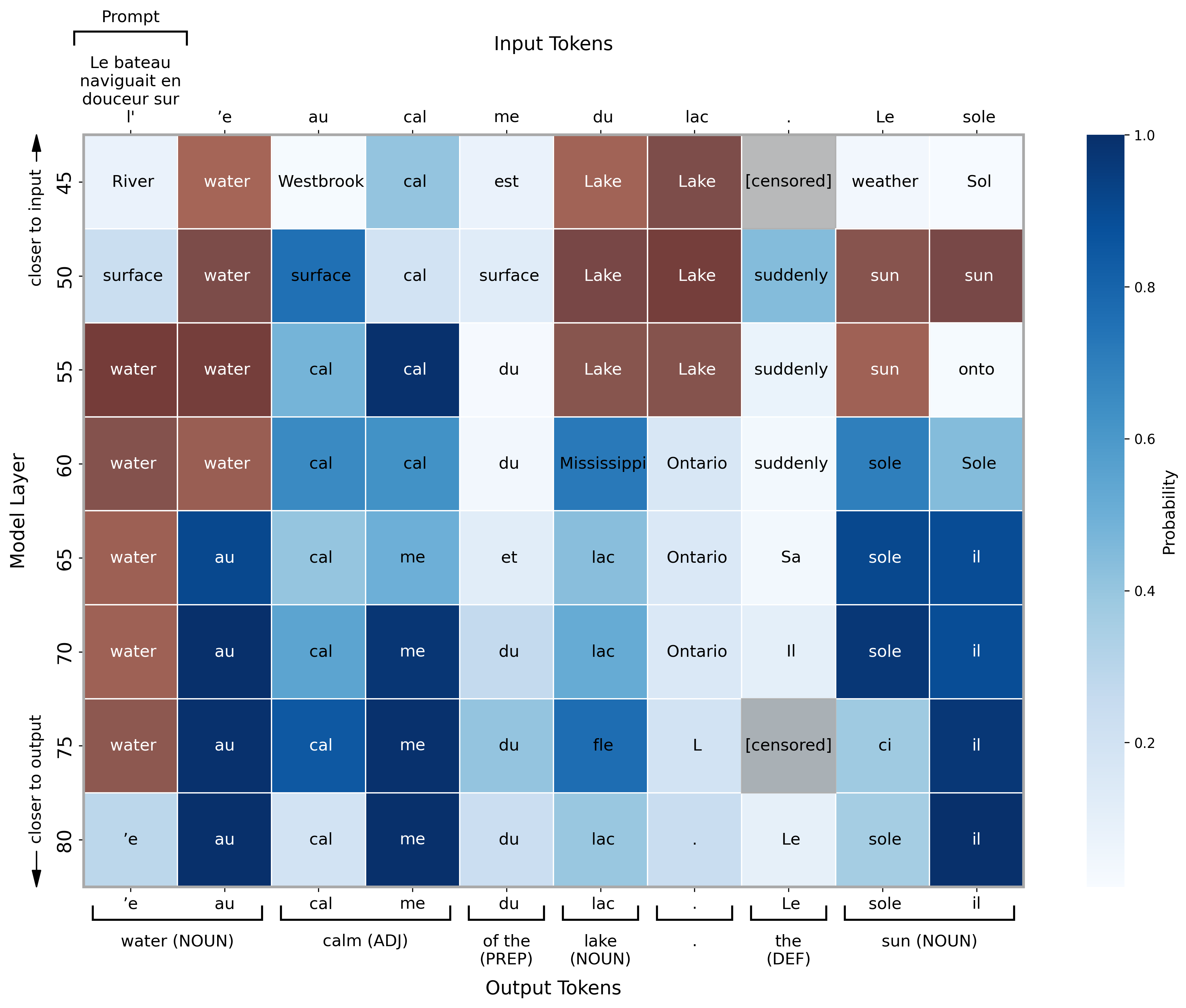} 
    \end{minipage}
    \caption{Logit lens applied to \llama's latent space, when prompted with \textit{Le bateau naviguait en douceur sur l'}. Each row depicts the decoded latent representations for one layer and each column corresponds to the generated token. Dark red boxes highlight words selected in English. The nouns `eau', `lac', and `soleil' are selected in English, whereas other parts of speech are not.}
    \label{fig:logit_lens}
    
\end{figure}

Large Language Models (LLMs) are predominantly trained on English data, yet are deployed across various languages, including some that are rarely seen during training. 
This raises an important question: \textit{how} do LLMs operate across different languages?

LLMs are hypothesized to operate in an abstract concept space  \citep{olahfeaturesllm, nanda2023progress, wendler2024llamasworkenglishlatent, dumas2024llamas}.
From the multilingual perspective, one main question is whether the concept space is language-specific or language-agnostic. 
We consider three different hypotheses: 
\begin{enumerate}
    \item LLMs `operate' in a space that is English-centric (or centered on the main pretraining language)
    \item LLMs `operate' in a language-agnostic space
    \item LLMs `operate' in a language-specific space, which is determined by the input language.
\end{enumerate}

We present evidence that the first hypothesis is true: LLMs reason in an English-centric way. Our work studies open-ended multi-token language generation, contrasting with prior work \citep{wendler2024llamasworkenglishlatent} which found evidence for the second hypothesis in the single token context. 

\begin{table*}[t]
\caption{Summary of pre-training dataset languages of the open-source large language models studied in this work.}
\label{model-summary}
\vskip -0.2in
\begin{center}
\begin{small}
\begin{sc}
\begin{tabular}{ p{6cm} p{2.3cm} ccccc}
\toprule
Model & Nr. Languages &\multicolumn{5}{c}{Trained on Language?} \\ 
& Trained On  & English & French & German  &Dutch & Mandarin \\ 
\midrule
Aya-23-35B {\citep{aryabumi2024aya}}  & 23 & $\surd$ & $\surd$ & $\surd$ & $\surd$  & $\surd$ \\
Llama-3.1-70B  \citep{dubey2024llama3herdmodels}  & 8 & $\surd$ & $\surd$ & $\surd$ & $\times$  & $\times$ \\
Mixtral-8x22B-v0.1 \citep{jiang2024mixtralexperts} & 5 & $\surd$ & $\surd$ & $\surd$ & $\times$  & $\times$ \\
Gemma-2-27b \citep{aryabumi2024aya}  & 1 & $\surd$ & $\times$ & $\times$ & $\times$  & $\times$ \\
\bottomrule
\end{tabular}
\end{sc}
\end{small}
\end{center}
\vskip -0.1in
\end{table*}

We study three aspects of language generation. 
First, we study how representations progress within the model, showing that for lexical words, English-focused representations often appear first before being transformed into the target language.
However, non-lexical words do not route through the English representation space. 
Second, we show that steering the representations is more effective using vectors constructed in English than in the target language.
Third, we show that the latent representation structure is consistent with the language and semantic context being represented separately. In more detail, we:

\paragraph{Decode the Representation Space}
LLMs make semantic decisions in English, even when prompted in a non-English language. 
Figure \ref{fig:logit_lens} shows the logit lens to \llama \ as it generates the French text \textit{Le bateau naviguait en douceur sur l'\textbf{eau au calme du lac. Le soleil ...}}, with the bold text representing Llama's output. The English translation is \textit{The boat sailed smoothly on the \textbf{calm water in of the lake. The sun ...}}. Lexical words like ``water," ``lake," and ``sun" are selected in English, whereas grammatical elements such as ``du" and ``le" are not. We find that this trend holds more generally for other models (Section \ref{sec:logit_lens_latent}), with Aya being the least English-centric and Gemma the most English-centric.

\paragraph{Manipulating the representations}
Non-English sentences generated can be steered more effectively using English-derived steering vectors than those derived from the target language. This surprising result provides further evidence that LLMs rely on an English-centric conceptual space for semantic reasoning. 
Further, we find that the steering vectors have a relatively high cosine similarity, however, they do encode a language-specific component. We can increase the similarity of steering vectors found in different languages by nudging them toward each other using a language steering vector.

\paragraph{Structure of the latent space}
Fact representations are shared between languages, allowing interpolation between a fact expressed in two languages while maintaining the correct answer—changing only the output language. 

We analyze four open source models  (\llama, \gemma, \aya \ and \mistral), which vary in architecture and language coverage.
In general, we find that LLMs make decisions in the representation space close to English, independent of the input or output language. 
This English-centric behavior of LLMs cause them to perform worse in other languages, whether in downstream tasks \citep{shafayat2024multifactassessingfactualitymultilingual, huang2023languagescreatedequalllms, bang2023multitaskmultilingualmultimodalevaluation, shi2022language}, or in fluency \cite{guo2024benchmarking}. Moreover, this impacts the fairness of these models -- which currently exhibit cultural biases \cite{shafayat2024multifactassessingfactualitymultilingual} -- and their robustness and reliability in diverse linguistic settings \cite{ marchisio2024understandingmitigatinglanguageconfusion, deng2024multilingualjailbreakchallengeslarge}.

\section{Background} \label{sec:background}

\subsection{Large Language Models}

Language models are trained to operate across different languages. 
Table \ref{model-summary} summarizes the four LLMs we study, which differ in the number of languages they were trained on. 
\aya \ supports the widest range of languages, while \gemma \ covers the fewest.

We evaluate these models across five languages, selected based on their varying levels of representation during training. English, the predominant training language, serves as a baseline. French and German represent high-resource, non-English languages, while Dutch and Chinese are lower-resource languages. Dutch is only a high-resource language in \aya \ and therefore provides an interesting comparison to German due to their linguistic similarity.
This analysis allows us to understand the performance disparities across languages with varying levels of representation in training.

\subsection{Methods}

Our goal is to understand whether LLMs have a universal representation space. 
To address this question, we use three mechanistic interpretability methods. 
The logit lens (Section \ref{sec:logit_lens}) allows us to examine the internal representations, while causal tracing provides insight into where facts are encoded in the model across different languages (Section \ref{sec:patching}). 
Finally, steering vectors let us intervene on the models' internal representations (Section \ref{sec:steering_vec}), which allows us to verify that the representations influence the output.

\begin{table*}[t]
\caption{LLM-Insight dataset examples: sentences and prompts for the word  animal.}
\label{data-example}
\vskip -0.25in
\begin{center}
\begin{small}
\begin{sc}
\begin{tabular}{ l p{8cm} p{6cm}  }
        \toprule
        {Language} & {Sentence Example} & {Prompt Example}   \\
        \midrule
        English & The zoo has a wide variety of animal species. & They adopted a    \\ 
        Dutch & De boerderij had elk type huisdier. & In de dierentuin zag ik een bijzonder  \\ 
        French & Le lion est un animal sauvage qui vit dans la savane. & Il a vu un \\ 
        Mandarin & \md{森林中生活着许多野生动物}& \md{每年都会有新的}     \\ 
        German & Der Zoo beherbergt viele faszinierende Tiere. & Sie liebt es Zeit mit ihrem    \\ 
        \bottomrule
    \end{tabular}
\end{sc}
\end{small}
\end{center}
\vskip -0.1in
\end{table*}

\subsubsection{Logit Lens} \label{sec:logit_lens}
The logit lens \citep{logitlens} decodes the internal representations of an LLM into tokens. LLMs take an input $x$ and output a probability distribution over the next token.
The logit lens decodes the intermediate representation $h_l(x)$ at layer $l$ into an output token, by applying the unembedding layer:
\begin{equation}
    \text{argmax}_t \  \text{softmax}(W_uh_l(\text{norm}(x)))
\end{equation}
where $x$ is the input, $W_u$ is the unembedding matrix of the model and the subscript $t$ corresponds to the token.
Figure \ref{fig:logit_lens} shows the logit lens applied to Llama when generating: 
\begin{displayquote}
``Le bateau naviguiait en douceur sur l'\textbf{eau au calme du lac. Le soleil ...}''. 
\end{displayquote}
For each layer (y-axis) and token position in the generation (x-axis), a token is decoded from the internal representation.
The decoded tokens from the middle layers onward are more interpretable, whereas early layers are less interpretable.

\subsubsection{Causal tracing} \label{sec:patching}
Causal tracing \citep{meng2022locating, vig2020investigating} uses causal mediation analysis to identify where facts are stored within a network. 
The method compares corrupted hidden states -- where the information necessary to retrieve the fact has been removed -- with clean hidden states --that successfully output the fact. This approach allows us to identify the part of the network that encodes the fact. 
Further details can be found in Appendix \ref{sec:causal_tracing}. 

\subsubsection{Steering Vectors} \label{sec:steering_vec}
Steering vectors \citep{subramani2022extracting, turner2023activation, panickssery2024steeringllama2contrastive} are used to nudge the behavior of the LLM in the desired direction. The main idea is to add activation vectors during the forward of a model to modify its behavior as follows:
\begin{equation}
    h_l(x) \leftarrow h_l(x) + \gamma v_l,
\end{equation}
where $v_l$ is the steering vector, and $\gamma \in \mathbb{R}^{+}$ is a scalar hyperparameter. Steering vectors are used to nudge the output of the LLM in the desired direction. For example, if we want the output to contain more `love', we can compute a steering vector as follows:
\begin{equation}
    v_l = h_l(\text{love}) -  h_l(\text{hate}).
\end{equation}
Further details can be found in \citet{subramani2022extracting} and  \citet{turner2023activation}.

\section{Datasets} \label{sec:datasets}

\paragraph{LLM-Insight}
We created a dataset to analyze the behavior of LLMs, which we will release alongside this paper. The dataset is specifically designed to study steering in LLMs. 

It includes 72 target words, each paired with 10 prompts and 10 sentences in English, Dutch, French, German and Mandarin. Table \ref{data-example} shows a sample of the data. 
The prompts are designed so the word could appear as the next token, but the prompts are also sufficiently open-ended so that semantically unrelated words can be used to complete the sentence.
For example ``They adapted a" can be completed with the word `animal' as well as `daughter'. 

The sentences provided in the dataset can be used to find steering vectors.
Some words in the dataset naturally form pairs that can be used to create steering vectors, such as the words `good' and `bad'. For words without a natural pairing, such as `thermodynamics', we provide a general set of sentences as the counter set to create the steering vector. Further details on the dataset can be found in Appendix \ref{sec:appendix_dataset}.

\paragraph{City facts \cite{ghandeharioun2024patchscope}}
We use this dataset to investigate how facts in different languages are encoded in LLMs. The task is to provide the capital city of a given country.
For example, when prompted with “The capital of Canada is”, the model should output “Ottawa”. 
This allows us to identify where in the network Ottawa is encoded.
To analyze cross-lingual representations, we augment the dataset by translating these facts into German, Dutch, and French. 

\section{Experiments}

\begin{figure*}[t]
\begin{minipage}{0.49\textwidth}
    \centering
    \subcaption{\aya}
    \includegraphics[width=\textwidth]{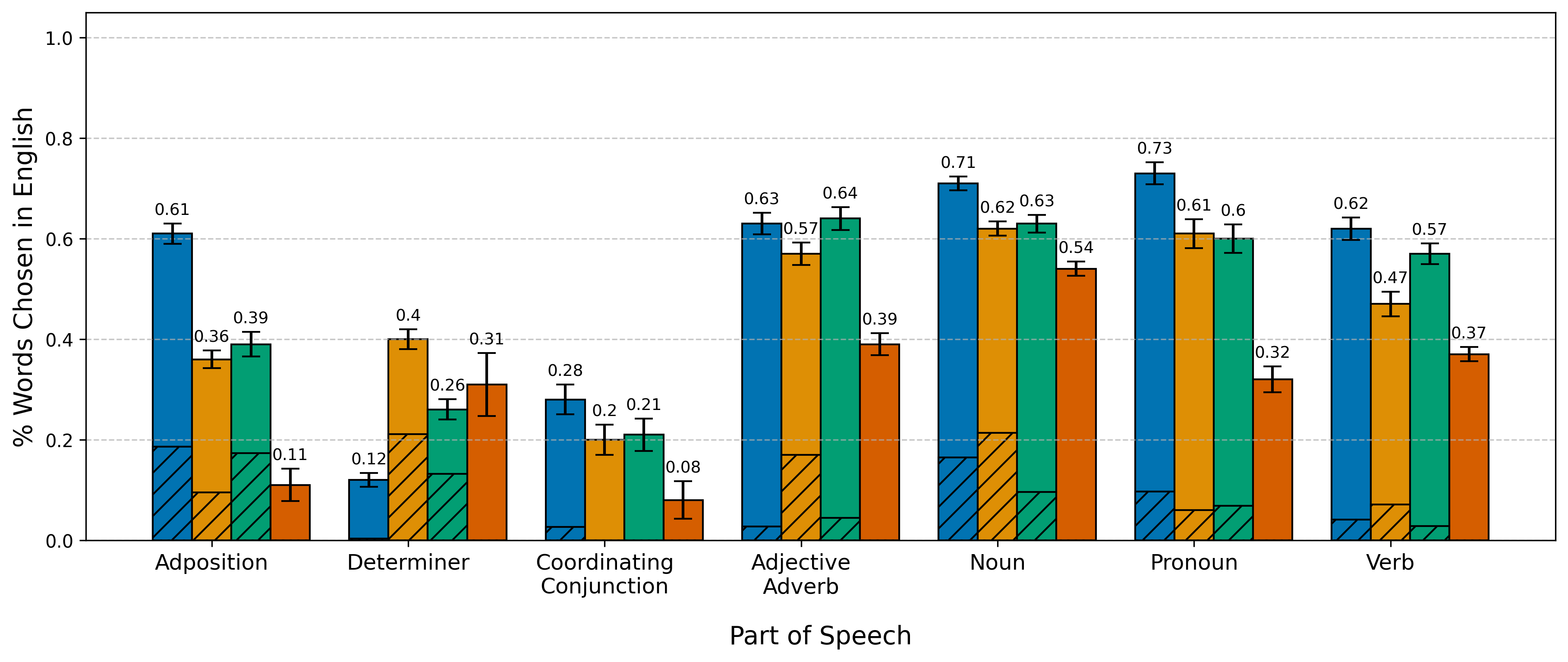} 
     
    \end{minipage}
    \begin{minipage}{0.49\textwidth}
    \centering
     \subcaption{\llama}
    \includegraphics[width=\textwidth]{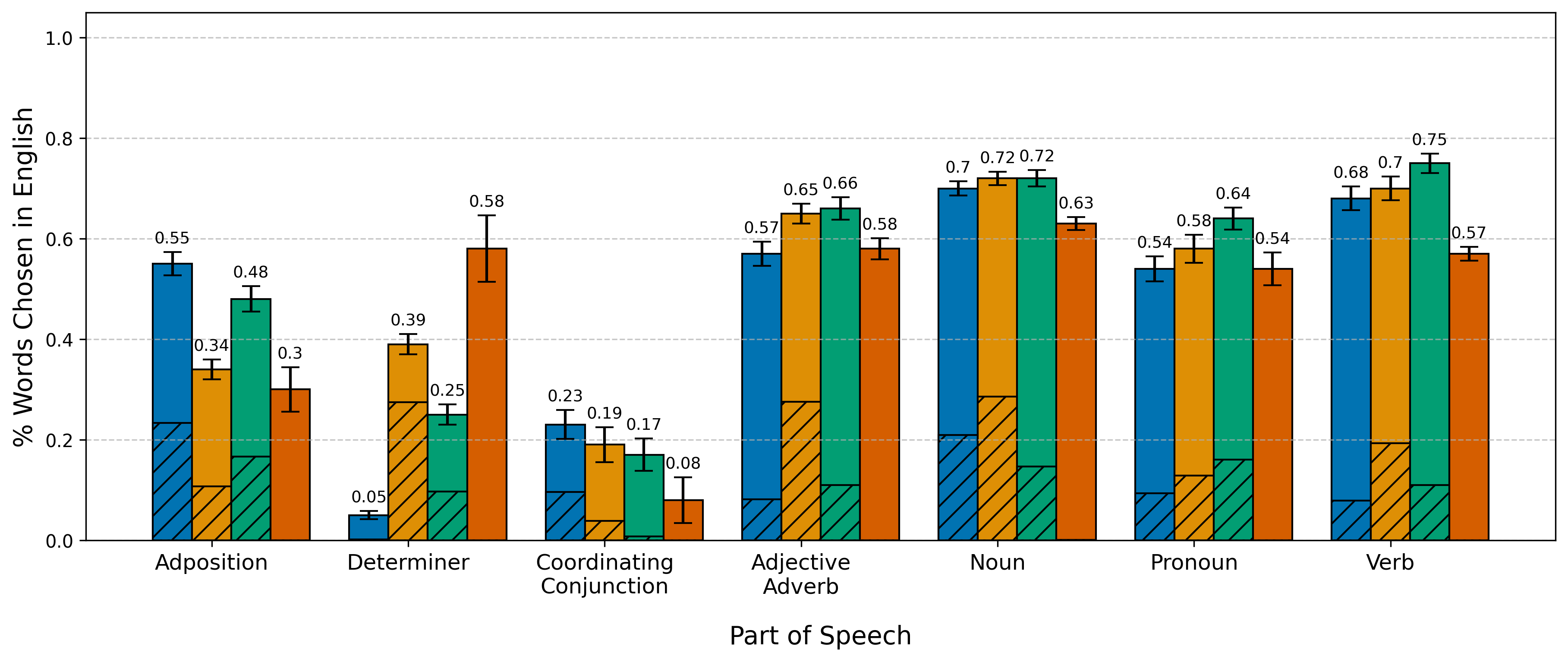} 
    \end{minipage}
    \begin{minipage}{0.49\textwidth}
    \centering
    \subcaption{\mistral}
    \includegraphics[width=\textwidth]{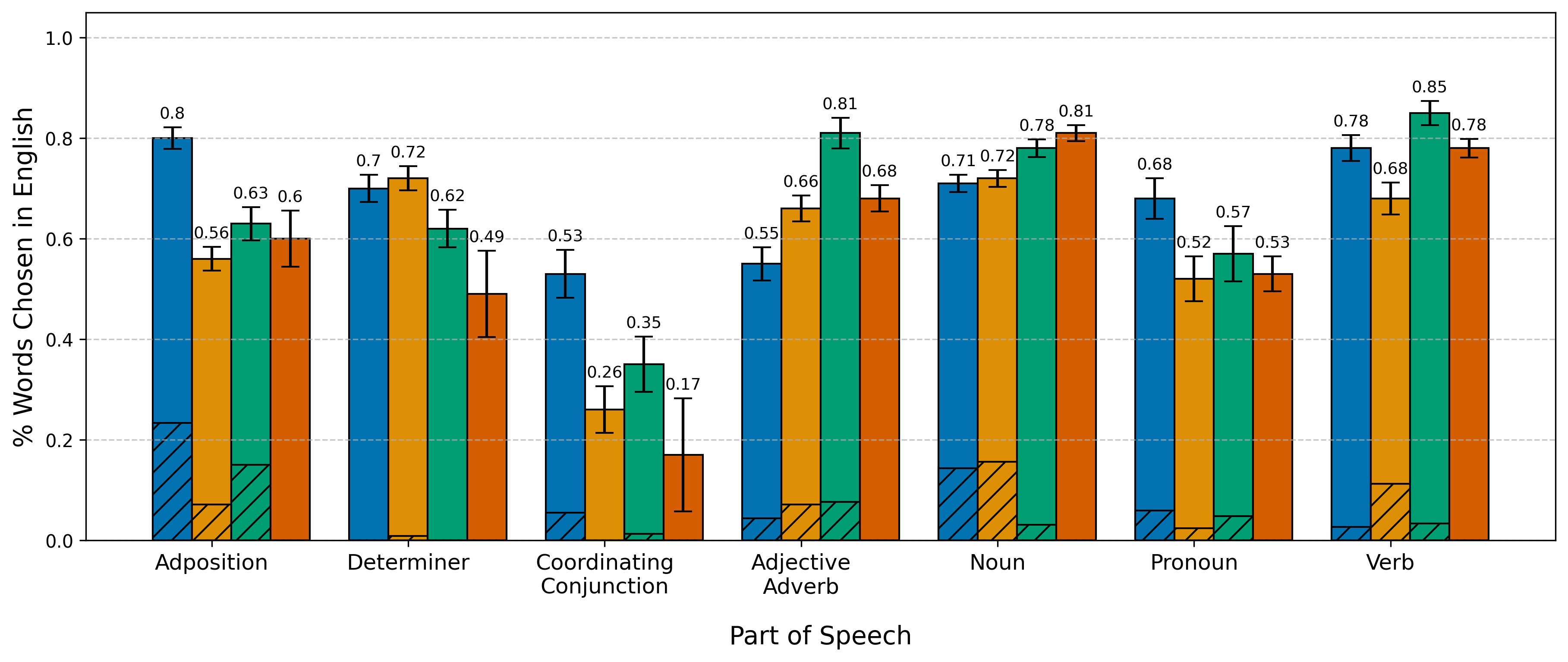}      
    \end{minipage}
    \begin{minipage}{0.49\textwidth}
    \centering
    \subcaption{\gemma}
    \includegraphics[width=\textwidth]{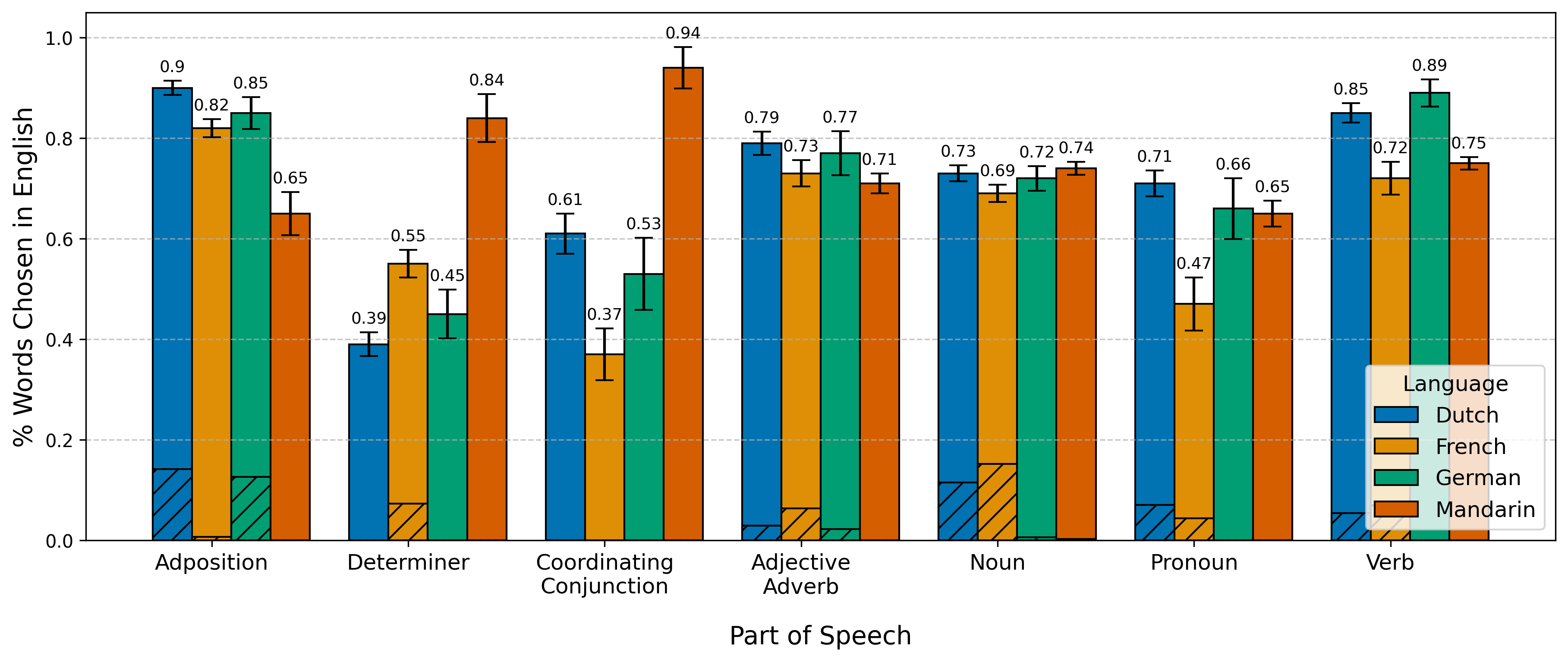} 
    \end{minipage}
    \caption{Logit lens analysis of LLMs routing through English. Each plot shows the proportion of words routed through the English representation space for each model. The shaded bars indicate the portion explained by homographs -- words that are spelled the same in English and the specified language. Overall, the degree of English-routing depends on the model: less diverse pretraining leads to more English-routing. Similarly, most routing occurs for lexical words.}
    \label{fig:pos_plots}
\end{figure*}

We want to understand whether LLMs process prompts differently depending on the output language. 
First, we analyze the latent space to find that LLMs make semantic decisions that are more closely aligned with the English representation space (Section \ref{sec:logit_lens_latent}).
Next, we show that we can steer activations better when using English steering vectors (Section \ref{sec:steering}). Lastly, in Section \ref{sec:latent_space_geometry}, we show that the representations of facts are shared across languages, but have an English-centric bias when decoded.

\subsection{Inspecting the latent space of LLMs using the Logit Lens} \label{sec:logit_lens_latent}

\begin{figure}
    \begin{minipage}{0.5\textwidth}
    \includegraphics[width=\textwidth]{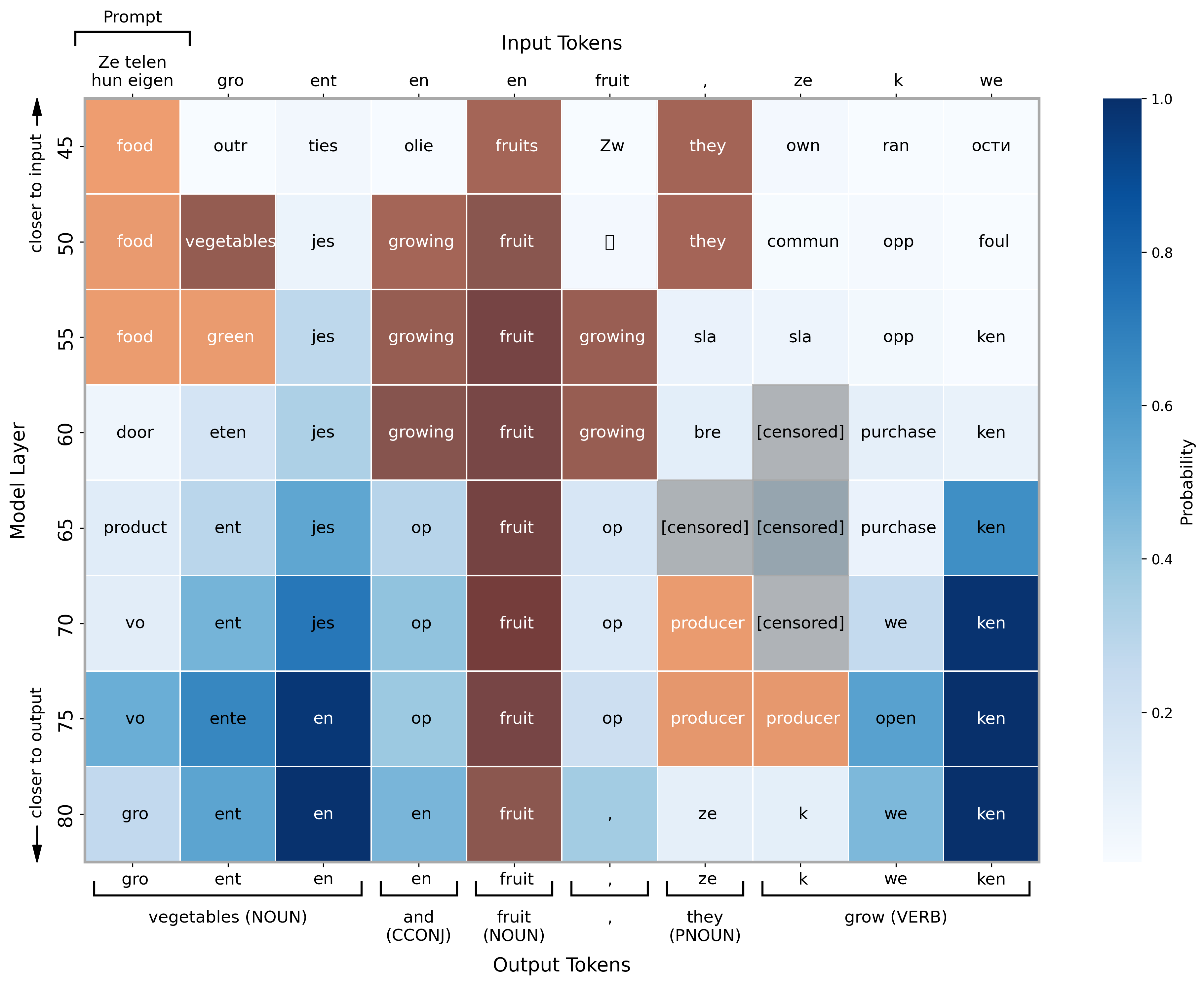} 
    \end{minipage}
     \caption{Logit lens applied to the latent space of \llama, prompted in Dutch with \textit{Ze telen hun eigen}. Each row depicts decoded latent representations across layers, and each column corresponds to the token generated at a specific time step. Orange boxes highlight words selected in English, darker red boxes highlight related words, while gray boxes indicate explicit terms omitted from the figure (see Appendix \ref{sec:explicit_text}). The nouns `fruit' and pronoun `they' are selected in English.}
    \label{fig:logit_lens_nl}
\end{figure}

\paragraph{Qualitative Examples}
To build an intuition on how LLMs operate when prompted in different languages, we analyze their latent space using the logit lens, which decodes the internal representations. 
In Figure \ref{fig:logit_lens}, nouns and pronouns are routed through English, whereas the coordinating conjunction is not. 
Similarly, Figure \ref{fig:logit_lens_nl} shows the logit lens applied to \llama \ for the Dutch prompt \textit{Ze telen hun eigen}. 
The noun `fruit', verb `kweken', and pronoun `they' are all routed through the English words, whereas the coordinating conjunction `en' is not routed through the English word `and'. 
Interestingly, the word growing appears in the latent space several tokens before `kweken' is generated, suggesting that the LLM may plan words in advance in English, which builds on \citet{Pal_2023}'s finding that LLMs encode future tokens in the latent space.

\begin{figure*}[t]
    \begin{minipage}{0.49\textwidth}
    \centering
    \subcaption{\aya}
    \includegraphics[width=\textwidth]{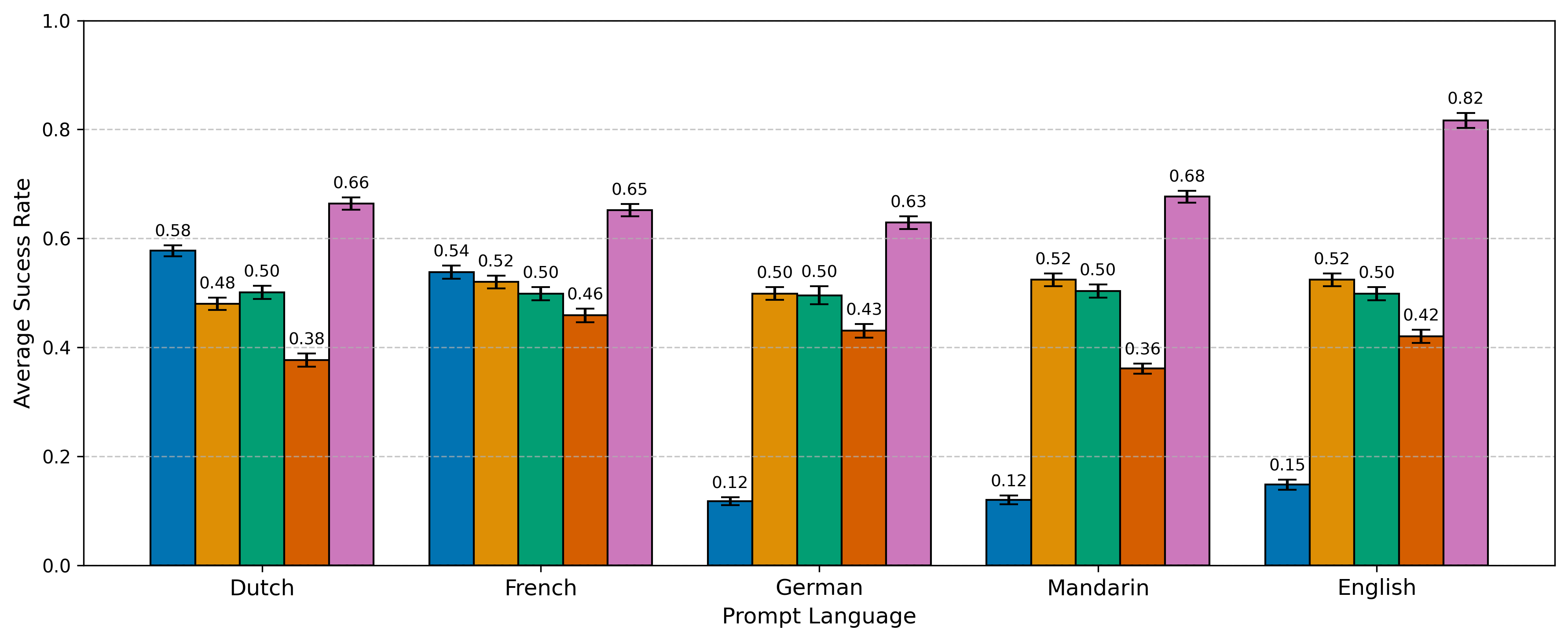} 
    \end{minipage}
    \begin{minipage}{0.49\textwidth}
    \centering
    \subcaption{\llama}
    \includegraphics[width=\textwidth]{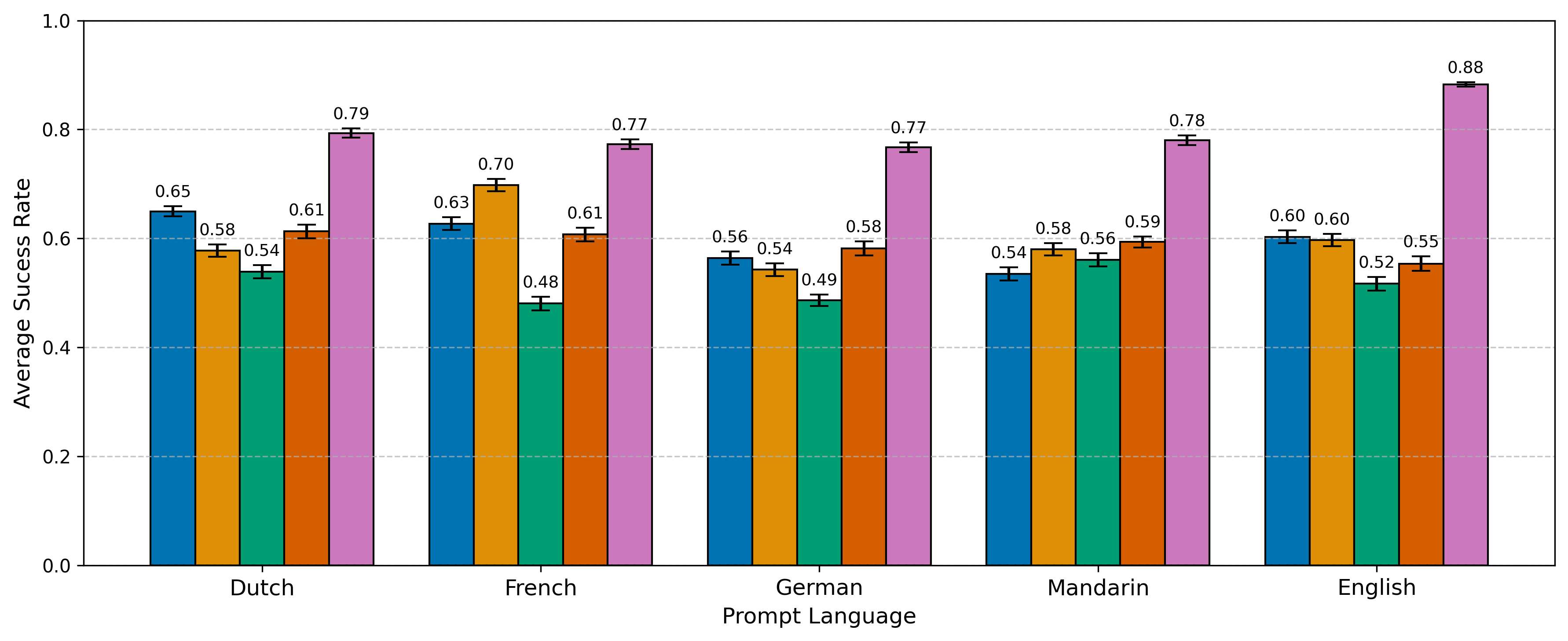} 
    \end{minipage}
    \begin{minipage}{0.49\textwidth}
    \centering
    \subcaption{\mistral }
    \includegraphics[width=\textwidth]{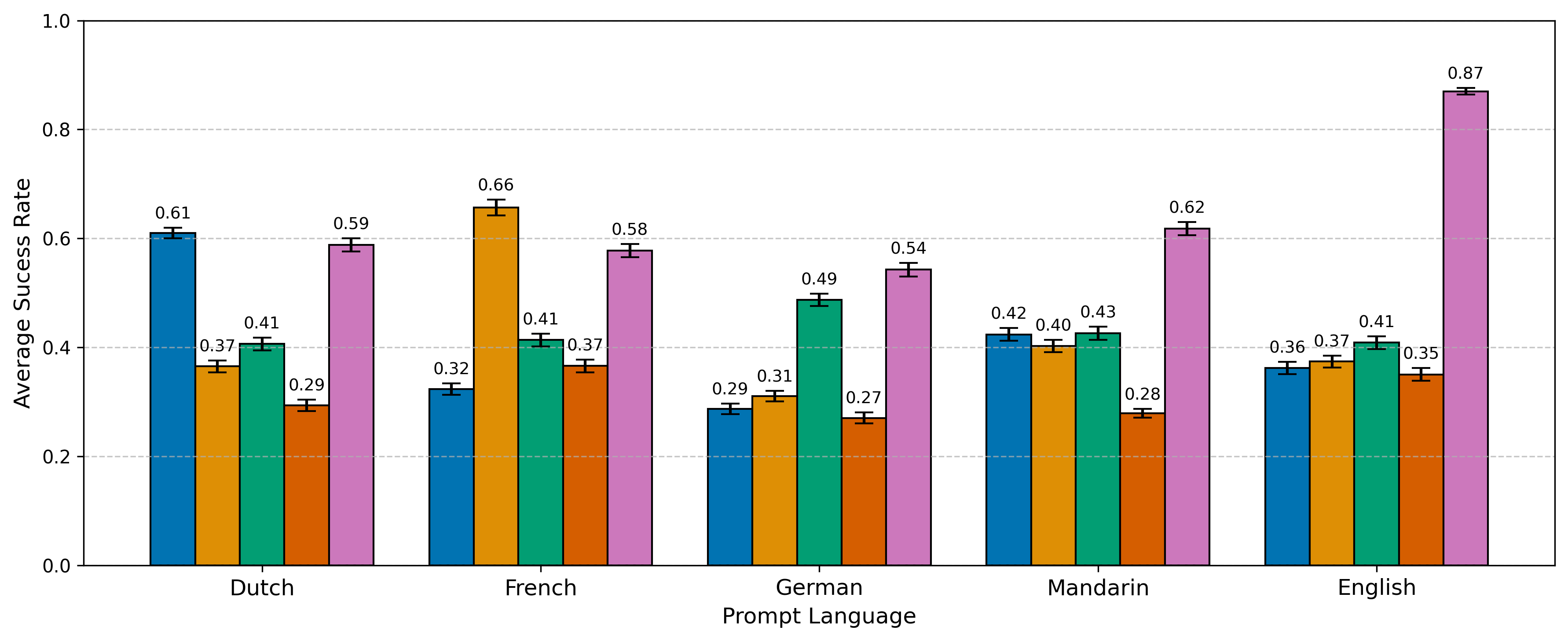} 
    \end{minipage}
        \begin{minipage}{0.49\textwidth}
      \subcaption{\gemma }
    \includegraphics[width=\textwidth]{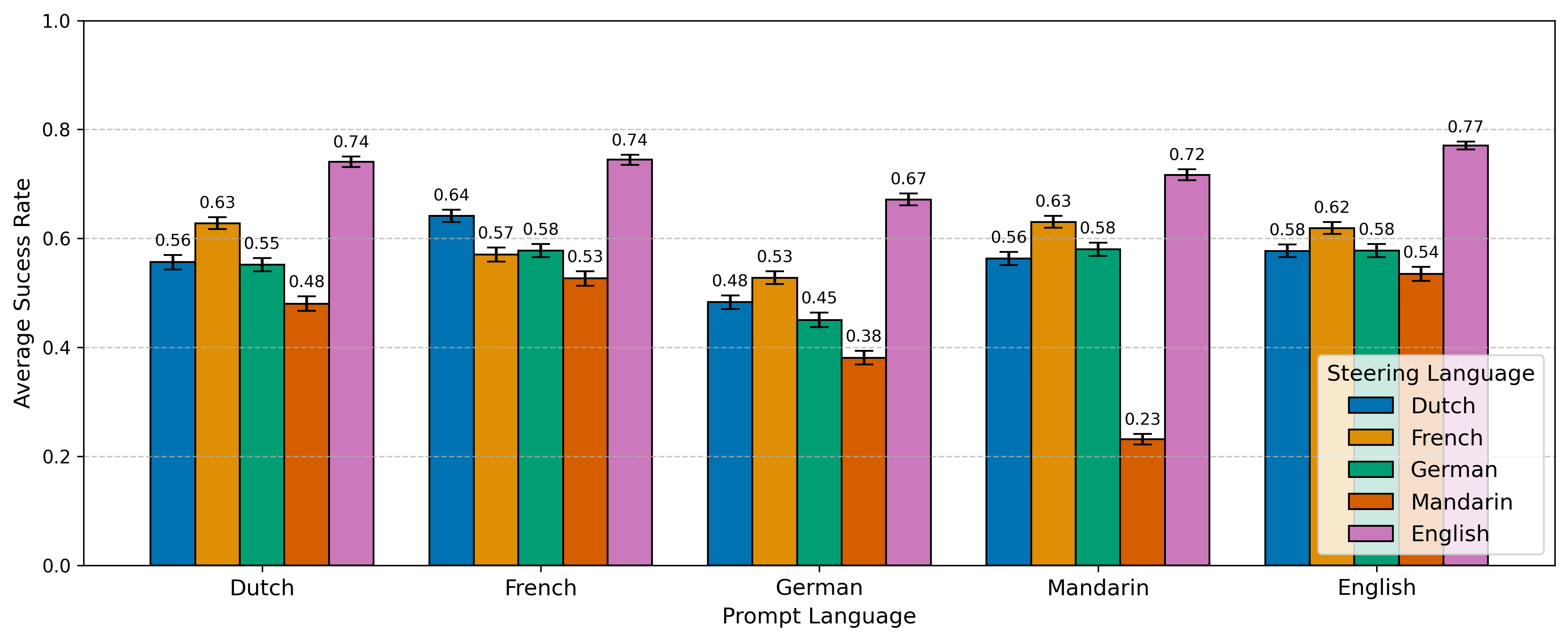} 
    \end{minipage}
    \caption{Cross-Lingual Steering LLMs: The language on the x-axis is the prompt and the desired output language, while the color of each bar indicates the language used to generate the topic steering vectors.}
    \label{fig:steering}
\end{figure*}

\paragraph{Quantitative Evaluation}
The qualitative examples shown in Figures \ref{fig:logit_lens} and \ref{fig:logit_lens_nl} suggest that the part of speech determines whether LLMs employ English routing.
To investigate this, we prompt each LLM to generate $720$ sentences. For each generated word, we evaluate whether the English equivalent of a word appears in the latent space. 
For example, in Figure \ref{fig:logit_lens_nl}, for the word groenten, we check whether the English equivalent, vegetables, appears in the decoded latent space.
We then aggregate the results across different parts of speech. Further implementation details are provided in Appendix \ref{sec:appendix_logit_lens_eval}.

Figure \ref{fig:pos_plots} shows the results for \aya, \llama, \mistral \ and \gemma. Each bar shows the percentage of words that route through the English representation space. The shaded part shows the proportion explained by cross-lingual homographs, words that are the same in English and the specified language (e.g., water in English and Dutch). For homographs, it is not possible to disambiguate whether the word routes through English. 

In general, lexical words -- nouns and verbs --  are often chosen in English. These parts of speech influence the semantic meaning of the sentence. 
Other parts of speech, such as adpositions, determiners and compositional conjugates are infrequently routed through English in \aya \ and \llama.

\begin{table*}[h]
\caption{English-routing in LLMs: percentage of generated words that are routed through English. \aya \ shows the least routing behavior, whereas \gemma \ shows the most routing behavior. }
\label{english-routing-model}
\vskip -0.1in
\begin{center}
\begin{small}
\begin{sc}
\begin{tabular}{l cccc|c}
        \toprule
        Model & Dutch & French & German & Mandarin & Average \\
        \midrule
        \gemma & 0.72 $\pm$ 0.01 & 0.67 $\pm$ 0.01 & 0.72 $\pm$ 0.01 & 0.71 $\pm$ 0.01 & 0.70 $\pm$ 0.00 \\
        \mistral & 0.69 $\pm$ 0.01 & 0.63 $\pm$ 0.01 & 0.71 $\pm$ 0.01 & 0.69 $\pm$ 0.01 & 0.68 $\pm$ 0.01 \\
        \llama & 0.51 $\pm$ 0.01 & 0.57 $\pm$ 0.01 & 0.58 $\pm$ 0.01 & 0.55 $\pm$ 0.01 & 0.55 $\pm$ 0.00 \\
        \aya & 0.58 $\pm$ 0.01 & 0.49 $\pm$ 0.01 & 0.50 $\pm$ 0.01 & 0.41 $\pm$ 0.01 & 0.50 $\pm$ 0.00 \\ 
        \bottomrule
\end{tabular}
\end{sc}
\end{small}
\end{center}
\vskip -0.1in
\end{table*}

The degree of English routing is model-dependent, as shown in Table \ref{english-routing-model}. 
One explanation is the degree of multilingualism in the pre-training data -- with more multilingual models, such as \aya, routing less through English, in contrast to the least multilingual model, \gemma, which routes the most through English.
However, this does not account for the differences observed between \mistral \ and \llama, for which French and German are both high-resource languages.
Another possible explanation is model size. Smaller models, such as \mistral\ and \gemma, route through English more frequently than larger models, potentially due to their more limited representation space.

\subsection{Cross-Lingual Steering} \label{sec:steering}

Our experiments in Section \ref{sec:logit_lens_latent} suggest that LLMs may first select topic words in an English representation space, before translating them into the output language in the later layers. 
To further investigate this hypothesis, we evaluate whether non-English model outputs can be modified using English steering vectors.

More concretely, we test whether we can steer models to generate a sentence in a specified output language using two types of steering vectors:
    \begin{itemize}
        \item \textbf{topic steering vector} -- encourages the LLM to generate a sentence with the given topic, such as animals.
        \item \textbf{language steering vector} -- encourages the model to generate text in the desired output language. 
    \end{itemize}
We evaluate the effectiveness of steering across various topics and prompts, using the LLM-Insight dataset (see Section \ref{sec:datasets}). 
We evaluate steering as successful if the generated sentence includes the target word associated with the steering vector while avoiding output collapse -- incoherent sentences or stuttering. 

Figure \ref{fig:steering} shows results when steering different LLMs.
In general, we observe that English steering vectors perform the best -- outperforming steering vectors generated using the desired output language. 
This suggests that the representation space is not universal -- if it were, we would expect the cross-lingual performance to be roughly equal across languages. Instead, this supports the hypothesis that these models select these words in English.

\paragraph{How similar are the steering vectors generated in different languages?}
The steering vectors for the same concepts generated in different languages have a relatively high cosine-similarity, particularly in the early middle layers (see Appendix \ref{sec:appendix_geo}). However, the steering vectors are not language-agnostic -- part of the dissimilarity of the steering vectors can be attributed to the difference in the language used to generate the vectors. This further supports the argument that the representation space is not universal. 

\subsection{Investigating the Representation Space} \label{sec:latent_space_geometry}

\begin{figure}
    \centering
    \includegraphics[trim={12.7cm 0cm 0cm 0cm},clip,width=0.48\textwidth]
    {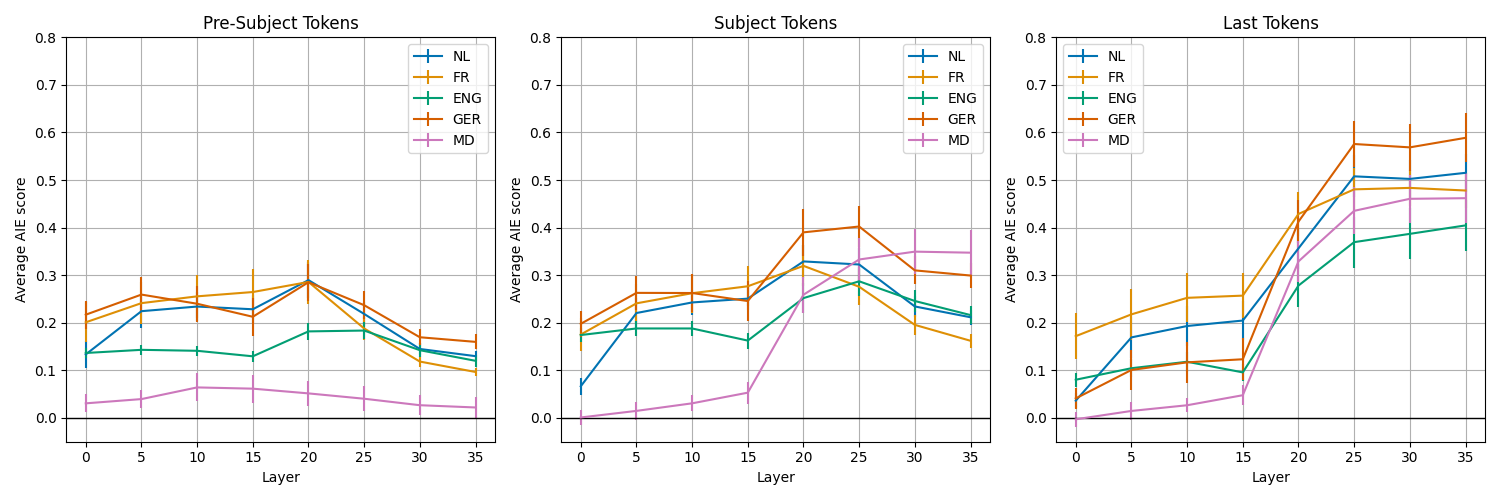}
    \caption{Causal traces for the City Facts dataset in \aya. The AIE scores are similar across different languages, suggesting that facts are localized in the same area of the model.  }
    \label{fig:aya_tracing}
\end{figure}

In this section, we study \textit{how} cross-lingual facts are encoded relative to each other using the city facts dataset (see Section \ref{sec:datasets}). 
First, we perform causal tracing to determine whether facts in different languages are encoded in the same part of the model. Figure \ref{fig:aya_tracing} shows the causal traces for \aya \ (see Appendix \ref{appendix:tracing} for other LLMs). We find that facts are generally localized in similar layers, regardless of the language.

Next, we want to understand if the representation of a fact is shared across different languages. In particular, if we have the same fact in two different languages, such as English and Dutch, can we decompose the representation as follows:
\begin{align} 
    h(\text{capital of Canada}) = h_{Ottawa} \ + h_{English} \\
    h(\text{hoofdstad van Canada}) = h_{Ottawa} \ + h_{Dutch},
\end{align}
where $h$ represents a vector in the latent space. If the above equations hold, we may be able to interpolate between the facts: 
\begin{align*}
    & \alpha h(\text{hoofdstad van Canada})  + (1-\alpha)h(\text{capital of Canada}),  \\ 
    & = h_{Ottawa} \ + \alpha h_{Dutch} + (1-\alpha)h_{English} 
\end{align*}
If we pushforward the interpolated hidden state, and the output is correct, then this suggests that we may be able to disentangle the language and semantic context. 

\begin{figure}[h]
\begin{minipage}{0.5\textwidth}
    \centering
    \includegraphics[trim={1cm 0.5cm 1.5cm 1.5cm},clip,width=\textwidth]{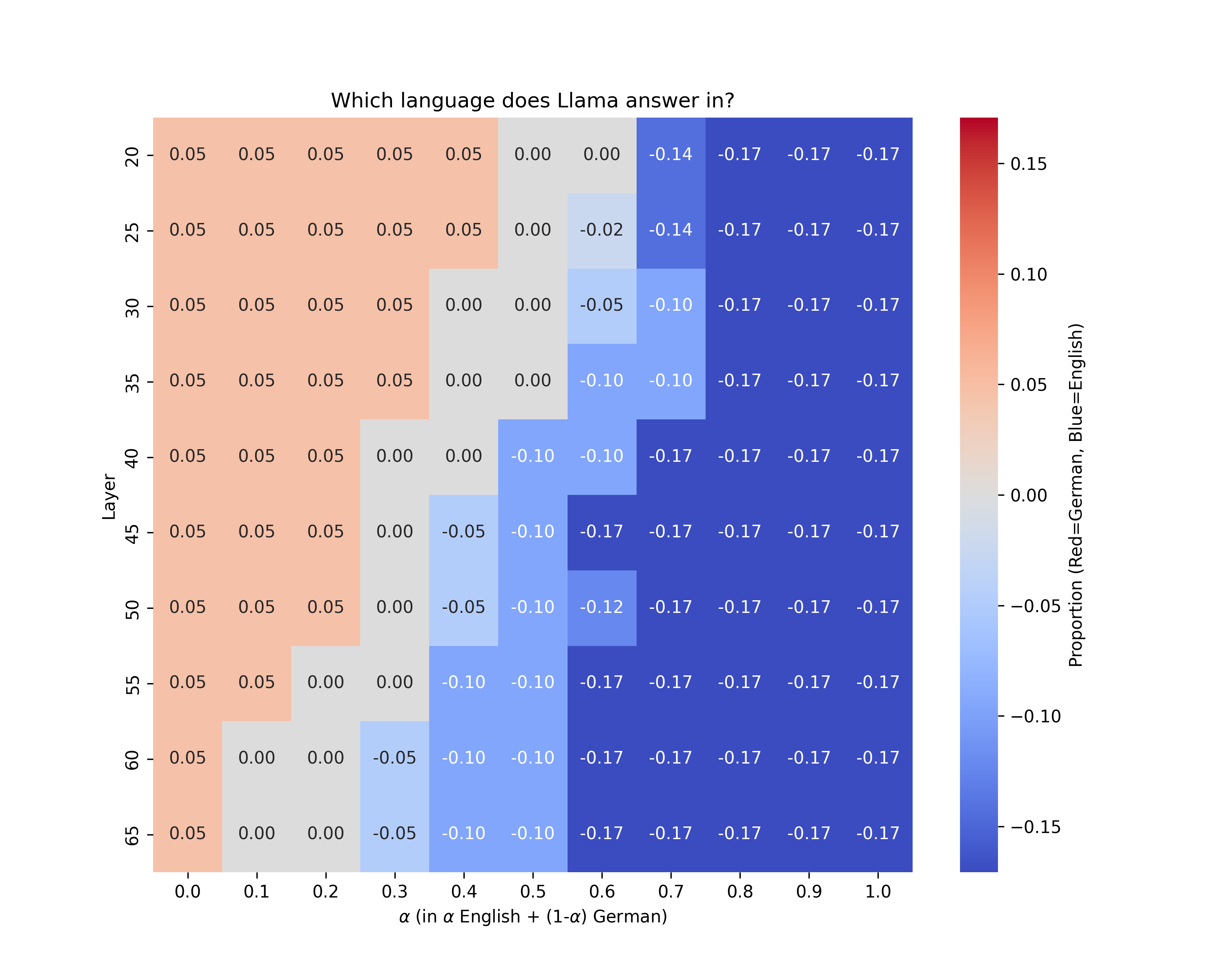} 
    \caption{Relative propensity of \llama \ to answer in German (red) vs English (blue). \llama \ is most likely to answer in English.}
\end{minipage}
\end{figure}

We find that we can interpolate between the hidden states without significant changes in accuracy; the accuracy generally interpolates between the accuracies of the two languages (see Appendix \ref{sec:hidden_state_interpolation}). Furthermore, we find that models have a propensity to answer in English.
This provides further evidence that the models likely operate in an English-centric space.

\section{Limitations}
Our work provides evidence suggesting that MLLMs primarily operate in English.
Below, we outline potential limitations and directions for future research.

\paragraph{Tokenization}
Sentences in different languages often vary in tokenization length \citep{rust-etal-2021-good, muller-etal-2021-unseen,petrov2024language}, which complicates cross-lingual comparisons. 
In this work, we provide heuristics (e.g., for causal tracing, which operates on a per-token level) to compare the results when tokenization lengths vary. 
However, tokenization remains an important consideration for the development of future interpretability methods designed to be used across multiple languages.

\paragraph{Language confidence and confusion}
Models often assign higher probabilities to outputs in certain languages, which can affect analyses such as causal tracing by requiring higher noise levels. Similarly, models often exhibit language confusion \citep{marchisio2024understandingmitigatinglanguageconfusion}, continuing to respond in English even when prompted in other languages. 
Both factors influence our analysis. We can mitigate some issues associated with the first problem -- e.g., in causal tracing, we ensure the probabilities all fall below a specified threshold when a prompt is noised. However, we do not actively address language confusion, as doing so could alter the natural behavior of the LLMs, which we aim to understand.

\paragraph{Factors affecting interpretability methods}
Interpretability methods are influenced by various factors. Steering performance, for example, depends on the intrinsic steerability of a prompt \citep{turner2023activation, tan2024analyzinggeneralizationreliabilitysteering}. To address this, we designed a custom dataset that, to the best of our knowledge, is equally steerable across all languages.
Another challenge is that steering could push activations outside the expected data distribution, leading to unintended outputs. 
To mitigate this, we checked for stuttering in the generated outputs. 
However, further work is needed to deepen our understanding of steering mechanisms and to develop more robust evaluation procedures.

\paragraph{Other Methods}
Exploring alternative methods could provide valuable insights. For example, sparse autoencoders (SAEs) \citep{olshausen1997sparse, hinton2006reducing, templeton2024scaling} are a popular interpretability tool. 
However, training SAEs for each layer is computationally expensive and beyond our computational budget. 
While some pre-trained SAEs are available, they are predominantly trained on English data, which introduces biases we aim to avoid \citep{lieberum2024gemmascopeopensparse}.

\section{Related Work}

We can think about understanding a model from two perspectives:
\begin{itemize}
    \item an \textbf{internal perspective}, focused on analyzing the model through the latent space and operations performed inside of the model. Examples of questions include: how do models represent knowledge across different languages? How does a model retrieve facts? 
    \item an \textbf{external perspective}, focused on analyzing the model output. For example, how well do models perform in different languages? 
\end{itemize}
Having a unifying theory that combines both perspectives is important -- the internal perspective helps us understand the mechanisms underlying behavior, while the external perspective examines the real-world impact of that behavior. Below, we summarize the research on multilingual language models from both perspectives.

\subsection{How do LLMs operate internally?}
The current main theory in mechanistic interpretability suggests that there are three general phases in the forward pass of an LLM \cite{olahfeaturesllm, nanda2023progress, wendler2024llamasworkenglishlatent, dumas2024llamas, fierro2025multilinguallanguagemodelsremember}:
\begin{enumerate}
    \item \textbf{Detokenization}:  In this phase, individual tokens are combined into abstract units that the model uses for analysis. These units can be referents -- for example, \cite{nanda2023progress} found evidence that the tokens [Michael] and [Jordan] are combined into a unit representing the basketball player Michael Jordan. Similarly, these units can encode instructions, as shown by \cite{dumas2024llamas}, where the model extracts the target language during translation tasks in these layers.
    \item \textbf{Processing}: In this phase, the model processes or reasons over abstract units. For instance, this stage may involve tasks like fact recall \cite{geva-etal-2023-dissecting, nanda2023progress}.
    \item {\textbf{Selecting the output}}: In this phase, the model selects the output. This may involve selecting the correct attribute \cite{nanda2023progress}, mapping an abstract concept to the corresponding word in the target language \cite{wendler2024llamasworkenglishlatent} and/or selecting the correct token for the intended word.  
\end{enumerate}

In the context of multilingual models, an important question is whether the concept space (in phase 2) is universal. 
Here, \textit{universal} means the representation is \textit{shared} across languages, i.e., the representation for `cat' (cat in English) and `kat' (cat in Dutch) is the same.

One stream of research argues that the concept space is \textit{universal}. When analyzing the latent space with the logit lens, \citet{wendler2024llamasworkenglishlatent} 
find that the concept space is language-agnostic, but more closely aligned with the English output space.
In their follow-up work, \citet{dumas2024llamas} used tracing to find further evidence of language-agnostic representations of concepts.\footnote{They used tracing to mitigate potential shortcomings of cosine-similarity \citep{steck2024cosine}}.
Concurrent to this work, \citet{brinkmann2025large} showed that models share morpho-syntactic concept representations across languages.

Other researchers found that the concept space is biased towards the training-dominant language. 
Using the logit lens, \citet{zhong2024beyond} focused on Japanese, showing that Swallow \citep{fujii2024continual}, which is fine-tuned on Japanese, and LLM-jap \cite{aizawa2024llm}, pre-trained on Japanese and English, are Japanese-centric. 
\cite{fierro2025multilinguallanguagemodelsremember} found that subject enrichment -- retrieving attributes about the subject -- is language-agnostic whereas object extraction is language-dependent, when studying EuroLLM \citep{martins2024eurollmmultilinguallanguagemodels}, XLGM \citep{lin2022fewshotlearningmultilinguallanguage} and mT5 \citep{xue-etal-2021-mt5}.

Overall, our findings suggest the presence of a language-specific latent space. Specifically, our results indicate that the concept space is largely English-centric, consistent with prior work \citet{zhong2024beyond, wu2024semantic}, where English is the dominant training language. However, unlike previous studies, we uncover additional nuance: non-lexical words are not necessarily routed through the English representation space.
Moreover, we find that the behavior varies across models. One potential explanation is that we focus on open language generation -- which is different from prior work that predominantly focuses on single-token and translation tasks. 
We provide a more in-depth discussion in Section \ref{sec:discussions}.

\subsection{Multilingual LLM behavior}

The internal mechanisms of LLMs affect their performance in several different ways, which we summarize below.

\paragraph{Performance}
The performance of multilingual language models varies across languages \cite{shafayat2024multifactassessingfactualitymultilingual, huang2023languagescreatedequalllms, bang2023multitaskmultilingualmultimodalevaluation, shi2022language, ahuja2023megamultilingualevaluationgenerative}, often performing best in English.
This can even be leveraged to improve performance in other languages, for example, through cross-lingual chain-of-thought reasoning \cite{chai2024xcotcrosslingualinstructiontuning}, or by modifying prompts, such as using multilingual instructions or asking the LLM to translate the task into English before completing it \cite{zhu2023extrapolatinglargelanguagemodels, etxaniz2023multilinguallanguagemodelsthink}.

\paragraph{Fluency and language confusion}
\citet{marchisio2024understandingmitigatinglanguageconfusion} has shown that English-centric models are prone to language confusion, i.e., providing the answer in the incorrect language. 
Moreover, even when LLMs output text in the correct language, they can produce unnatural sentences in other languages, akin to an accent \cite{guo2024benchmarking}. 

\paragraph{Bias and culture}
 Moreover, LLMs tend to be biased toward certain cultures, with content performing better when dealing with facts originating from Western contexts \cite{naous2024havingbeerprayermeasuring, shafayat2024multifactassessingfactualitymultilingual}, falling short when answering questions on other cultures \cite{chiu2024culturalbenchrobustdiversechallenging}.
\citet{liu2024multilingualllmsculturallydiversereasoners} investigate the cultural diversity of LLMs using proverbs and find that these models often struggle to reason with or effectively use proverbs in conversation. Their understanding appears limited to memorization rather than true comprehension, creating a notable ``culture gap" when translating or reasoning with culturally specific content across languages.

\section{Conclusion}

Our results provide evidence that semantic decisions in LLMs are predominantly made in a representation space close to English, while non-lexical words are processed in the prompt language. 
However, we find that this behavior varies across models, likely due to differences in multilingual proficiency and model size. 
The English-centric behavior is further validated by our findings that steering non-English prompts using vectors derived from English sentences is more effective than those from the prompt language.

Exploring the structure of the latent space, we find that factual knowledge across languages is stored in roughly the same regions of the model. 
Interpolating between the latent representations of these facts in different languages preserves predictive accuracy, with the only change being the output language. 
This suggests that facts encoded in different languages likely share a common representation. 
However, when interpolating, we find that model output is most frequently in English, further underlining the English-centric bias of the latent space.

The English-centricity of the latent space is consistent with prior observations about LLM behavior. 
In particular, \citet{etxaniz2023multilinguallanguagemodelsthink} found that instructing LLMs to first translate a non-English prompt into English improves model performance. 
However, this bias can be detrimental. If the latent space is English-centric, this may lead the LLMs toward exhibiting Western-centric biases \citep{naous2024havingbeerprayermeasuring, shafayat2024multifactassessingfactualitymultilingual}. 

\section{Discussion} \label{sec:discussions}
There are currently two perspectives in interpretability research on concept representations in multilingual models: (1) concept representations are universal; and (2) concepts have language-centric representations, where the language is the training-dominant language. 
Our work aligns more closely with the second perspective, as well as a third perspective -- namely, that LLMs encode language-specific representations, where the language is the input/output language. Below, we discuss how the different theories may be reconciled. 

\citet{wendler2024llamasworkenglishlatent} and \citet{dumas2024llamas} argue that 
the concept space is universal, but likely more aligned with the English output space. 
However, our findings contest this conclusion, as we find that interventions in the latent space are more effective when using English text, even when the target language is not English. 
If the concept space were truly universal, we would expect interventions using all languages to perform equally well. Our findings are consistent with concurrent work by \citet{wu2024semantic}, who similarly find that steering using English performs comparably to, or slightly better than, the target language.

One possible way to reconcile the two theories is via the difference between concepts that are encoded and concepts that are used \citep[as discussed in][]{brinkmann2025large}. 
There may be multiple representations of any given concept \citep{hase, mcgrath2023hydraeffectemergentselfrepair}, or a concept may be represented in an LLM but not used during generation.
\citet{wendler2024llamasworkenglishlatent} focus more on the encoding of concepts, whereas our work focuses more on the generation of text. 

An alternative explanation is that different behavior is captured in the tasks. In \citet{wendler2024llamasworkenglishlatent, dumas2024llamas}, the tasks are designed to generate a single token. In this setting, the task is to select the correct token, and we expect a high probability mass on a single token. In contrast, we focus on a more open-ended setting where there are several different possible continuations. These two settings are inherently different, leading to different conclusions about the behavior of LLMs. Even within the same task of fact retrieval, prior work found that different components of the forward pass are language-specific and language-agnostic \citep{fierro2025multilinguallanguagemodelsremember}. 

More generally, the open-generation setting allows us to analyze different parts of speech. 
This leads to the second main difference in conclusions, which is that LLMs encode language-specific representations. 
For semantically loaded words, we find evidence that the latent space is English-centric (in LLMs where English is the dominant training language). 
This is consistent with one line of prior work, which generally focuses on nouns \citep{wu2024semantic, zhong2024beyond}.
However, we find that the same pattern does not hold for non-lexical words. 

This is in contrast to concurrent work by \citet{brinkmann2025large}, who showed that models share morpho-syntactic concept representations across languages in Llama-3-7b and Aya-8B. In line with their previous work \citep{wendler2024llamasworkenglishlatent}, they argue that the representations are universal. 
While our high-level conclusions differ, our findings also support the hypothesis that smaller models emit more shared representations than larger models, which permit more language-specific representations. 

In summary, our findings indicate that the extent to which representations are shared across languages is more nuanced than previously thought. 
Contrasting our work with previous work suggests that the task and model size likely influence the observed behavior. 
Fully understanding these nuances is important to ensure the fairness and robustness of LLMs.

\section*{Impact Statement}
Large Language Models (LLMs) are increasingly deployed across a wide range of applications, making it crucial to understand and evaluate their performance to ensure both safety and fairness. 
A key characteristic of LLMs is their English-centric nature, which influences their behavior, as shown in this paper. 
In particular, we find further evidence that LLMs perform semantic decisions in English. 
This likely leads to biased behaviour \citep{naous2024havingbeerprayermeasuring, shafayat2024multifactassessingfactualitymultilingual}. 
Understanding this is essential to equitable and reliable outcomes in diverse linguistic and cultural contexts.

Moreover, our findings may further be relevant for improving the safety of LLMs. 
When analysing non-lexical words, we found that LLMs do not emit an English-centric bias. 
If knowledge representation is not universal across languages, LLMs may require language-specific safety tuning. 
When introspecting the latent space, we observed varying levels of vulgar terms depending on the language, particularly in cases where the models are not safety-tuned on the language.
While this does not necessarily mean the model’s output will be vulgar, it could make the model more vulnerable to jailbreaks \citep{deng2024multilingualjailbreakchallengeslarge, ghandeharioun2024whosaskinguserpersonas}.

\section*{Acknowledgments}
We thank Jannik Kossen, Kunal Handa, Emily Reif, Veniamin Veselovsky and Lewis Hammond for their useful feedback and discussions during the writing of this paper. 
Lisa Schut is funded by the EPSRC Centre for Doctoral Training in Autonomous Intelligent Machines and Systems grant (EP/S024050/1) and DeepMind.
Yarin Gal is supported by a Turing AI Fellowship financed by the UK government’s Office for Artificial Intelligence, through UK Research and Innovation (EP/V030302/1) and delivered by the Alan Turing Institute.

\bibliographystyle{abbrvnat}
\bibliography{references.bib}

\appendix
\onecolumn

\section{Appendix}

\subsection{Causal Tracing} \label{sec:causal_tracing}
Causal tracing \citep{meng2022locating, vig2020investigating} uses causal mediation analysis to identify where facts are stored within a network.
For example, imagine that we want to find where the fact ``The capital of Canada is Ottawa" is represented in an LLM. 
We could prompt the model with ``The capital of Canada is" to find where ``Ottawa'' is stored in the network.
There are two main steps in causal tracing:
\begin{enumerate}
    \item \textbf{corrupt the signal}: destroy the information so that the model no longer outputs the fact. 
    \item \textbf{restore the signal}: determine where in the network the representation needs to be restored so that the LLM can recover the correct output.
\end{enumerate}

Let $e^{\text{clean}} \in \mathbb{R}^{m,d}$ be the embedding of the prompt ``The capital of Canada is'', where $m$ is the number of tokens and $d$ is the embedding dimension. 
In the first step, the information is ``destroyed" by adding noise to the embedding of the subject token:
\begin{equation}
  e^{\text{corrupted}}_{j} =
    \begin{cases}
      e^{\text{clean}}_{j} + \varepsilon & \text{if token $j$ is a subject token}\\
      e^{\text{clean}}_{j} & \text{otherwise}
    \end{cases}       
\end{equation}
where $\varepsilon $ is noise sampled from an isotropic Gaussian distribution, and $e^{\text{corrupted}}$ is the corrupted embedding.
We pushforward corrupted embeddings $e^{\text{corrupted}}$ through the network to obtain the probability that the model outputs Ottawa, $p[\text{Ottawa}|e^{\text{corrupted}}]$.

Next, we want to find out which part of the hidden states encodes the relevant information to restore the correct output. 
At a given layer $l$ in the network, we `restore' part of the corrupt hidden state by copying back part of the clean hidden state $h$ at position $p$:
\begin{equation}
  h^{\text{restored}}_{j,l} =
    \begin{cases}
      h^{\text{clean}}_{j,l} & \text{if} \  j = p\\
      h^{\text{corrupted}}_{j,l} & \text{otherwise},
    \end{cases}       
\end{equation}
where the hidden state $h^{\text{clean}}_{l} = [h^{\text{clean}}_{1,l}, \dots, h^{\text{clean}}_{m,l}]$ is obtained by pushing the original embeddings, $e^{\text{clean}}$, through the network. 

Finally, we propagate $h^{\text{restored}}_l$ through the remaining layers produce the output probability $p[\text{Ottawa}|h^{{\text{restored}}}_{l,p}]$.
The difference $p[\text{Ottawa}|h^{{\text{restored}}}_{l,p}] - p[\text{Ottawa}|h_l^{\text{corrupted}}]$, measures the importance of layer $l$ and token position $p$ in encoding a fact. 
Through this approach, causal tracing helps identify which parts of the representation are sufficient to retrieve the correct output.

\subsection{LLM Training Data Languages}

Table \ref{model} summarizes the languages the different models are trained on. 

\begin{table*}[h]
\caption{LLMs: High resource training languages}
\label{model}
\vskip 0.15in
\begin{center}
\begin{small}
\begin{sc}
\begin{tabular}{p{6cm} p{8cm}} 
\toprule
{\bf Model} & {\bf Languages} \\ \midrule 
Llama-3.1-70B \citep{dubey2024llama3herdmodels} & English, German, French, Italian, Portuguese, Hindi, Spanish, and Thai \\  
Mixtral-8x22B-v0.1 \citep{jiang2024mixtralexperts} & English, French, Italian, German, and Spanish \\  
Aya-23-35B \citep{aryabumi2024aya} & Arabic, Chinese (simplified and traditional), Czech, Dutch, English, French, German, Greek, Hebrew, Hindi, Indonesian, Italian, Japanese, Korean, Persian, Polish, Portuguese, Romanian, Russian, Spanish, Turkish, Ukrainian, and Vietnamese \\  
Gemma-2-27b  \citep{gemma_2024} & English \\ 
\bottomrule
\end{tabular}
\end{sc}
\end{small}
\end{center}
\vskip -0.1in
\end{table*}

\subsection{LLM-Insight Dataset} \label{sec:appendix_dataset}

Our goal is to generate a dataset that can be used for cross-lingual interpretability. We wanted a dataset that can be used to introspect LLM internal representations and analyze LLM behavior when the internal representations are intervened on.
Additionally, the dataset focuses on open-ended sentence generation rather than being restricted to specific tasks like fact recall or sentiment analysis, as text generation is an important real-world application. 

\subsubsection{Text generation}
We use GPT-4o to generate sentences and prompts. For each target word, we generate:
\begin{itemize}
    \item 10 unique sentences containing a version of the word -- for example, for the verb `(to) see', a suitable sentence is `She saw a bird in the sky.'
    \item a list containing the version of the word used in each sentence. In the previous example, the version of the word is `saw'. 
    \item 10 unique prompts, designed to be completed with the target word.
    \item a list containing a version of the word used in each sentence
\end{itemize}

We instruct GPT-4o to generate prompts that can be completed with the target word, as well as semantically distinct words. 
However, we observe that the model sometimes produces sentences and prompts that do not meet the criteria.

An example of a sentence that does not meet the criteria is:
\begin{displayquote}
\textbf{Target word}: bouquet (boeket in Dutch) \\
\textbf{Sentence}: Het boeket was gevuld met levendige rozen en lelies. \\
\textbf{Translation}: The bouquet was filled with live roses and lilies.
\end{displayquote}
The issue with this sentence is its unnatural phrasing—the word "live" is not typically used in this context.

An example of a prompt that does not meet the criteria is:
\begin{displayquote}
\textbf{Target word}: money \\
\textbf{Prompt}: He went to the bank to withdraw
\end{displayquote}
In this case, the only plausible continuation is "money." While the prompt is coherent, it lacks the open-endedness needed to analyze how interventions influence model behavior.

An example of a well-constructed prompt is:
\begin{displayquote}
\textbf{Target word}: bus \\
\textbf{Prompt}: She took a 
\end{displayquote}
This can be completed with the intended word "bus", as well as semantically different alternatives such as "walk" or "long road trip".

To ensure data quality, we asked native speakers to review and correct the data. The original version of the data and the corrections are provided in the dataset. 

\FloatBarrier

\subsubsection{Dataset summary}

We selected words that vary in the number of tokens (in non-English languages), whether the word is a homograph with the English version of the word, and the part of speech. Table \ref{word-translations} summarizes the words used.

\begin{longtable}{%
    >{\raggedright\arraybackslash}p{2.5cm} 
    >{\raggedright\arraybackslash}p{2.5cm} 
    >{\raggedright\arraybackslash}p{2.5cm} 
    >{\raggedright\arraybackslash}p{2.5cm} 
    >{\raggedright\arraybackslash}p{2.5cm} 
    >{\raggedright\arraybackslash}p{2cm}}
\caption{Word Translations Across Languages} \label{word-translations} \\ 
\toprule
\textbf{Word} & \textbf{English} & \textbf{Dutch} & \textbf{French} & \textbf{German} & \textbf{Mandarin} \\ 
\midrule
\endfirsthead

\toprule
\textbf{Word} & \textbf{English} & \textbf{Dutch} & \textbf{French} & \textbf{German} & \textbf{Mandarin} \\ 
\midrule
\endhead

\bottomrule
\endfoot    

animal & animal, animals & dier, dieren & animal, animaux & Haustier, Tier, Tiere & \md{动物} \\
bad & bad & slecht, slechte & mal, mauvais, mauvaise, mauvaises & schlecht, schlechte, schlechten & \md{不好, 坏了} \\
ballet & ballet & ballet & ballet & Ballett & \md{芭蕾} \\
bank & bank & bank & banque & Bank & \md{银行} \\
beautiful & beautiful & mooi, mooie & beau, bel, belle, magnifique & schön, schöne, schönen & \md{美丽} \\
big & big, tall & groot, groots, grote & grand, grande, grands, gros & große, großen, großer, großes & \md{大} \\
bouquet & bouquet & boeket & bouquet & Strauß & \md{花束} \\
brother & brother & broer & frère & Bruder & \md{哥哥, 弟弟, 兄弟} \\
bus & bus & bus & bus & Bus & \md{公交车 / 巴士} \\
cat & cat & kat & chat & Katze, Katzen, Mutterkatze & \md{猫, 小猫, 流浪猫} \\
centre & centre & centrum & centre & Zentrum & \md{中心, 市中心, 研究中心, 社区中心, 艺术中心} \\
chair & chair & stoel & chaise & Stuhl, Stuhls, Stühlen & \md{椅子, 椅子, 摇椅} \\
chauffeur & chauffeur, driver & chauffeur & chauffeur, chauffeuse & Chauffeur & \md{司机} \\
child & child & kind & enfant & Kind & \md{孩子} \\
club & club & club, vereniging & club & Club & \md{俱乐部} \\
cold & cold & ijskoud, kou, koud, koude & froid, froide, froides & kalt & \md{冷, 寒冷} \\
computer & PC, computer, laptop, machine, rig, system & computer & ordinateur & Computer, Computern, Laptop & \md{电脑} \\
culture & culture, cultures & culturen, cultuur & culture, cultures, la culture & Kultur & \md{文化} \\
day & day & dag & jour, journée & Tag & \md{天, 日子} \\
dog & dog & hond & chien & Hund & \md{小狗, 狗} \\
drink & drink & drank, drankje, drinken & boire, boisson, drinken & trinken & \md{饮料} \\
eat & eat & eten & dîner, manger, repas & essen & \md{吃} \\
fast & fast & fast, snel, snelle & rapide, vite & schnell & \md{快} \\
film & film & film, films & film & Film, film & \md{电影, 影片, 纪录片} \\
food & cuisine, cuisines, dish, food, meals & gerecht, gerechten, voedsel & cuisine, nourriture & Essen, Futter, Nahrung & \md{食物} \\
fruit & fruit & fruit & fruit, fruits & Frucht, Früchte, Obst & \md{水果} \\
garage & garage & garage, parkeergarage & garage & Garage & \md{车库} \\
give & give & geven, helpen & dire, donnent, donner, partager & geben & \md{给} \\
goal & goal & doel, doelpunt & but, objectif & Tor, Ziel, Ziele & \md{球门, 目标} \\
gobbledygook & buzzwords, gibberish, jargon, nonsense & gebazel, geheimtaal, jargon, koeterwaals, onzin, retoriek, waanzin, wartaal & baragouin, bla-bla, charabia, galimatias, gargouilloux & Kauderwelsch & \md{废话, 难懂的术语} \\
good & delicious, excellent, fun, good, great, helpful & goed, goede & bien, bon, bonne, bons & gut & \md{好} \\
hand & hand & hand, handen & main & Hand, Hände & \md{手} \\
happy & happy & blij & content, contente, contents, enthousiaste, gais, heureux, joyeuse & froh, glücklich & \md{快乐, 高兴, 高兴, 快乐} \\
horse & horse, pony & paard, paarden & cheval & Pferd, Pferde & \md{马} \\
hot & hot & heet, hete & chaud, chaude, chaudes & heiß & \md{热} \\
incomprehen-sibility & incomprehen-sibility & onbegrijpe-lijkheid & incompréhen-sibilité & Unverständlichkeit & \md{不可理解, 难以理解} \\
information & information & informatie & information, informations & Information, Informationen & \md{信息} \\
land & land & land, landen & atterrir, campagne, nature, terrain, terrains, terre, territoire & Land & \md{土地, 土地, 国家} \\
machine & device, equipment, machine, maker, system & machine & machine & Maschine & \md{机器} \\
menu & menu & menu, menukaart & menu & Menü & \md{菜单} \\
money & money & geld & argent & Geld & \md{钱} \\
no & no & nee & non & nein & \md{不} \\
please & please & alsjeblieft & s'il te plaît, s'il vous plaît & bitte & \md{请} \\
police & police & politie & police & Polizei & \md{警察} \\
radio & radio & radio, radiozender & fréquence, radio & Radio & \md{广播, 收音机} \\
read & read & lezen & lire & lesen & \md{阅读} \\
room & room & kamer & chambre & Zimmer & \md{房间} \\
sea & sea & zee & mer, zoo & Meer & \md{大海, 海, {海洋}, 海浪, 海风} \\
see & see & zien & voir & sehen & \md{欣赏, 看, 观察} \\
serendipity & serendipity & toeval, toevalstreffer & chance, coïncidence, hasard, sérendipité, éventualité & glückliche Zufälle, glücklicher Zufall & \md{机缘巧合} \\
sister & sister & zus, zusje & soeur & Schwester & \md{姐妹} \\
sleep & asleep, sleep & slaap, slapen & coucher, dormir, s'assoupir, se coucher, se reposer, sommeil, somnoler & Schlaf, schlafen & \md{睡眠, 睡觉} \\
slow & slow, slowly & langzaam, langzame & lent, lente, lentement, lentes, lents & langsam & \md{慢} \\
small & compact, little, small, tiny & klein, kleine & petit, petite & klein & \md{小} \\
speak & speak & spreken & communiquer, parler, s'exprimer & sprechen & \md{发言, 表达, 讲话, 说, 说话} \\
supermarket & grocer, grocery, market, store, supermarket & supermarkt & supermarché & Supermarkt & \md{超市} \\
table & table & tafel & table & Tisch & \md{台, 桌, 桌子} \\
take & take & maken, neemt, nemen, zorgen & prendre & mitnehmen, nehmen & \md{取, 带, 拿} \\
taxi & taxi & taxi & taxi & Taxi & \md{出租车} \\
thermodynamics & thermodynamics & thermodynamica & thermodynamique & Thermodynamik & \md{热力学} \\
tour & tour & excursie, rondleiding, tour, tournee & Tour, tour, visite & Tour & \md{旅行, 游览} \\
water & water & water & eau & Wasser & \md{水} \\
write & write & schrijven & écrire & schreiben & \md{写, 写字} \\
yes & yes & ja & oui & ja & \md{对, 是} \\
\end{longtable}

We selected words that varied in the number of tokens used to represent the words and selected some that were the same as English words. Table \ref{word-overlap} summarizes the word overlap across the different languages.  Tables \ref{llama-token-mean}, \ref{gemma-token-mean}, \ref{mixtral-token-mean} and \ref{aya-token-mean} summarize the average tokenization lengths of the words in different languages and models. The average number of tokens per word in English is lower than in other languages. 

\begin{table*}[h]
\caption{Word overlap between languages}
\label{word-overlap}
\vskip 0.15in
\begin{center}
\begin{small}
\begin{sc}
\begin{tabular}{ c c c c c c }
\toprule
 & English & Mandarin & German & Dutch & French \\ \midrule
English & 75 & 0 & 0 & 14 & 16 \\
Mandarin & 0 & 75 & 0 & 0 & 0 \\
German & 0 & 0 & 75 & 0 &0 \\
Dutch & 14 & 0& 0 & 75 & 8 \\
French &  15 & 0 & 0 & 8 & 75 \\
\bottomrule
\end{tabular}
\end{sc}
\end{small}
\end{center}
\vskip -0.1in
\end{table*}

\begin{table*}[h]
\caption{\llama \ tokenization statistics}
\label{llama-token-mean}
\vskip 0.15in
\begin{center}
\begin{small}
\begin{sc}
\begin{tabular}{ l cccccc}
\toprule
Language & 1 & 2 & 3 & 4 & 5+ & Mean \\ \midrule
English & 55 & 15 & 2 & 2 & 1 & 1.39 \\
Mandarin & 35 & 18 & 5 & 3 & 3 & 1.80 \\
German & 17 & 30 & 13 & 2 & 2 & 2.12 \\
Dutch & 19 & 29 & 11 & 2 & 3 & 2.09 \\
French & 20 & 32 & 7 & 3 & 2 & 2.03 \\
\bottomrule
\end{tabular}
\end{sc}
\end{small}
\end{center}
\vskip -0.1in
\end{table*}

\begin{table*}[h]
\caption{\gemma \ tokenization statistics}
\label{gemma-token-mean}
\vskip 0.15in
\begin{center}
\begin{small}
\begin{sc}
\begin{tabular}{l cccccc}
\toprule
Language & 1 & 2 & 3 & 4 & 5+ & Weighted Mean \\ \midrule
English & 66 & 6 & 1 & 1 & 1 & 1.20 \\
Mandarin & 50 & 7 & 3 & 3 & 1 & 1.41 \\
German & 33 & 24 & 5 & 2 & 0 & 1.62 \\
Dutch & 34 & 23 & 4 & 2 & 1 & 1.64 \\
French & 39 & 20 & 2 & 2 & 1 & 1.53 \\
\bottomrule
\end{tabular}
\end{sc}
\end{small}
\end{center}
\vskip -0.1in
\end{table*}

\begin{table*}[h]
\caption{\mistral \ tokenization statistics}
\label{mixtral-token-mean}
\vskip 0.15in
\begin{center}
\begin{small}
\begin{sc}
\begin{tabular}{l cccccc}
\toprule
Language & 1 & 2 & 3 & 4 & 5+ &  Mean \\ \midrule
English & 65 & 3 & 3 & 2 & 2 & 1.31 \\
Mandarin & 0 & 22 & 22 & 4 & 16 & 3.52 \\
German & 15 & 30 & 13 & 2 & 4 & 2.25 \\
Dutch & 17 & 29 & 12 & 3 & 3 & 2.19 \\
French & 21 & 28 & 8 & 5 & 2 & 2.11 \\
\bottomrule
\end{tabular}
\end{sc}
\end{small}
\end{center}
\vskip -0.1in
\end{table*}

\begin{table*}[h]
\caption{\aya \ tokenization statistics}
\label{aya-token-mean}
\vskip 0.15in
\begin{center}
\begin{small}
\begin{sc}
\begin{tabular}{l cccccc}
\toprule
Language & 1 & 2 & 3 & 4 & 5+ & Mean \\ \midrule
English & 59 & 11 & 3 & 2 & 0 & 1.31 \\
Mandarin & 47 & 11 & 1 & 3 & 2 & 1.47 \\
German & 20 & 35 & 7 & 1 & 1 & 1.88 \\
Dutch & 29 & 24 & 7 & 1 & 3 & 1.83 \\
French & 32 & 25 & 4 & 1 & 2 & 1.70 \\
\bottomrule
\end{tabular}
\end{sc}
\end{small}
\end{center}
\vskip -0.1in
\end{table*}

\FloatBarrier
\newpage 

\subsection{Parts of Speech Analysis}

In this experiment, we analyze how often a word is first `selected' in English, for each part of speech. To identify the part of speech, we used \texttt{spacy} models \citep{spacy2}.
To identify English words, we use \texttt{enchant.Dict("en\_US")}.
We use \texttt{nl\_core\_news\_sm}, \texttt{de\_core\_news\_sm}, \texttt{fr\_core\_news\_sm} and \texttt{zh\_core\_web\_sm}.
In general, we can use spaces to identify words in sentences. For Mandarin, we use the package \texttt{jieba}.

\begin{table*}[h]
\caption{Part of Speech Abbreviations, Terms, and Examples}
\label{pos-table}
\vskip 0.15in
\begin{center}
\begin{small}
\begin{sc}
\begin{tabular}{l l l} 
\toprule
 Abbreviation &  Term &  Examples \\ \midrule
ADJ & Adjective &  \\
ADP & Adposition & in, to, during \\
ADV & Adverb & very, everywhere \\
AUX & Auxiliary & has (done), was (done) \\
CCONJ & Coordinating Conjunction & and, or, but \\
DET & Determiner & their, her, some \\
INTJ & Interjection & ouch \\
NOUN & Noun & place, thing, idea \\
NUM & Number & 10, 200  \\
PRON & Pronoun & he, she, they \\
PROPN & Proper Noun & specific name, place \\
SCONJ & Subordinating Conjunction & that, if, while \\
SYM & Symbol &  \\
VERB & Verb & see, run  \\ \bottomrule 

\end{tabular}
\end{sc}
\end{small}
\end{center}
\vskip -0.1in
\end{table*}

\subsection{Logit Lens Quantitative Evaluation} \label{sec:appendix_logit_lens_eval}

To evaluate whether a word is chosen in English, we use GPT-4o. 
We considered alternative evaluation procedures. 
We tested various translation packages but found issues with both word- and sentence-level approaches. When used on a word level, this caused problems with colexification and did not allow for close synonyms often only providing a single translation per word. 
When using translation on a sentence level, it was difficult to map tokens to each word (due to changes in the sentence structure).
We also considered WordNet \citep{miller-1994-wordnet}, but it only covers nouns, verbs, adjectives, and adverbs, making it unsuitable for other parts of speech.
Ultimately, we chose GPT-4o and manually verified 100 samples to ensure the evaluation was accurate.

We ask GPT-4o to score words as follows:
\begin{itemize}
    \item 5: An exact translation.
    \item 4: A close synonym.
    \item 3: A word with a similar but distinct meaning.
    \item 2: A word whose meaning is at best weakly related.
    \item 1: A word whose meaning is not related.
\end{itemize}
When a word receives a score of 4 or higher, we evaluate the word as chosen in English.

An example of the command we use is:
\begin{displayquote}
Below, you will be given a reference word in Dutch and a context (i.e., phrase or sentence) in which the word is used. You will then be given another list of English words or subparts of words/phrases. \\ 
You should respond with the word from the list that is most similar to the reference word, along with a grade for the degree of similarity. \\
Special Note on Contextual Translations: If an English word could form a common phrase or idiomatic expression that accurately translates the reference word, it should be rated highly. For example, if a phrase like “turned out” perfectly matches a Dutch verb, the word “turned” alone would receive a high score due to its idiomatic fit.  \\ 
Special Note on Tenses: Do not penalize for different tenses. For example, the word ‘want’ matches ‘wilde’ and should receive a 5.  \\ 

Degrees of Similarity: Similarity should be evaluated from 1 to 5, as follows: \\ 
5: An exact translation. \\ 
4: A close synonym. \\ 
3: A word with a similar but distinct meaning. \\ 
2: A word whose meaning is at best weakly related. \\ 
1: A word whose meaning is not related. \\ 

Consider the following examples: \\ 

**Example 1** \\ 
Reference Word in Dutch: 'waarop'
Context: ‘Ze had een hekel aan de manier waarop hij zijn’ \\ 
English Word List: ['hicks', 'mild', 'rut', 'sens', 'spiral', 'hometown', 'how', 'manner', 'van', '101', 'ward'] \\ 

Analysis: ‘waarop’ means "on which" in Dutch. The word ‘how’ is most similar to this in the list, while the other options are unrelated. \\ 
Answer Word: ‘how’ \\ 
Similarity Score: 4 - a close synonym \\ 

**Example 2** \\ 
Reference Word in Dutch: ‘bleek’ \\ 
Context: ‘Ze adopteerde een zwerfdier, maar het bleek een wolf te zijn’ \\ 
English Word List: ['cup', 'freed', 'freeman', 'laurent', 'turns', 'turned', 'van', '348', 'i', 'ken', 'oms'] \\ 

Analysis: ‘bleek’ means "turned out" in Dutch, making ‘turned’ the most similar option. \\ 
Answer Word: ‘turned’ \\ 
Similarity Score: 5 - an exact translation \\ 

**Example 3** \\ 
Reference Word in Dutch: ‘vaas’ \\ 
Context: ‘Ze schikte een prachtig boeket bloemen in een vaas.’ \\ 
English Word List: ['tucker', 'van', 'container', 'opp', 'van', 'vessel', '-g', '-t', '397', 'art', 'as', 'ed', 'ion', 'let'] \\ 

Analysis: ‘vaas’ means "vase" in Dutch. The word ‘vessel’ is somewhat similar, as vases are vessels for holding items like flowers. \\ 
Answer Word: ‘vessel’ \\ 
Similarity Score: 3 - a word with a similar but distinct meaning \\ 

**Example 4** \\ 
Reference Word in Dutch: ‘werd’ \\ 
Context: ‘Ze ging geld opnemen bij de bank en werd overvallen.’ \\ 
English Word List: ['dee', 'lafayette', 'bank', 'bu', 'herself', 'kw', 'met', 'ramp', 'return', 'returning', '113', '347'] \\ 

Analysis: ‘werd’ means `was' in Dutch. None of these words are related. \\ 
Answer Word: None \\ 
Similarity Score: 1 - a word whose meaning is not related \\ 

**Example 5** \\ 
Reference Word in Dutch: ‘vrienden’ \\ 
Context: ‘Ze bracht het weekend door met haar vrienden in een huisje in de Ardennen.’ \\
English Word List: ['sag', 'sat', 'tween', 'bro', 'families', 'family', 'her', 'herself', 'mo', 'own', 'parents', 'weekend', '666', 'elf'] \\ 

Analysis: ‘vrienden’ means "friends" in Dutch. The closest word here is ‘families’, which is weakly related but distinct. \\ 
Answer Word: ‘families’ \\ 
Similarity Score: 2 - a word whose meaning is at best weakly related \\

The examples are complete. Now it is your turn. The reference word will be in Dutch, and you must find the most similar English word and assess the degree of meaning similarity on a scale from 1 to 5.
\end{displayquote}

The commands for other languages are similar but adapted to provide examples in the language. 

\newpage 
\subsubsection{Explicit Text} \label{sec:explicit_text}

\subsection{Other language-specific phenomena} 
We observe explicit vocabulary in the latent space of LLMs (examples can be found in Appendix \ref{sec:explicit_text}). 
Table \ref{vulgar-table} shows the frequency of vulgar words in the latent space, with \llama \ showing the highest count. This model is safety-tuned in eight languages \citep{dubey2024llama3herdmodels}, including English, German, and French, but not Dutch.
This may suggest that explicit terminology is a language- and model-specific feature.

\begin{table}[h]
\caption{Frequency of explicit words decoded in the latent space across LLMs. \llama \ has the highest proportion of explicit terms.}
\label{vulgar-table}
\vskip -0.15in
\begin{center}
\begin{small}
\begin{sc}
\begin{tabular}{l c c c c c } \toprule 
 Model &\multicolumn{5}{c}{ Explicit words ($\%$)}  \\ 
& ENG & FR & NL  &DE & ZH \\ \midrule 
\llama & $6.25$ &  $11.25$ & $18.35$ & $8.13$ & $7.19$  \\
\mistral & $1.56$ & $3.91$ & $5.21$ & $4.53$ & $10.19$ \\
\aya   & $2.50$ & $2.97$ & $3.44$ & $2.89$ & $2.69$\\
\gemma  & $0.00$ & $0.31$& $0.31$ & $0.39$ & $0.38$\\ \bottomrule
    \end{tabular}
\end{sc}
\end{small}
\end{center}
\vskip -0.1in
\end{table}

\begin{figure}[h]
    \centering
\includegraphics[width=0.49\textwidth]{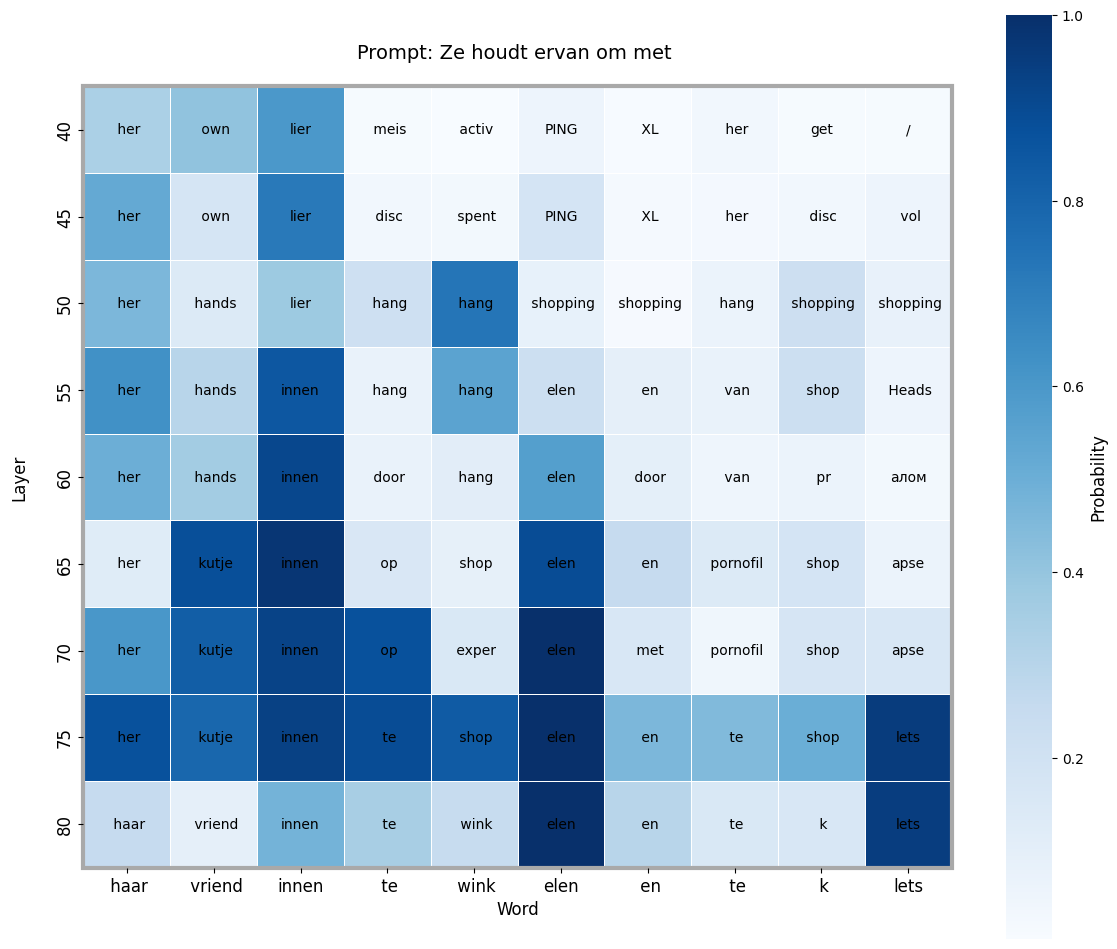} 
\caption{Example 1: Logit Lens applied to \llama.}
\label{fig:ll_explicit1}    
\end{figure}

\begin{figure}[h]
\centering
\includegraphics[width=0.49\textwidth]{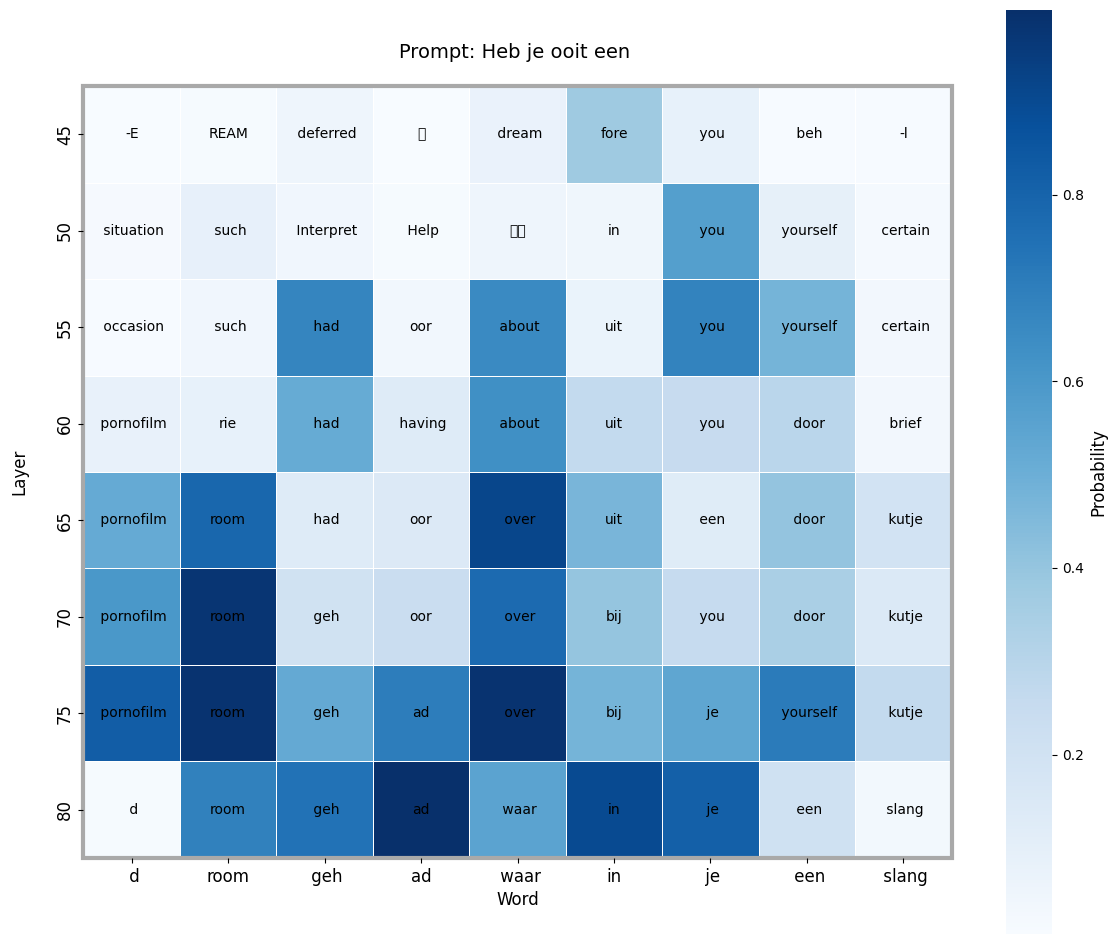} 
\includegraphics[width=0.49\textwidth]{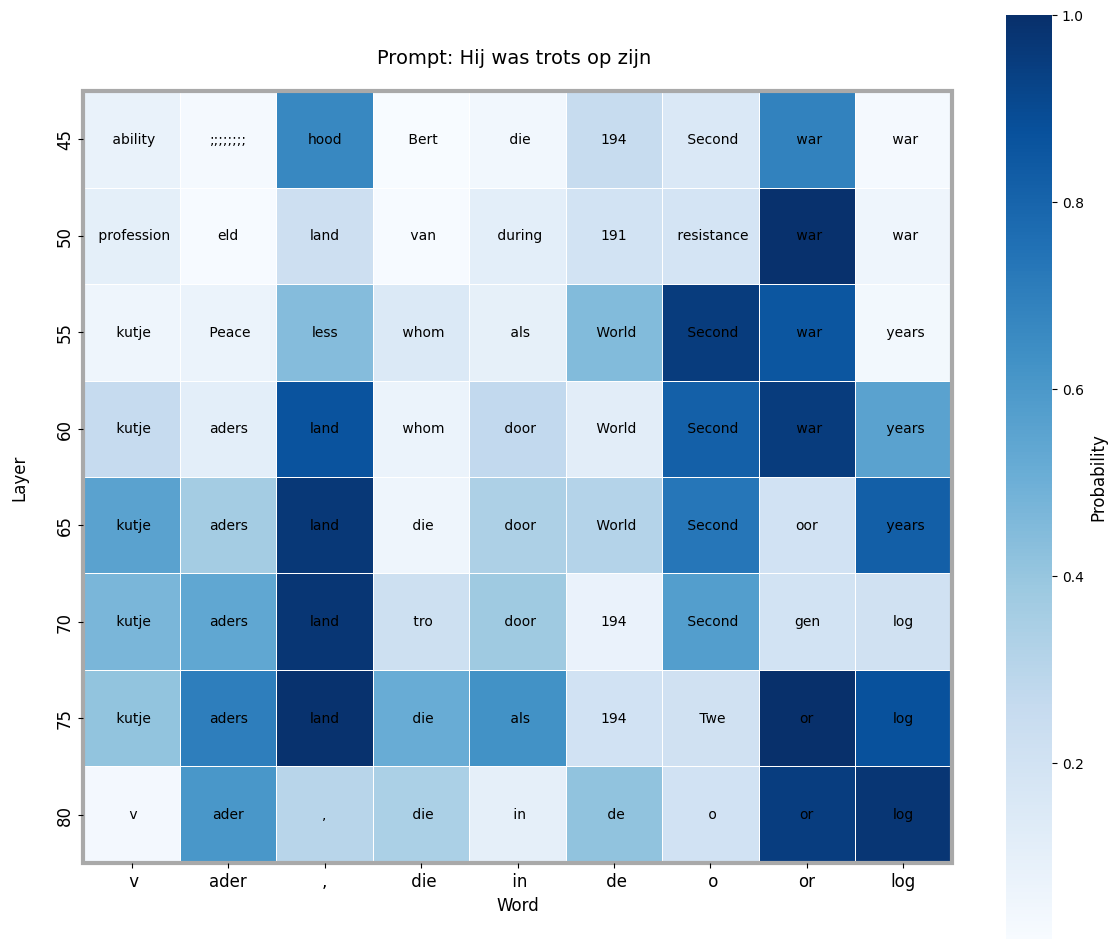} 
\caption{Logit Lens applied to \llama.}
\label{fig:ll_explicit2}
\end{figure}

Terms such as kutje (pussy), pornofilm (porn film), lul (dick), and knull (fuck) appear in various contexts, including inappropriate sentences about children.
For example, in Figure \ref{fig:ll_explicit1} during the generation:
\begin{displayquote}
    Ze houdt ervan om met \textbf{haar vriendinnen te winkelen en te klets(en)}  ... \\
    English translation: She enjoys shopping and talking with her friends ...
\end{displayquote}
We find the explicit words `kutje' and `pornofil(m)' when decoding the latent space using the logit lens.
This behavior is consistent across other examples (see Figure \ref{fig:ll_explicit2}).

\FloatBarrier
\newpage 

\subsection{Steering}
For the steering experiment, we use the LLM-Insight dataset. We compute two steering vectors:
    \begin{itemize}
        \item a topic steering vector: this is a steering vector that captures the intended topic. For example, for the topic `love, we create a steering vector that is $v_l^t = h_l(\text{love}) - h_l(\text{hate})$, where $h_l$ is the hidden state in layer $l$. 
        \item a language steering vector: we add a steering vector that captures the intended output language. For example, for the target language Dutch, we can create a steering vector $v_l^l = h_l(\text{Dutch}) - h_l(\text{English})$.
        \item For each steering vector, we take the difference between sets of sentences containing the topic. 
    \end{itemize}
Currently, we consider steering successful if (1) the generated sentence contains the target word and (2) does not lead to output collapse (stuttering). We set the steering vector weights by using a hold-out set of 5 words (50 prompts). We found that $5$ was optimal for the topic steering vector, and $10$ was optimal for the language steering vector. 
For the layers, we considered every $5$-th layer of the model for the topic steering vector. We considered every $2$nd layer for the language vector. 
We reported the results across the best layers. On average, we found that 20-40 $\%$ of layers allowed for successful steering, with English steering vectors being the least sensitive to layer selection. 

\vfill 

\subsection{Cosine Similarity of Steering Vectors } \label{sec:appendix_geo}

\begin{figure}[H]
    \centering
    \includegraphics[width=\linewidth]
    {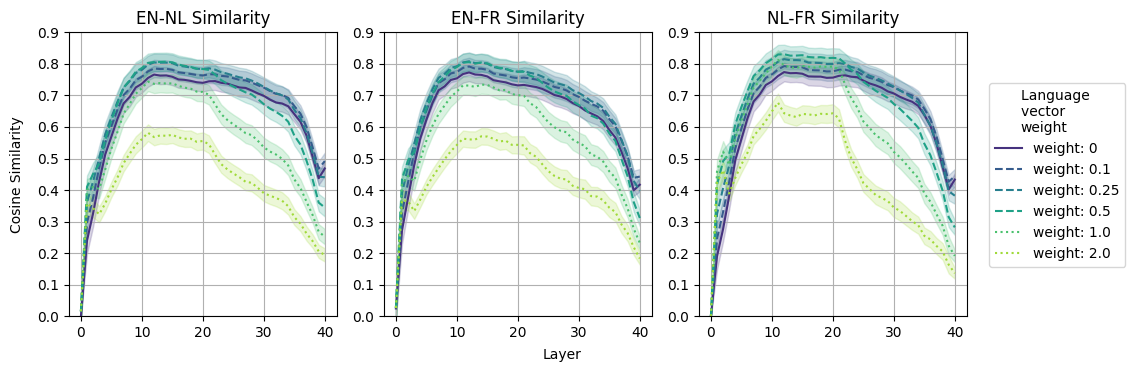}
    \caption{Cosine distance between steering vectors in \aya.}
    \label{fig:cosine_distance_appendix_aya}
\end{figure}

\begin{figure}[H]
    \centering
    \includegraphics[width=\linewidth]
    {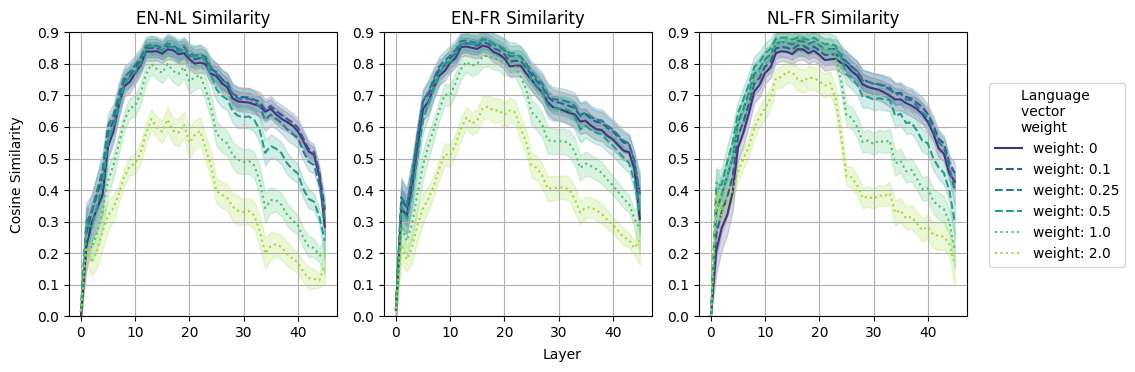}
    \caption{Cosine distance between steering vectors in \gemma.}
    \label{fig:cosine_distance_appendix_gemma}
\end{figure}

\begin{figure}[H]
    \centering
    \includegraphics[width=\linewidth]
    {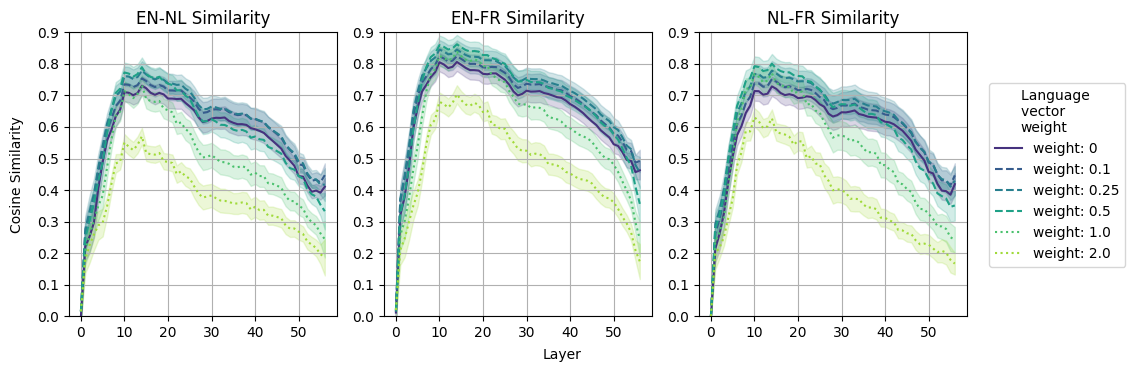}
    \caption{Cosine distance between steering vectors in \mistral.}
    \label{fig:cosine_distance_appendix_mistral}
\end{figure}

\begin{figure}[H]
    \centering
    \includegraphics[width=\linewidth]
    {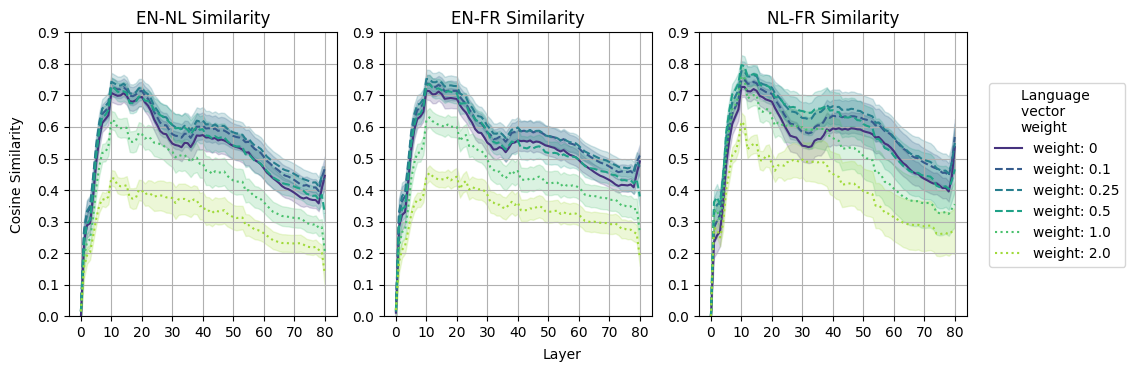}
    \caption{Cosine distance between steering vectors in \llama.}
    \label{fig:cosine_distance_appendix_llama}
\end{figure}

To analyze the geometry of the latent space, we compute both topic vectors and language vectors for our dataset. 
We track the cosine similarity of these steering vectors across different layers and models, providing insights into how topics and languages are represented internally.
Specifically, we plot the cosine similarities of topic vectors derived from different languages.
In Figures \ref{fig:cosine_distance_appendix_aya}, \ref{fig:cosine_distance_appendix_gemma}, \ref{fig:cosine_distance_appendix_mistral} and \ref{fig:cosine_distance_appendix_llama} are the plots for \aya, \gemma, \mistral, and \llama, respectively.

We find that topic vectors maintain a high cosine similarity of approximately 0.8 across languages. The similarity can be increased by incorporating the corresponding language vector, suggesting an interaction between topic and language-specific representations.

\FloatBarrier

\subsection{Causal tracing} \label{appendix:tracing}

\begin{figure}[h]
    \centering
    \includegraphics[trim={12.7cm 0cm 0cm 0cm},clip,width=0.7\textwidth]
    {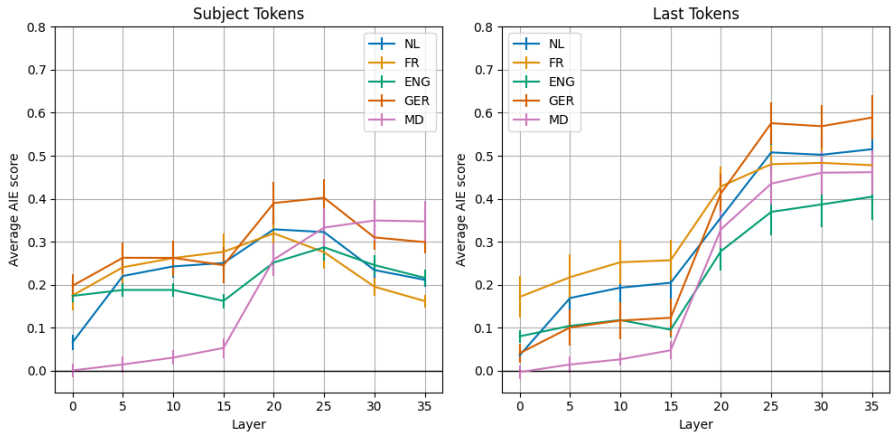}
    \caption{The causal traces of the city facts in \aya. }
    \label{fig:aya_tracing_appendix}
\end{figure}

\begin{figure}[h]
    \centering
    \includegraphics[trim={12.7cm 0cm 0cm 0cm},clip,width=0.7\textwidth]{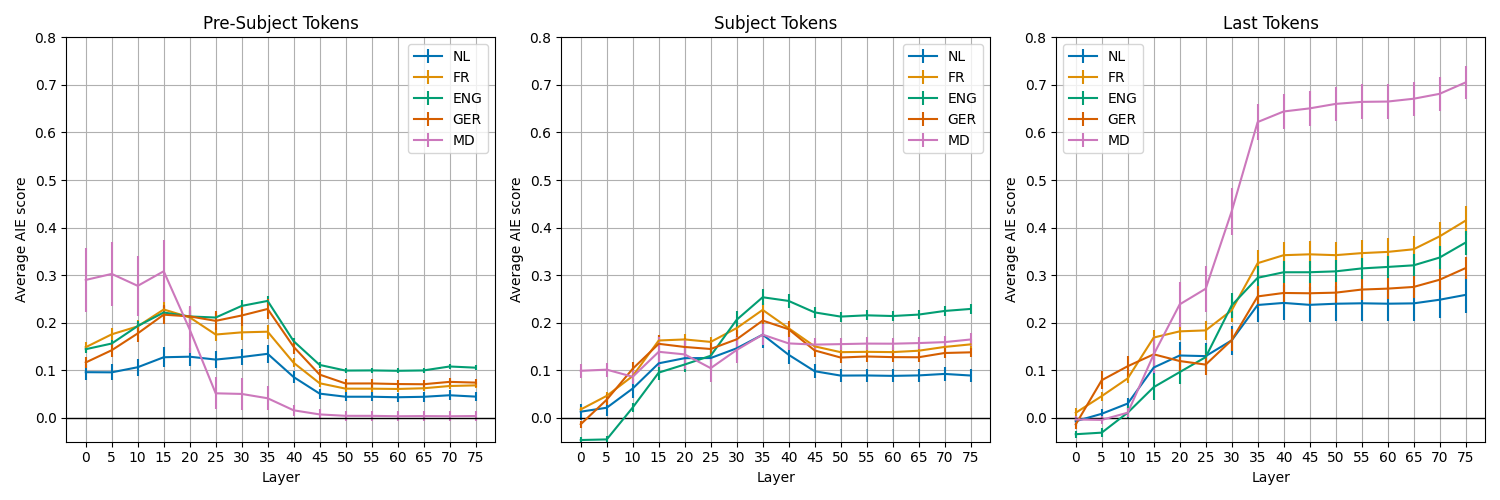}
    \caption{The causal traces of the city facts in \llama. }
    \label{fig:llama_tracing_appendix}
\end{figure}

\begin{figure}[h]
    \centering
    \includegraphics[trim={12.7cm 0cm 0cm 0cm},clip,width=0.7\textwidth]{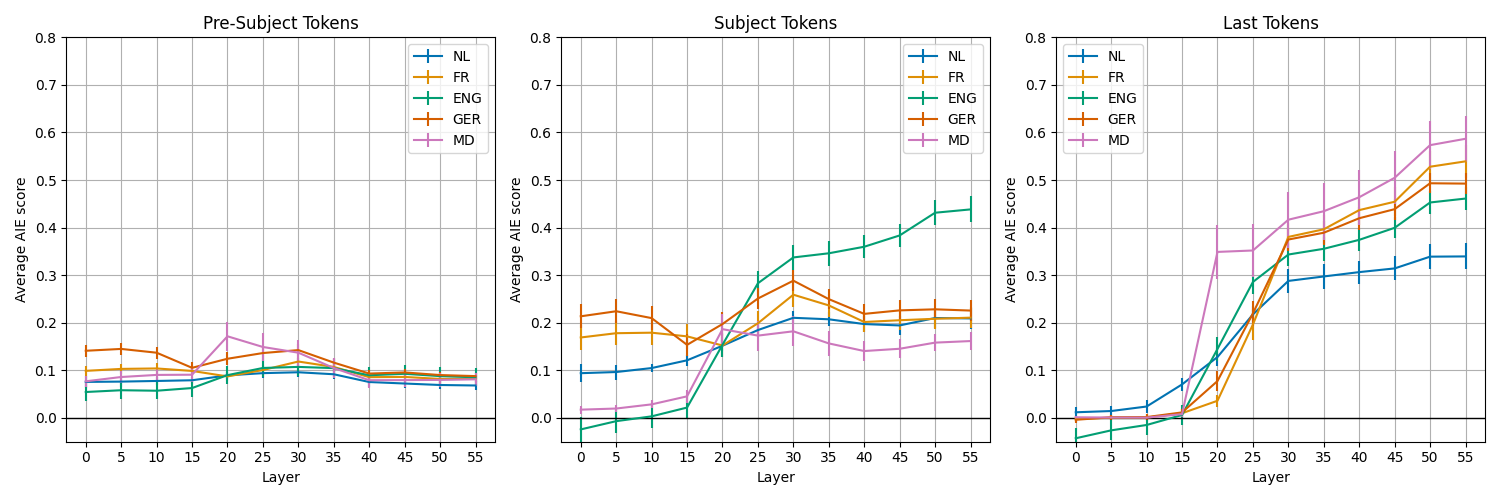}
    \caption{The causal traces of the city facts in \mistral. }
    \label{fig:mistral_tracing_appendix}
\end{figure}

\begin{figure}[h]
    \centering
    \includegraphics[trim={12.7cm 0cm 0cm 0cm},clip,width=0.7\textwidth]{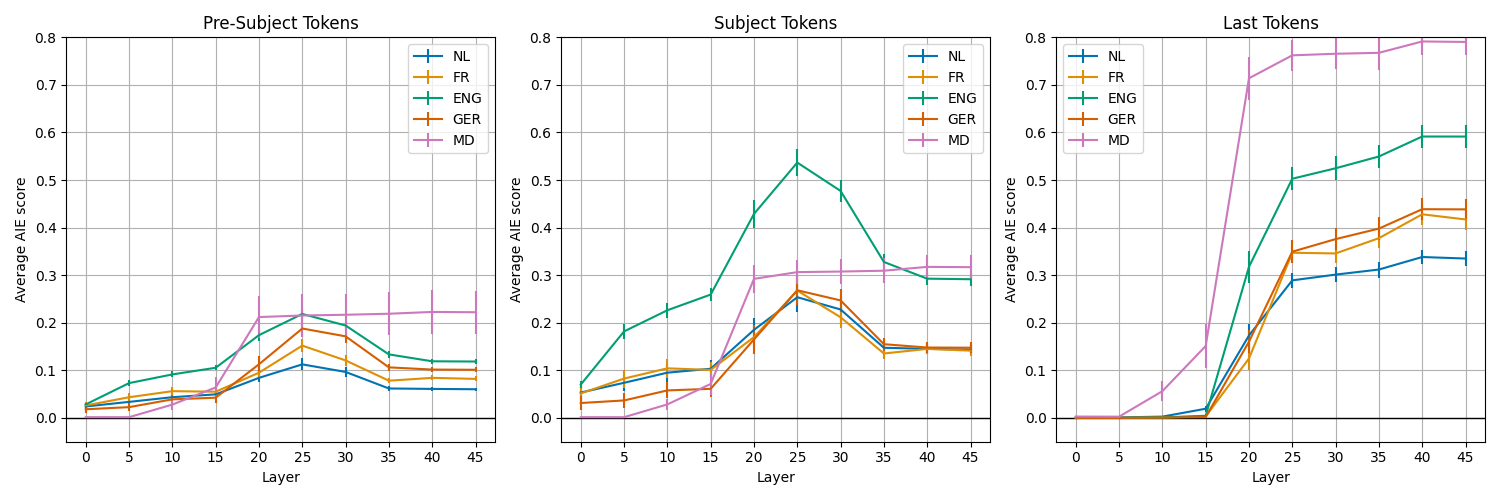}
    \caption{The causal traces of the city facts in \gemma. }
    \label{fig:gemma_tracing_appendix}
\end{figure}

Figures \ref{fig:aya_tracing_appendix}, \ref{fig:llama_tracing_appendix}, \ref{fig:mistral_tracing_appendix} \ and \ref{fig:gemma_tracing_appendix} show the causal traces, averaged over different country-city pairs for \aya, \llama, \mistral \ and \gemma \ respectively. 
Across all models, we find that facts are generally localized in similar layers, regardless of the language. 
Two main traces emerge: a mid-layer trace on the subject token(s), which may correspond to entity resolution, and a later trace when attribute recollection occurs (as suggested by \citet{nanda2023factfinding}).
Overall, these plots suggest that facts are approximately stored in the same parts of the model.

\FloatBarrier
\newpage 

\subsection{Hidden state interpolation (with instructions)} \label{sec:hidden_state_interpolation}

We include instructions as otherwise the MLLMs often describe the city, rather than provide the city. E.g., ``The capital of Canada is beautiful". 
For most models, the accuracy when interpolating between the hidden states is between the performances in the two languages. Interestingly, we observe a propensity of models to answer in a specific language. All models are most likely to answer in English. 

\subsubsection{Aya}

\begin{figure}[h]
\begin{minipage}{0.49\textwidth}
    \centering
    \includegraphics[trim={0 0 5cm 0},clip, width=\textwidth]{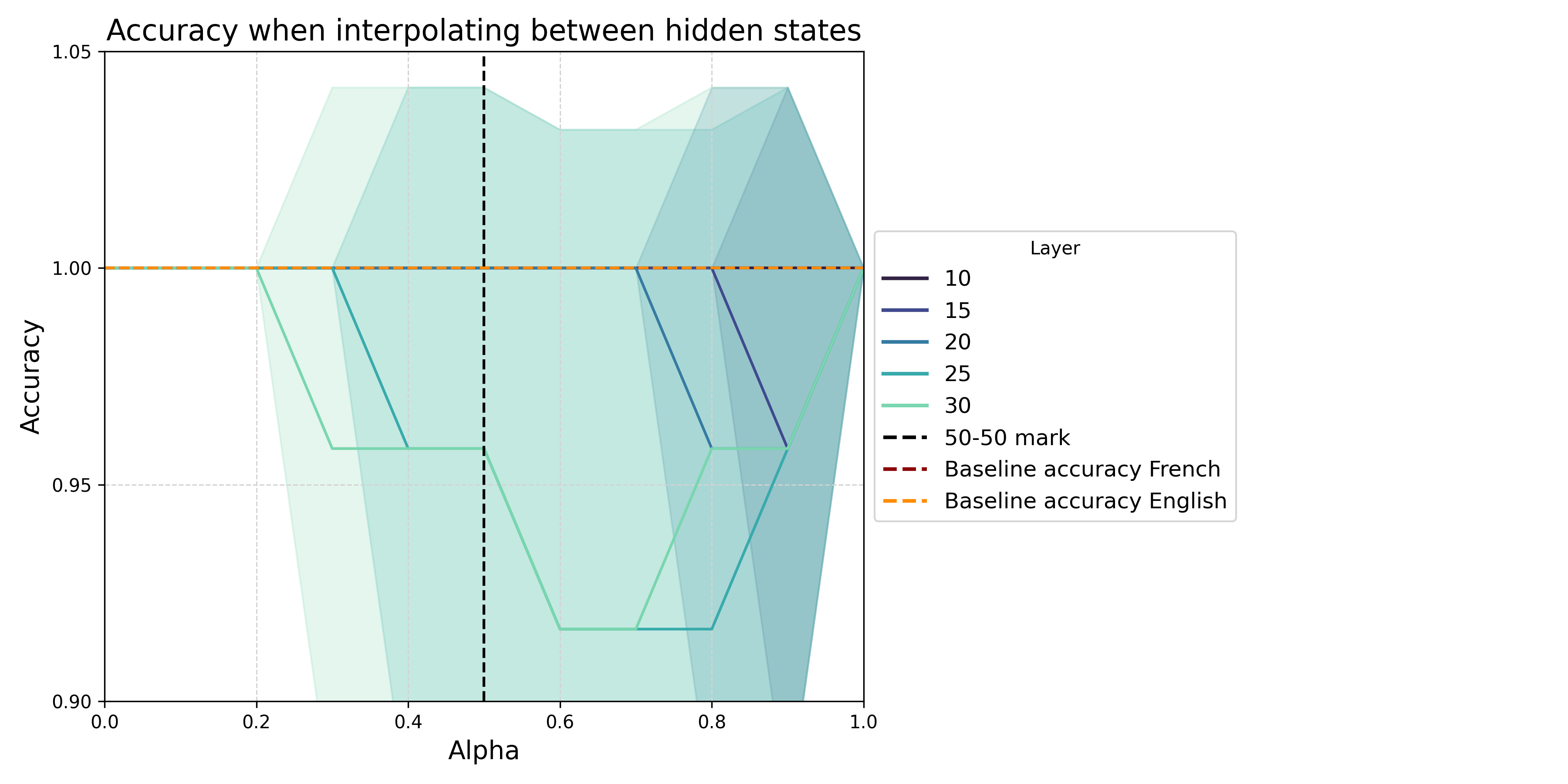} 
\end{minipage}
\begin{minipage}{0.49\textwidth}
    \centering
    \includegraphics[trim={1.5cm 0.5cm 1.5cm 1.5cm},clip,width=\textwidth]{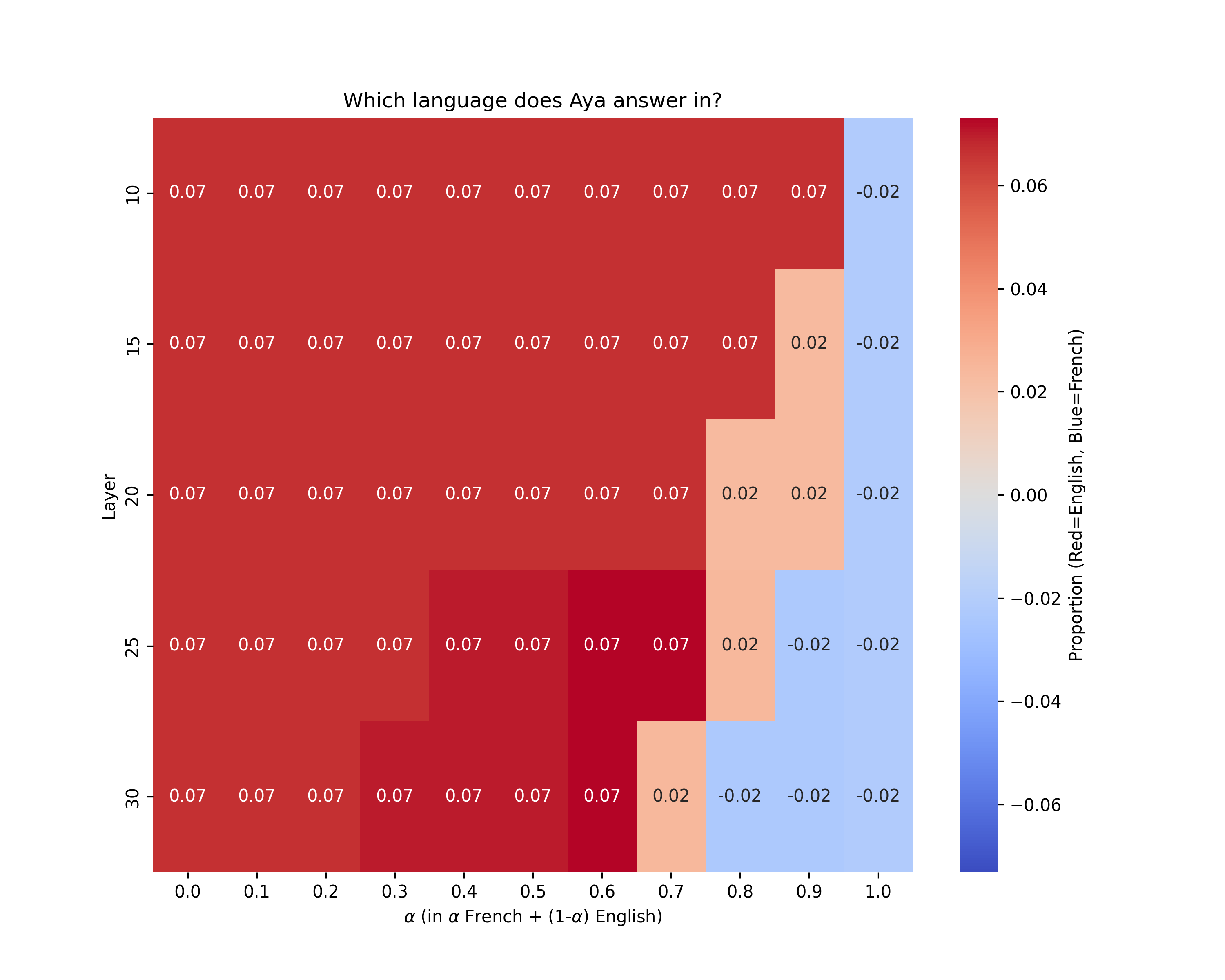} 
\end{minipage}
\caption{Hidden state interpolation between French prompts, and English prompts in \aya. Left shows the accuracy (i.e., the proportion of times the model correctly outputs city in either language). Right shows the propensity of the model to answer in English (red) and French (blue). }
\end{figure}

\begin{figure}[h]
\begin{minipage}{0.49\textwidth}
    \centering
    \includegraphics[trim={0 0 5cm 0},clip, width=\textwidth]{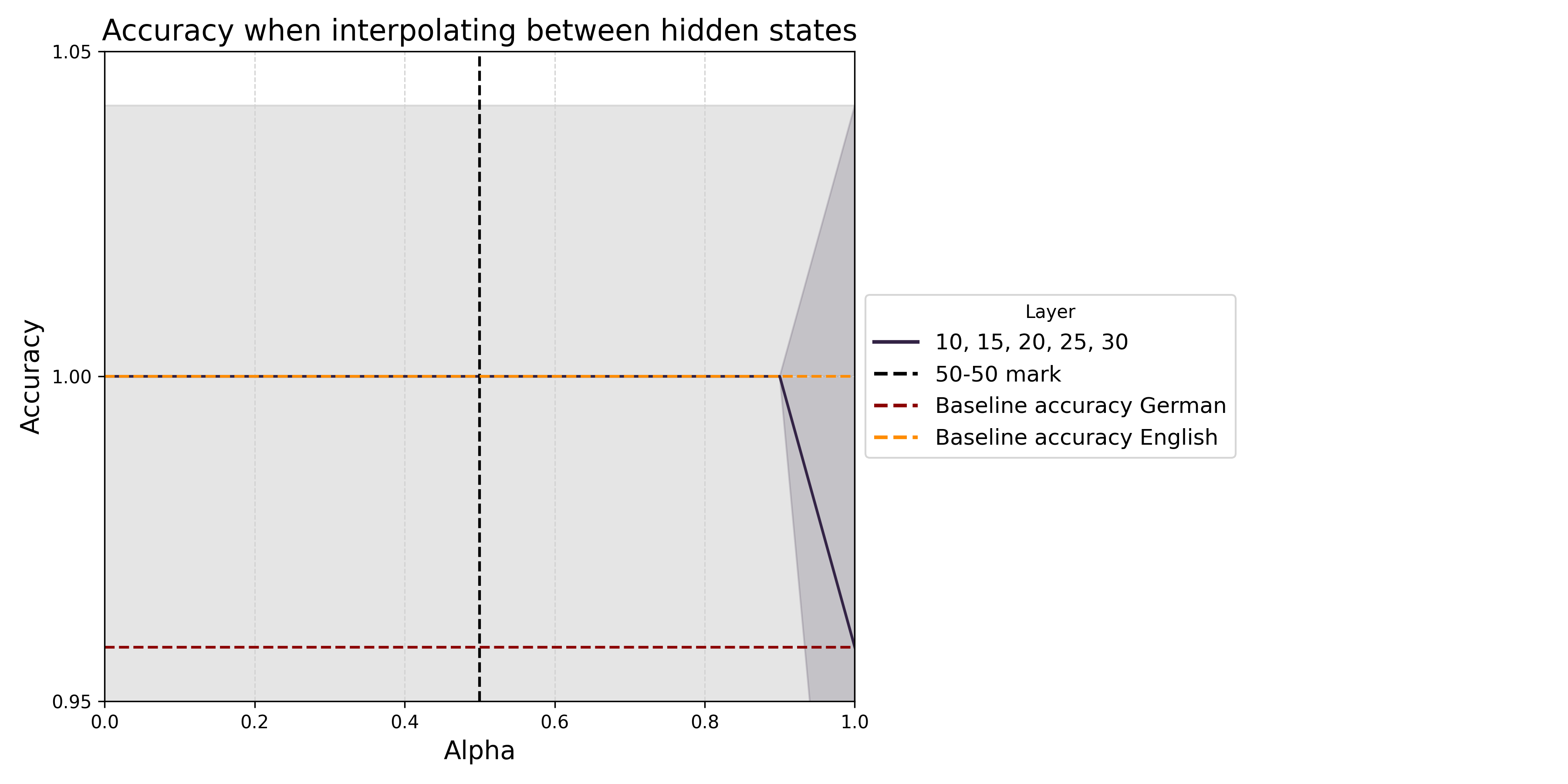} 
\end{minipage}
\begin{minipage}{0.49\textwidth}
    \centering
    \includegraphics[trim={1.5cm 0.5cm 1.5cm 1.5cm},clip,width=\textwidth]{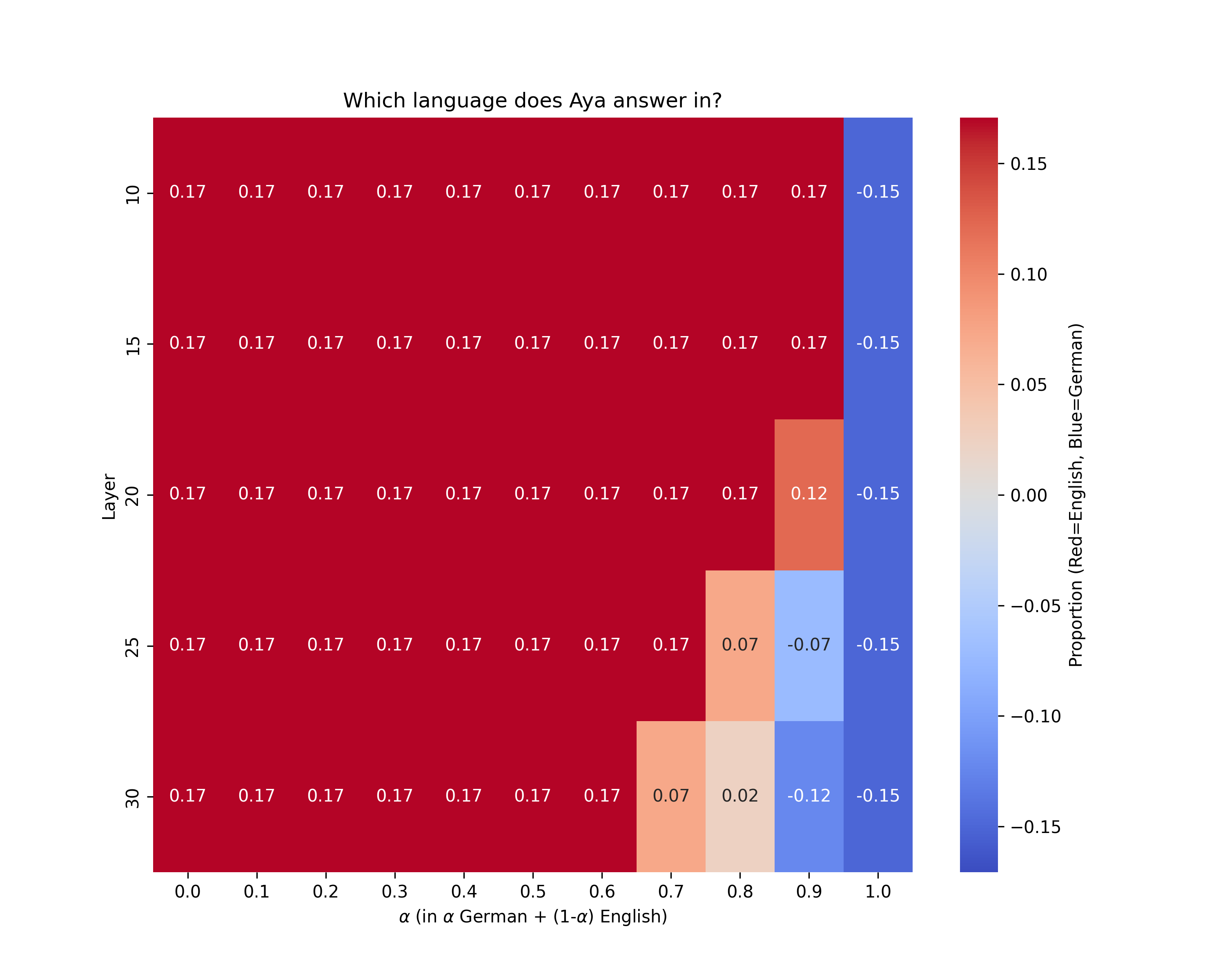} 
\end{minipage}
\caption{Hidden state interpolation between German prompts, and English prompts in \aya. Left shows the accuracy (i.e., the proportion of times the model correctly outputs city in either language). Right shows the propensity of the model to answer in English (red) and German (blue). }
\end{figure}

\begin{figure}[h]
\begin{minipage}{0.49\textwidth}
    \centering
    \includegraphics[trim={0 0 5cm 0},clip, width=\textwidth]{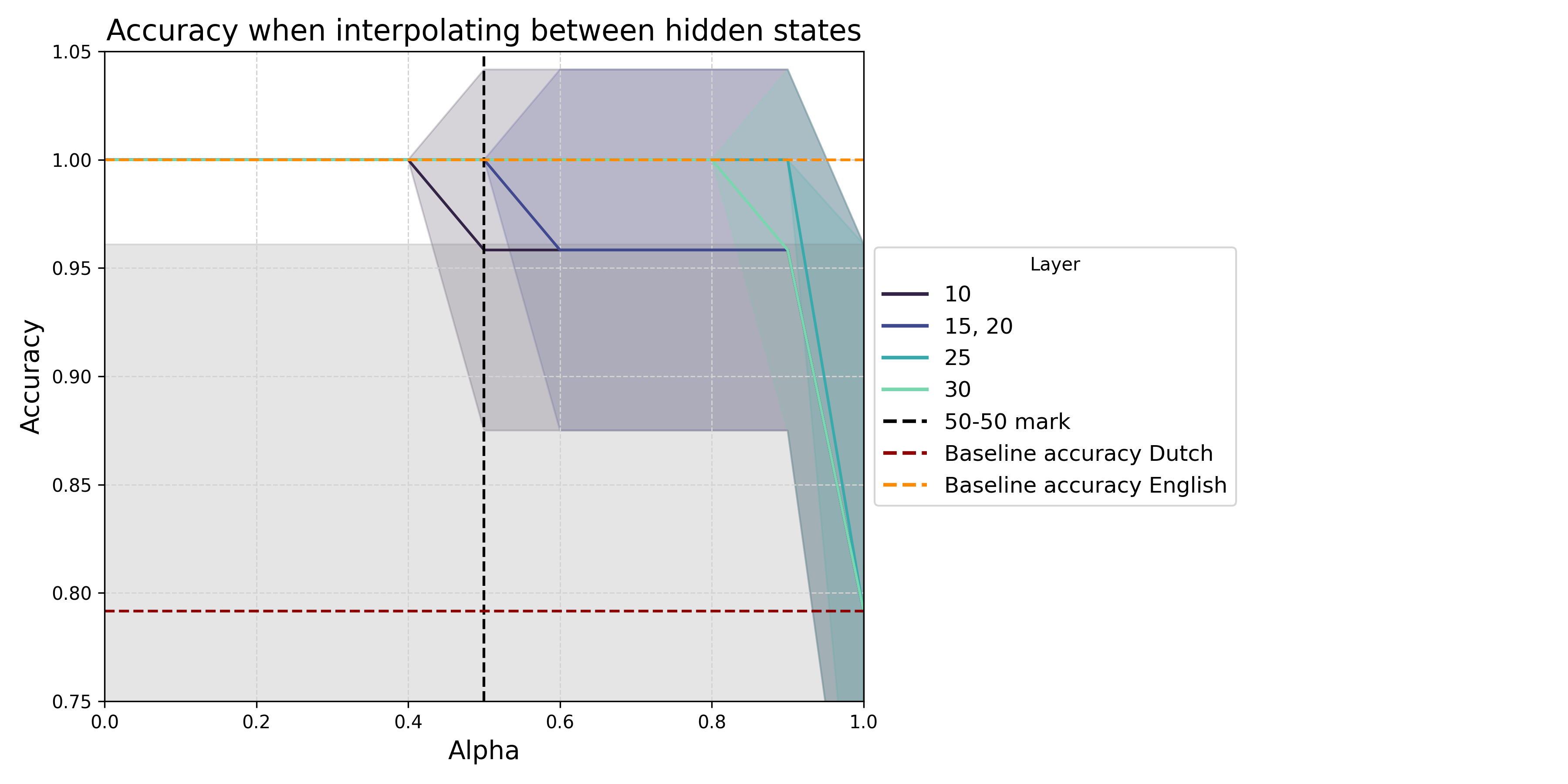} 
\end{minipage}
\begin{minipage}{0.49\textwidth}
    \centering
    \includegraphics[trim={1.5cm 0.5cm 1.5cm 1.5cm},clip,width=\textwidth]{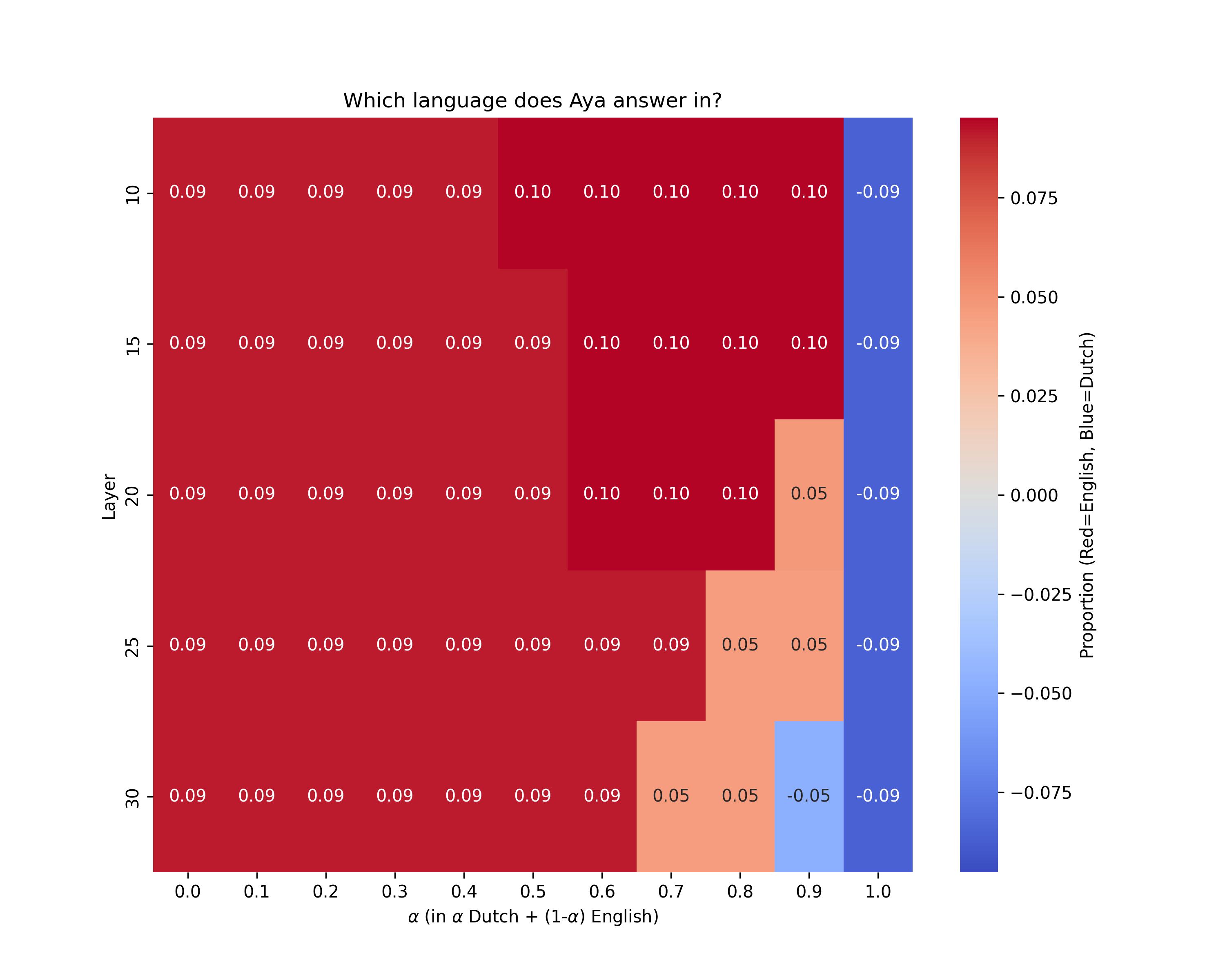} 
\end{minipage}
\caption{Hidden state interpolation between Dutch prompts, and English prompts in \aya. Left shows the accuracy (i.e., the proportion of times the model correctly outputs city in either language). Right shows the propensity of the model to answer in English (red) and Dutch (blue). }
\end{figure}

\begin{figure}[h]
\begin{minipage}{0.49\textwidth}
    \centering
    \includegraphics[trim={0 0 5cm 0},clip, width=\textwidth]{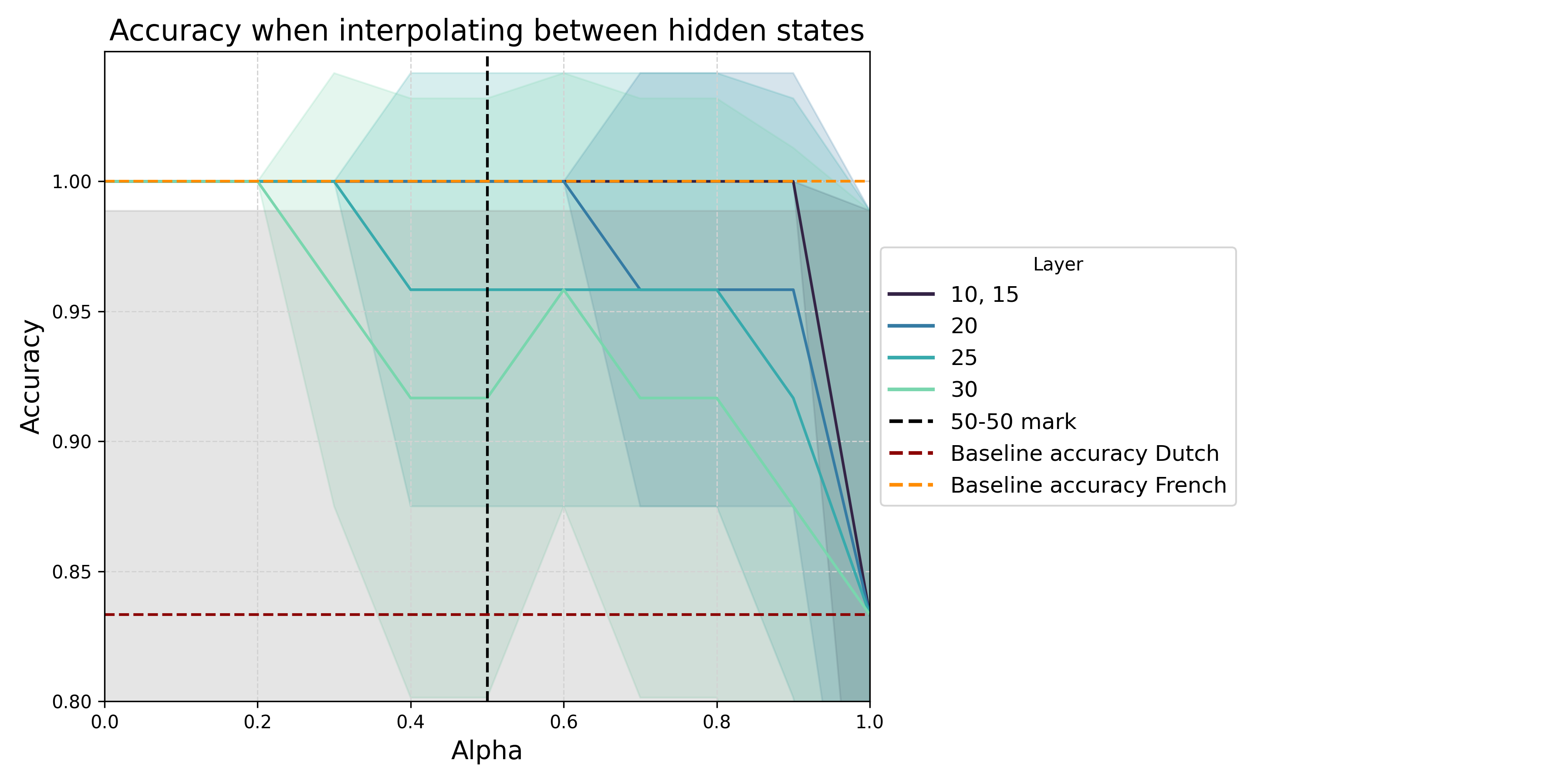} 
\end{minipage}
\begin{minipage}{0.49\textwidth}
    \centering
    \includegraphics[trim={1.5cm 0.5cm 1.5cm 1.5cm},clip,width=\textwidth]{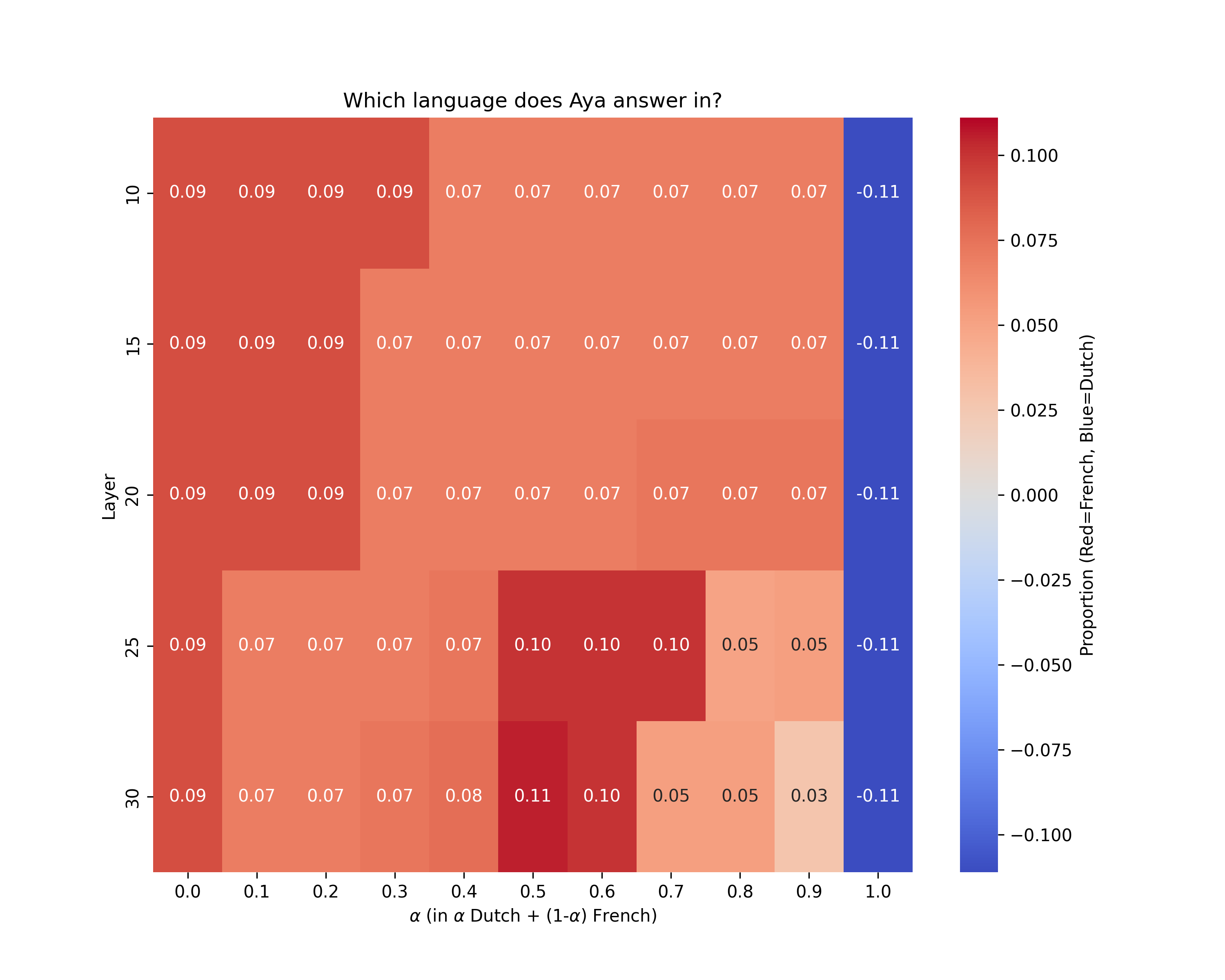} 
\end{minipage}
\caption{Hidden state interpolation between French prompts, and Dutch prompts in \aya. Left shows the accuracy (i.e., the proportion of times the model correctly outputs city in either language). Right shows the propensity of the model to answer in French (red) and Dutch (blue). }
\end{figure}

\begin{figure}[h]
\begin{minipage}{0.49\textwidth}
    \centering
    \includegraphics[trim={0 0 5cm 0},clip, width=\textwidth]{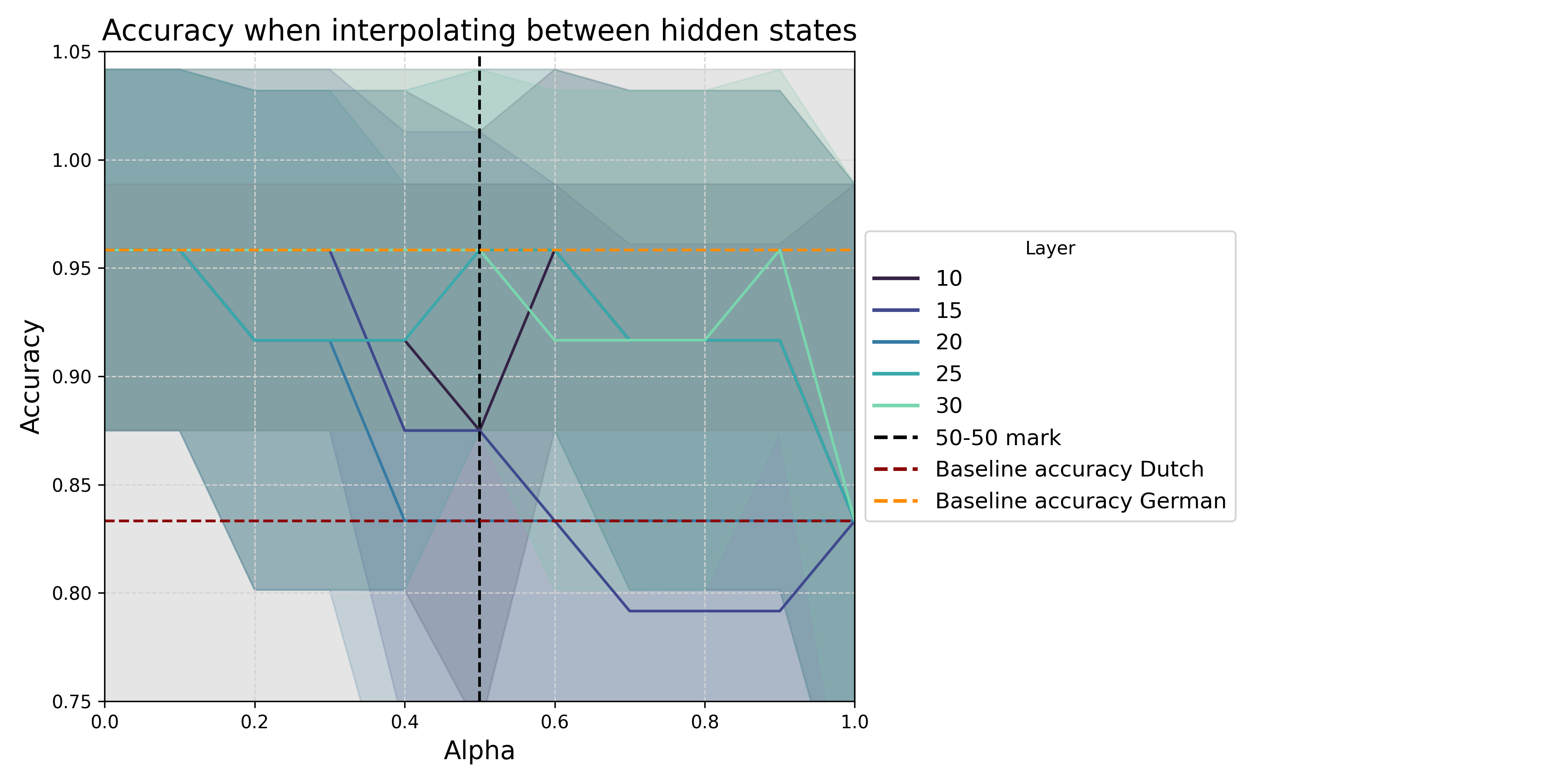} 
\end{minipage}
\begin{minipage}{0.49\textwidth}
    \centering
    \includegraphics[trim={1.5cm 0.5cm 1.5cm 1.5cm},clip,width=\textwidth]{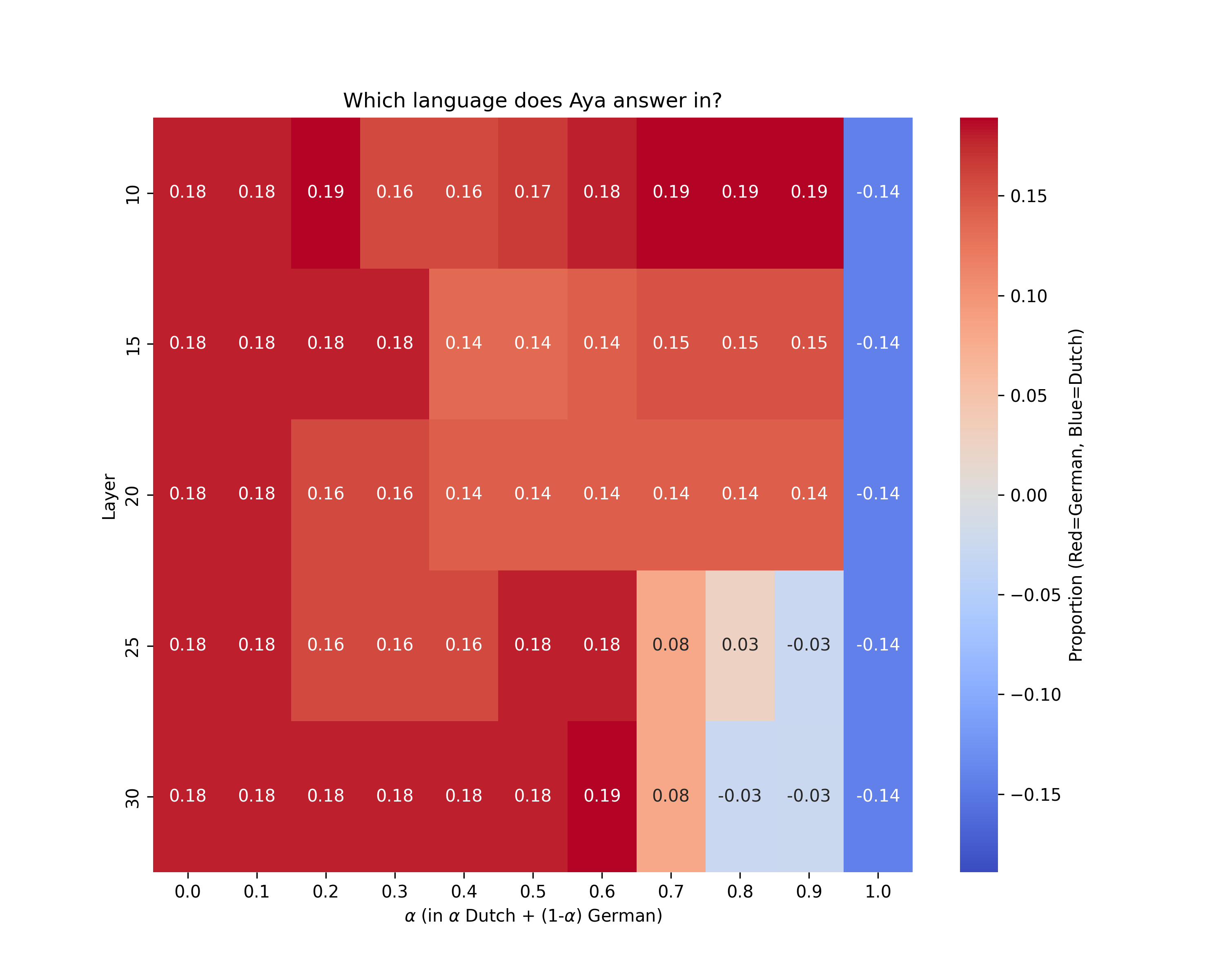} 
\end{minipage}
\caption{Hidden state interpolation between German prompts, and Dutch prompts in \aya. Left shows the accuracy (i.e., the proportion of times the model correctly outputs city in either language). Right shows the propensity of the model to answer in German (red) and Dutch (blue). }
\end{figure}

\begin{figure}[h]
\begin{minipage}{0.49\textwidth}
    \centering
    \includegraphics[trim={0 0 5cm 0},clip, width=\textwidth]{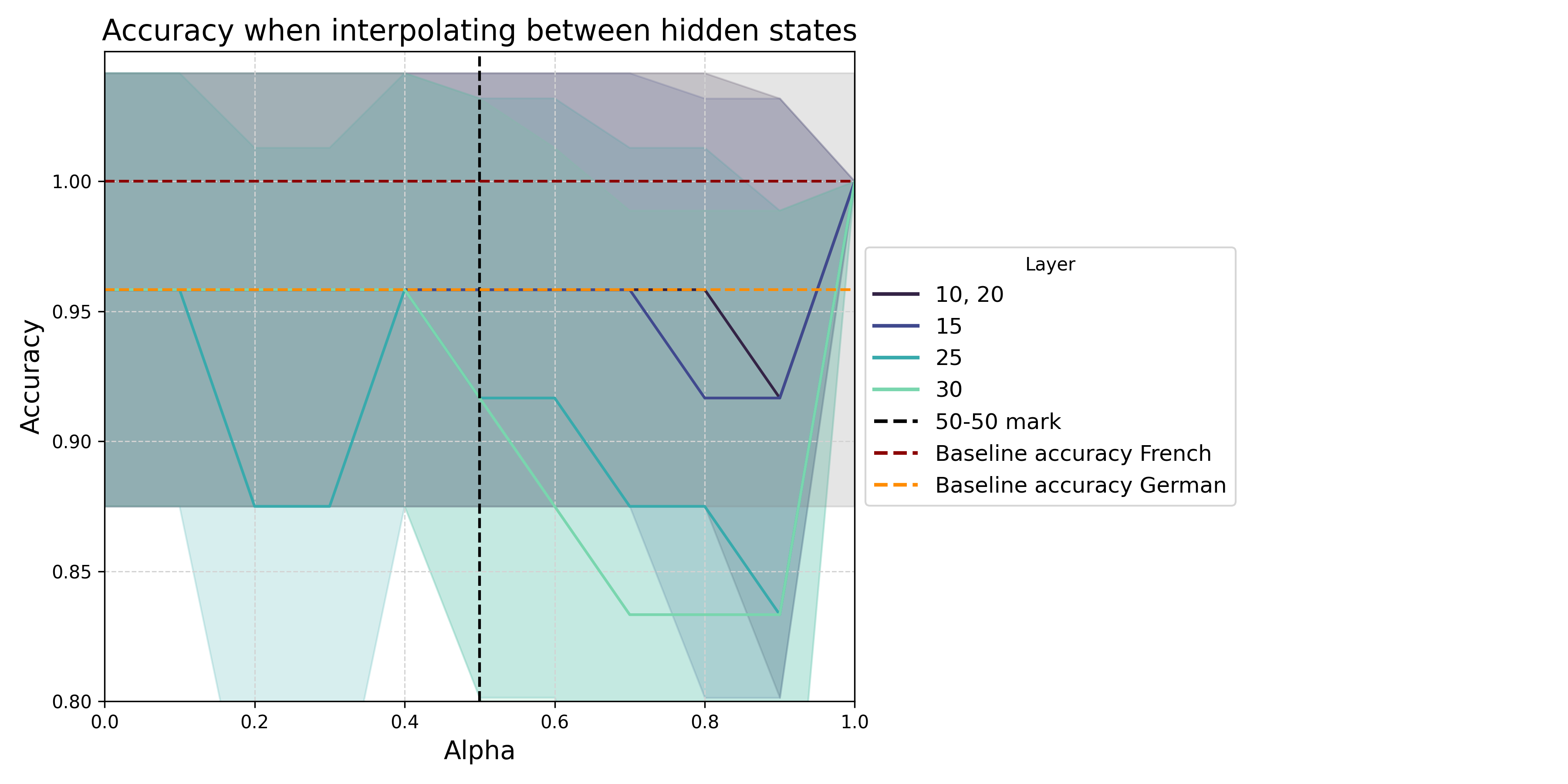} 
\end{minipage}
\begin{minipage}{0.49\textwidth}
    \centering
    \includegraphics[trim={1.5cm 0.5cm 1.5cm 1.5cm},clip,width=\textwidth]{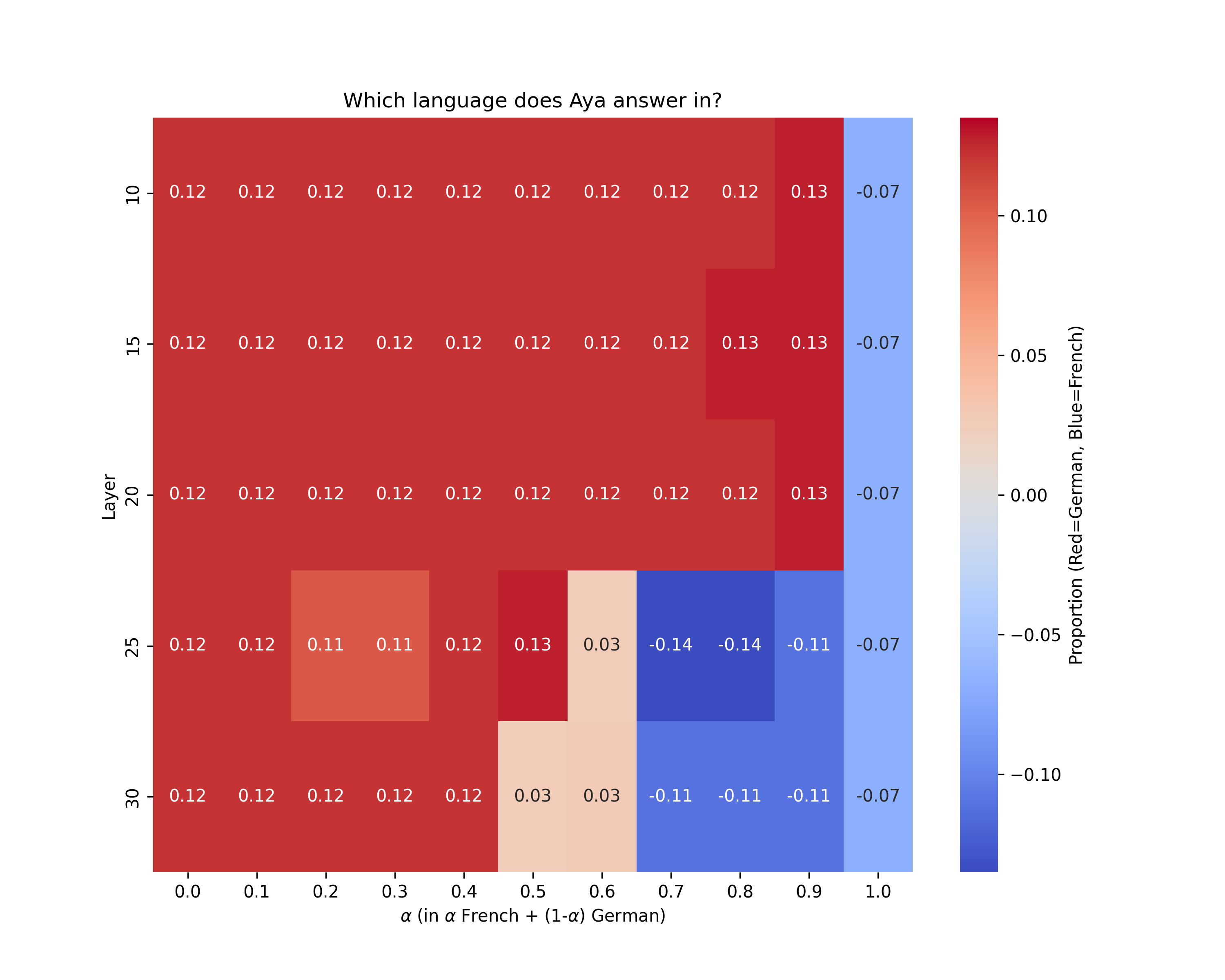} 
\end{minipage}
\caption{Hidden state interpolation between French prompts, and German prompts in \aya. Left shows the accuracy (i.e., the proportion of times the model correctly outputs city in either language). Right shows the propensity of the model to answer in German (red) and French (blue). }
\end{figure}

\begin{figure}[h]
\begin{minipage}{0.49\textwidth}
    \centering
    \includegraphics[trim={0 0 5cm 0},clip, width=\textwidth]{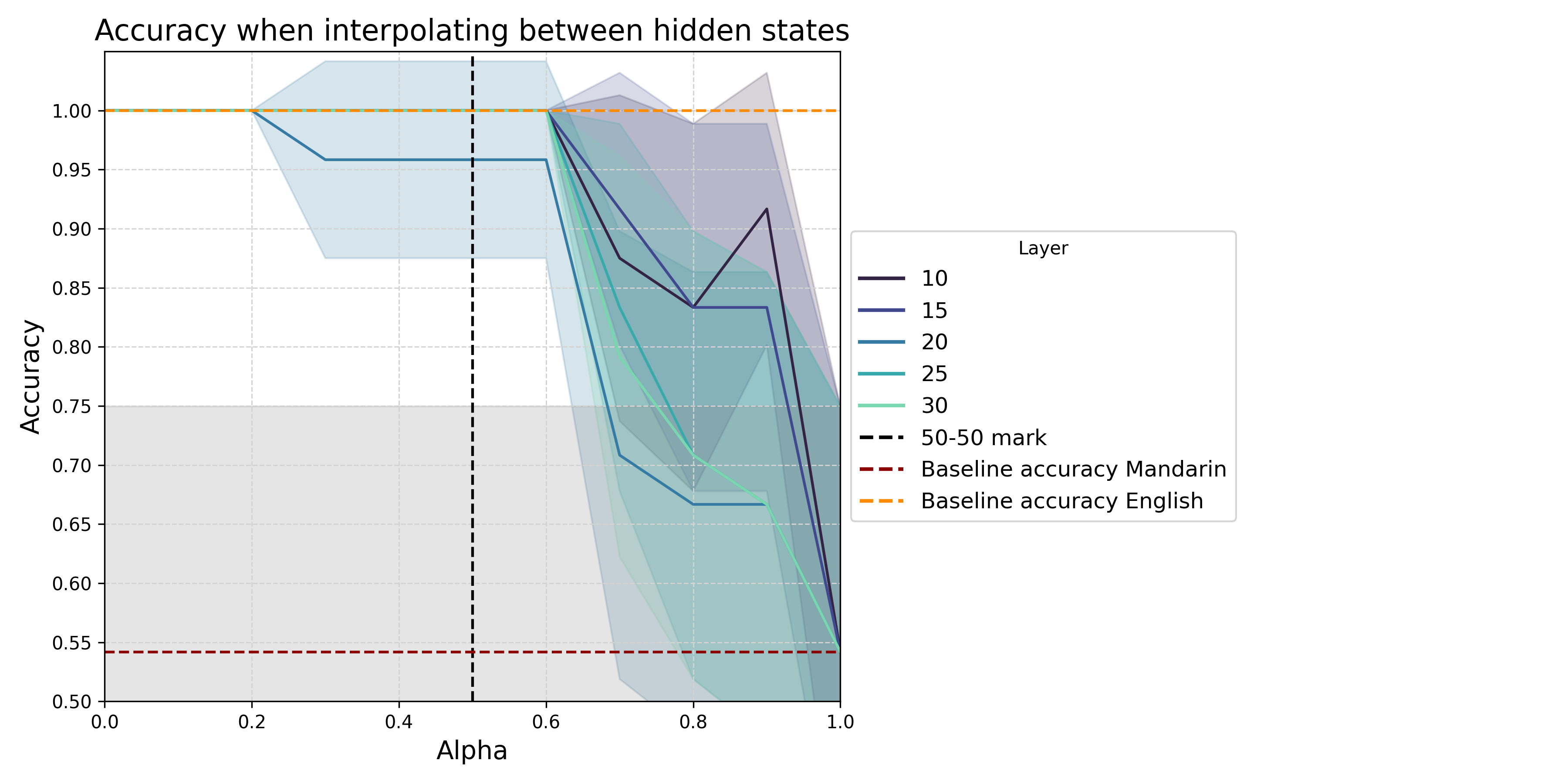} 
\end{minipage}
\begin{minipage}{0.49\textwidth}
    \centering
    \includegraphics[trim={1.5cm 0.5cm 1.5cm 1.5cm},clip,width=\textwidth]{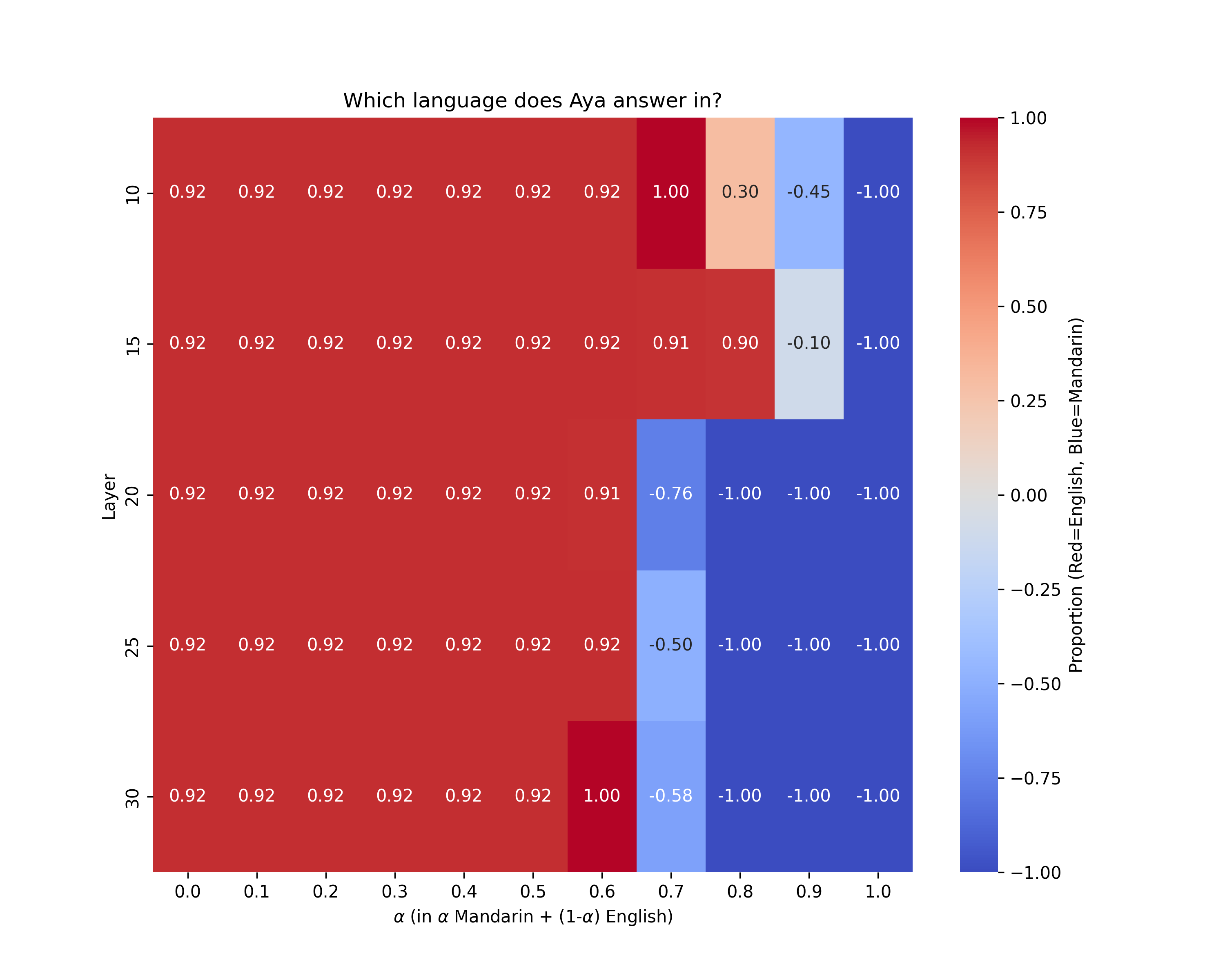} 
\end{minipage}
\caption{Hidden state interpolation between Mandarin prompts, and English prompts in \aya. Left shows the accuracy (i.e., the proportion of times the model correctly outputs city in either language). Right shows the propensity of the model to answer in English (red) and Mandarin (blue). }
\end{figure}

\begin{figure}[h]
\begin{minipage}{0.49\textwidth}
    \centering
    \includegraphics[trim={0 0 5cm 0},clip, width=\textwidth]{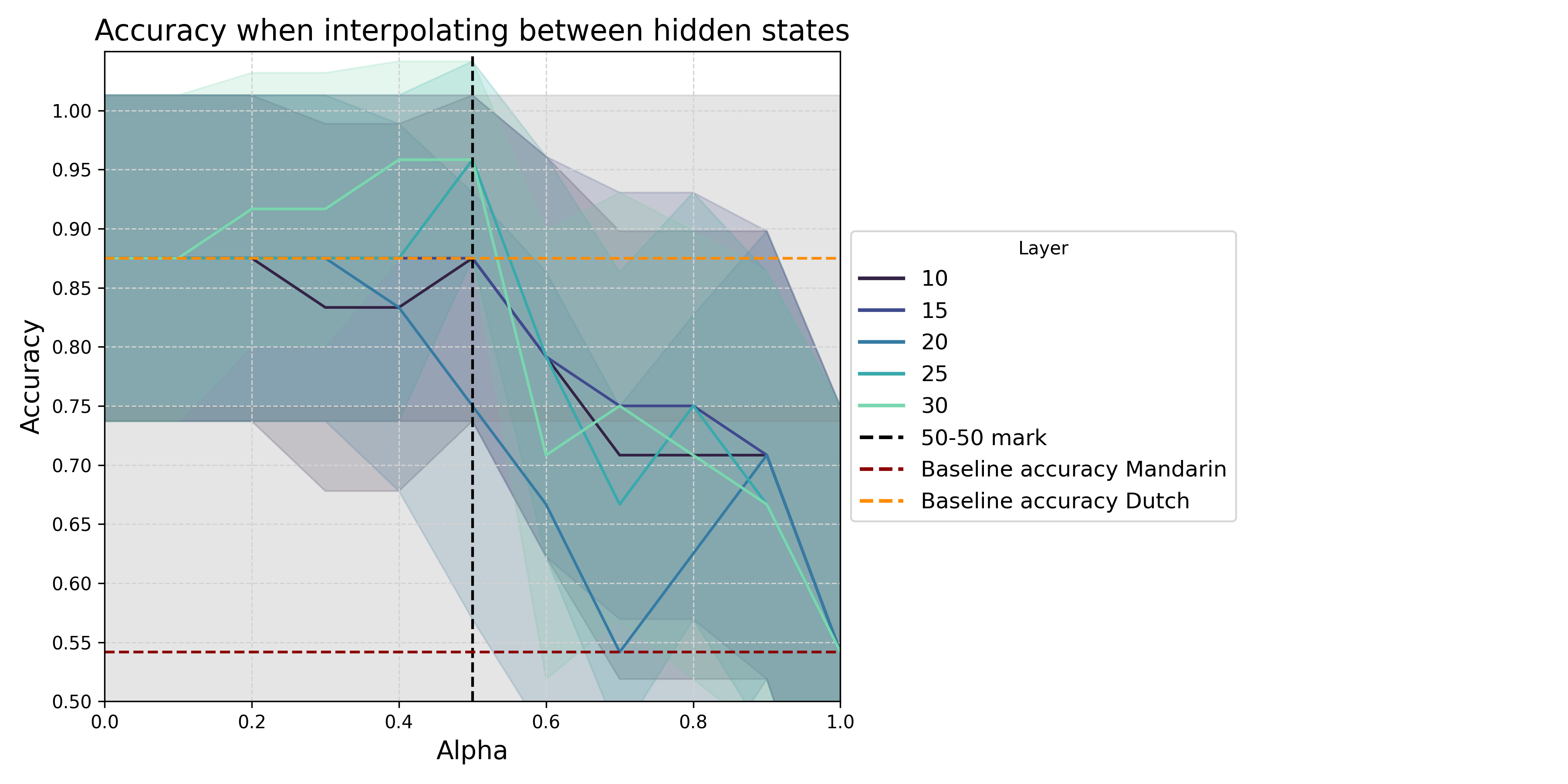} 
\end{minipage}
\begin{minipage}{0.49\textwidth}
    \centering
    \includegraphics[trim={1.5cm 0.5cm 1.5cm 1.5cm},clip,width=\textwidth]{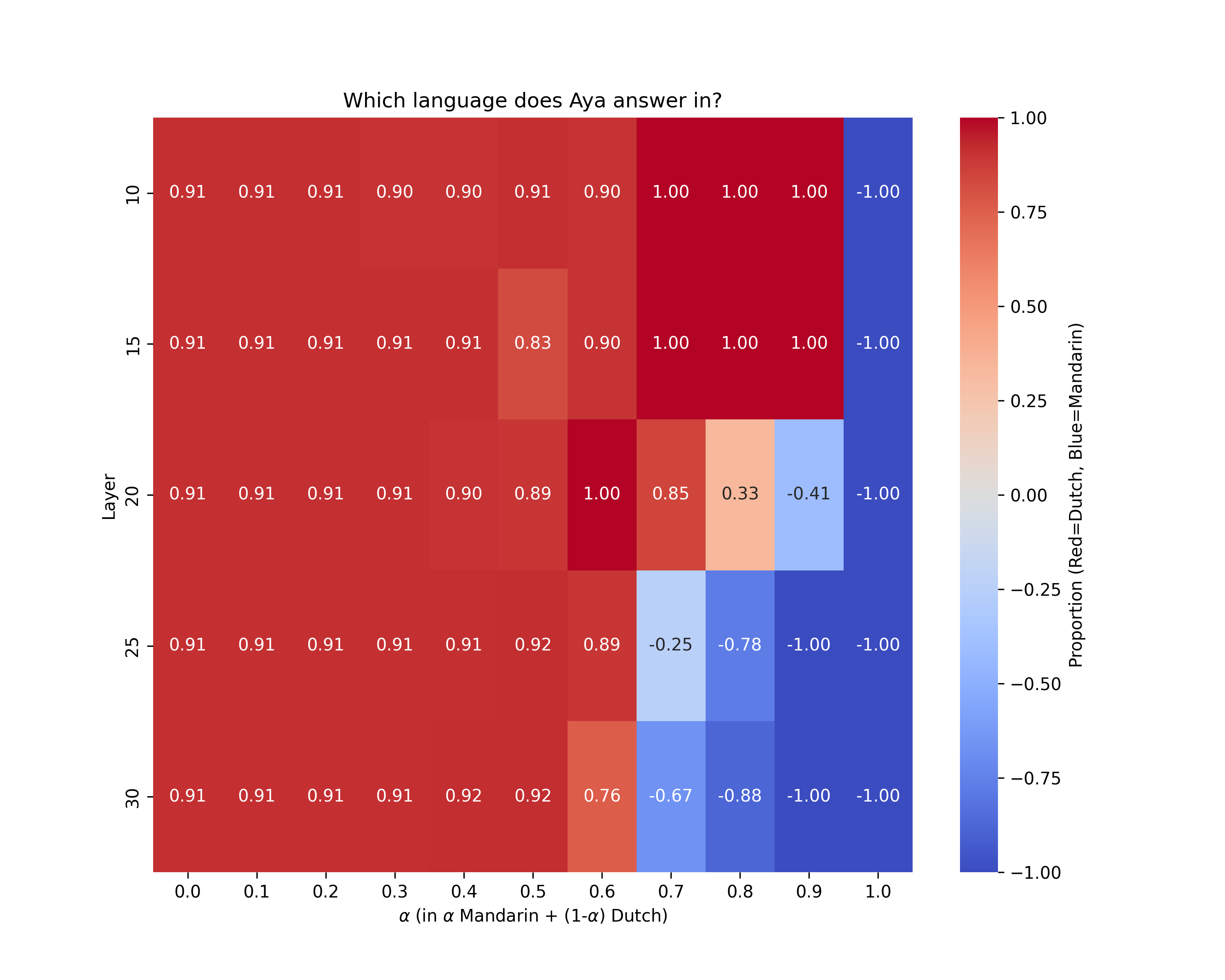} 
\end{minipage}
\caption{Hidden state interpolation between Mandarin prompts, and Dutch prompts in \aya. Left shows the accuracy (i.e., the proportion of times the model correctly outputs city in either language). Right shows the propensity of the model to answer in Dutch (red) and Mandarin (blue). }
\end{figure}

\begin{figure}[h]
\begin{minipage}{0.49\textwidth}
    \centering
    \includegraphics[trim={0 0 5cm 0},clip, width=\textwidth]{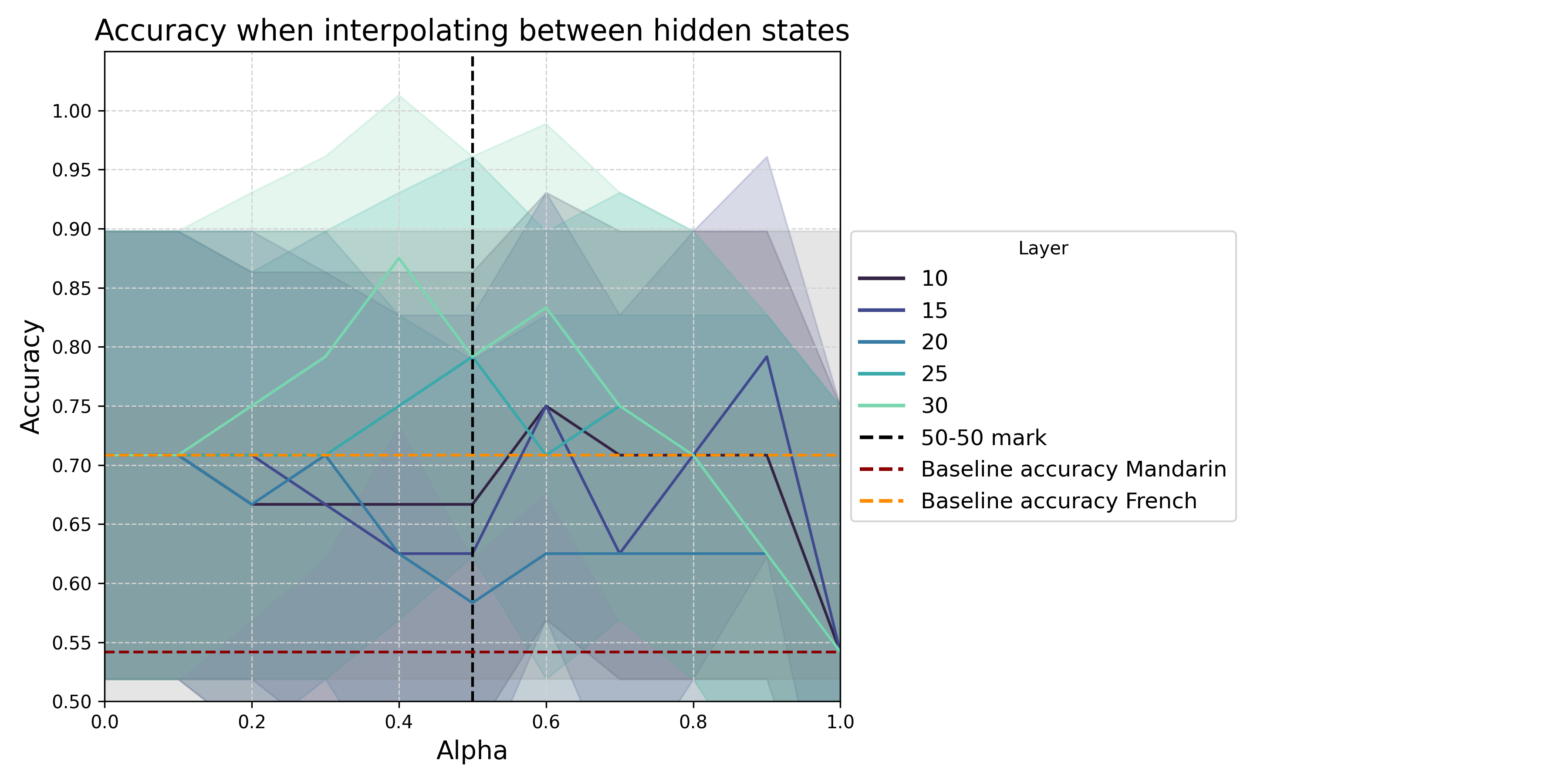} 
\end{minipage}
\begin{minipage}{0.49\textwidth}
    \centering
    \includegraphics[trim={1.5cm 0.5cm 1.5cm 1.5cm},clip,width=\textwidth]{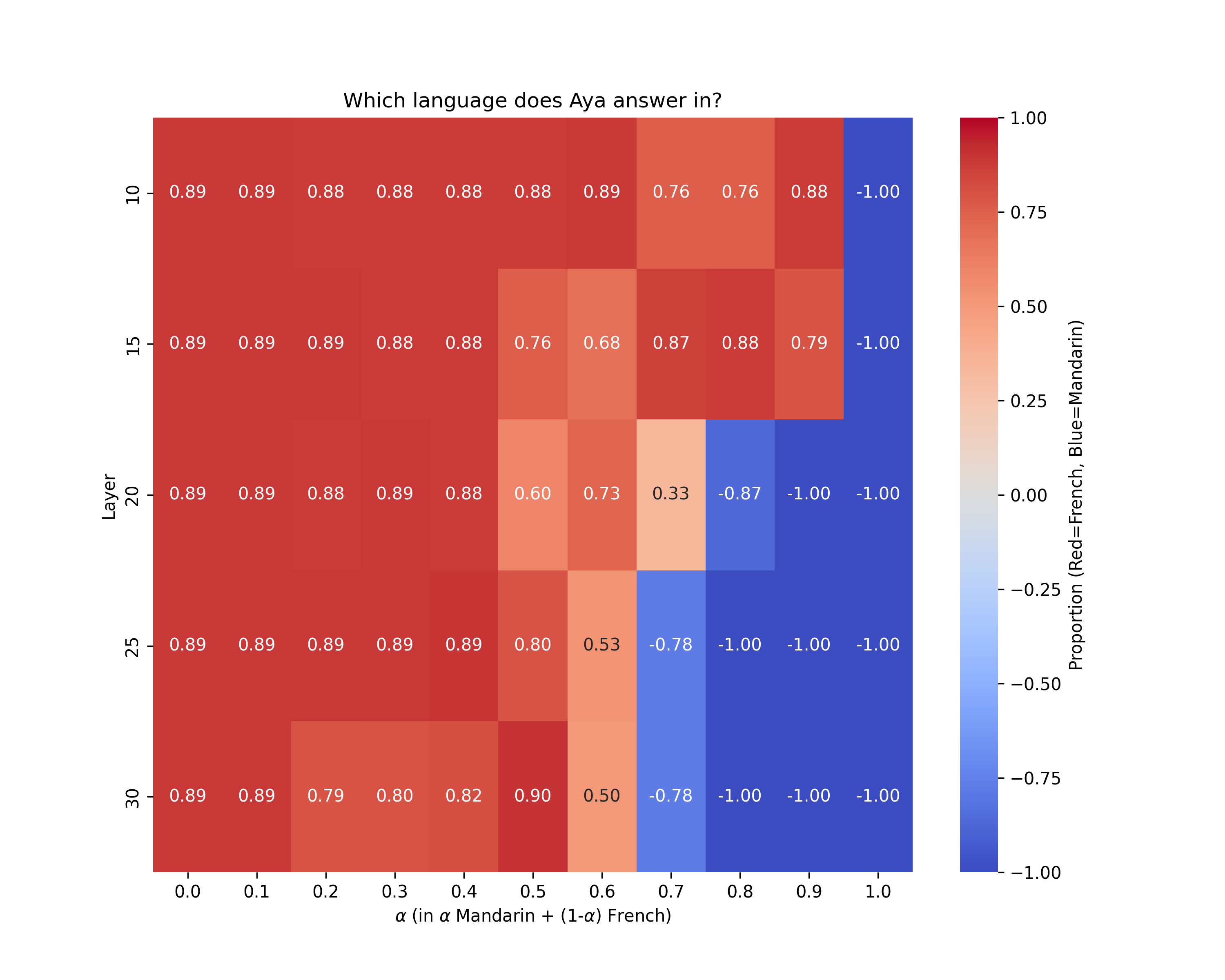} 
\end{minipage}
\caption{Hidden state interpolation between Mandarin prompts, and French prompts in \aya. Left shows the accuracy (i.e., the proportion of times the model correctly outputs city in either language). Right shows the propensity of the model to answer in French (red) and Mandarin (blue). }
\end{figure}

\begin{figure}[h]
\begin{minipage}{0.49\textwidth}
    \centering
    \includegraphics[trim={0 0 5cm 0},clip, width=\textwidth]{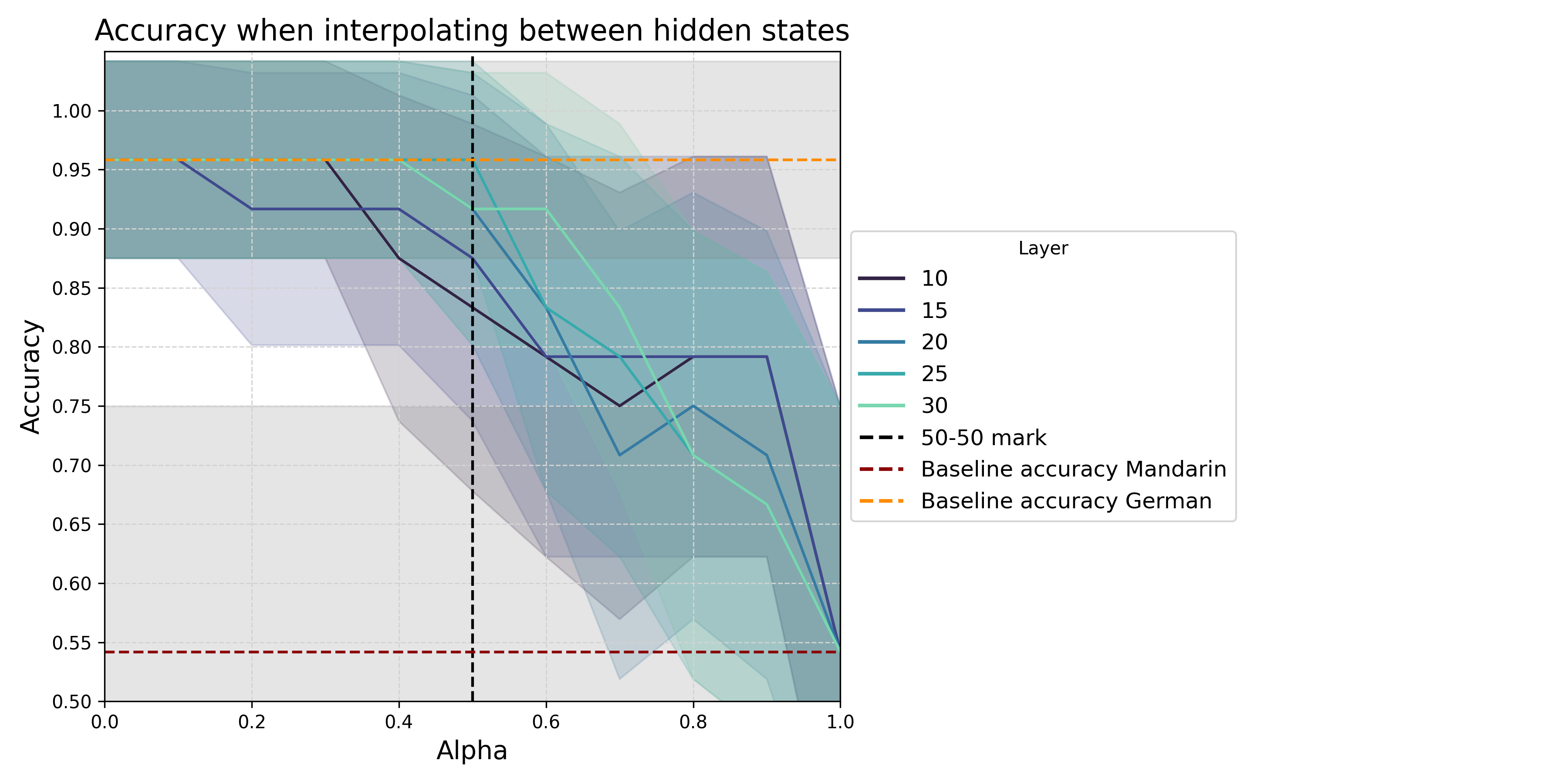} 
\end{minipage}
\begin{minipage}{0.49\textwidth}
    \centering
    \includegraphics[trim={1.5cm 0.5cm 1.5cm 1.5cm},clip,width=\textwidth]{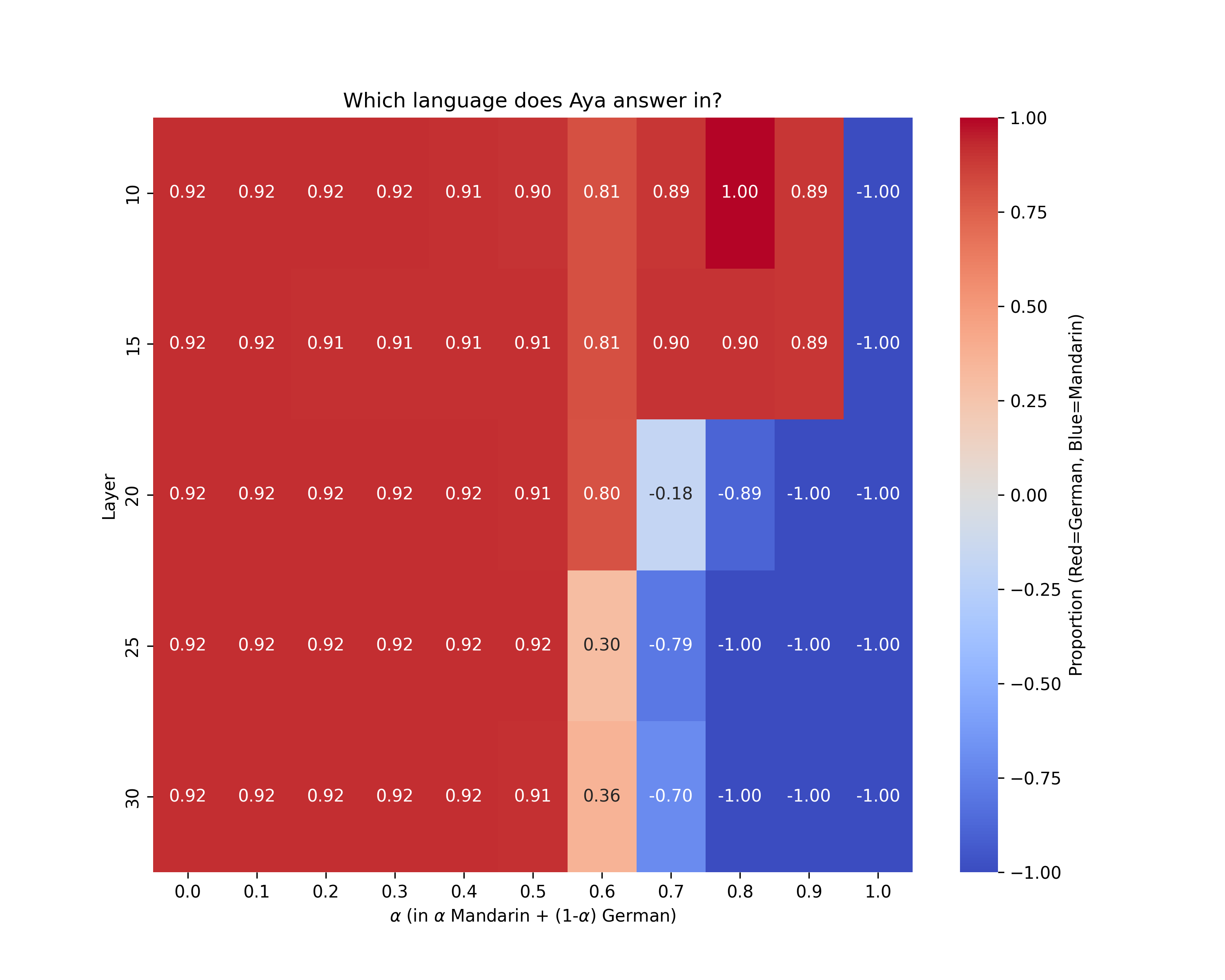} 
\end{minipage}
\caption{Hidden state interpolation between Mandarin prompts, and German prompts in \aya. Left shows the accuracy (i.e., the proportion of times the model correctly outputs city in either language). Right shows the propensity of the model to answer in German (red) and Mandarin (blue). }
\end{figure}

\FloatBarrier
\newpage 

\subsubsection{Llama} 

\begin{figure}[h]
\begin{minipage}{0.49\textwidth}
    \centering
    \includegraphics[trim={0 0 5cm 0},clip, width=\textwidth]{figures/interpolate/FR_ENG_aya__interpolate_results.png} 
\end{minipage}
\begin{minipage}{0.49\textwidth}
    \centering
    \includegraphics[trim={1.5cm 0.5cm 1.5cm 1.5cm},clip,width=\textwidth]{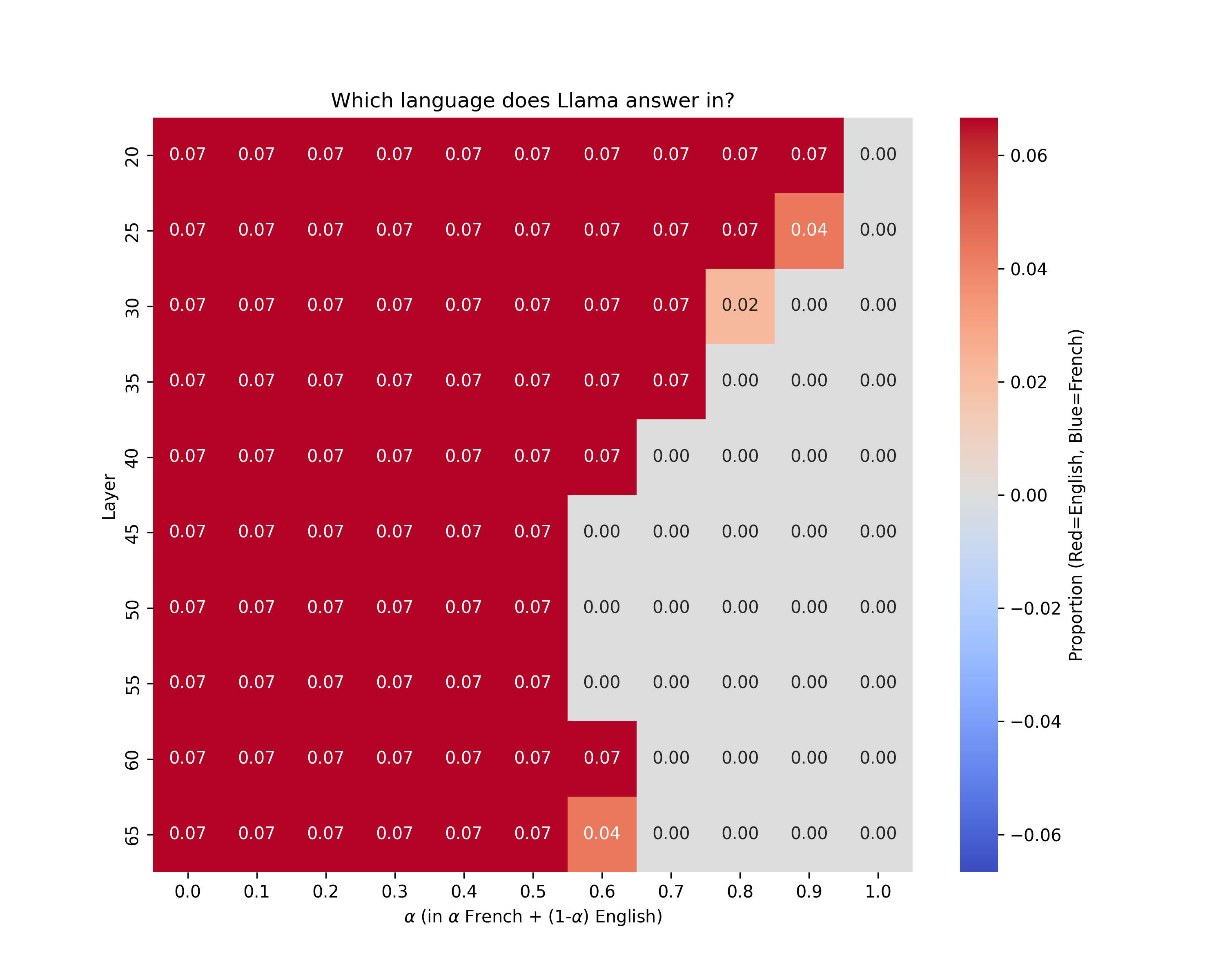} 
\end{minipage}
\caption{Hidden state interpolation between English prompts, and French prompts in \llama. Left shows the accuracy (i.e., the proportion of times the model correctly outputs city in either language). Right shows the propensity of the model to answer in English (red) and French (blue). }

\end{figure}

\begin{figure}[h]
\begin{minipage}{0.49\textwidth}
    \centering
    \includegraphics[trim={0 0 5cm 0},clip, width=\textwidth]{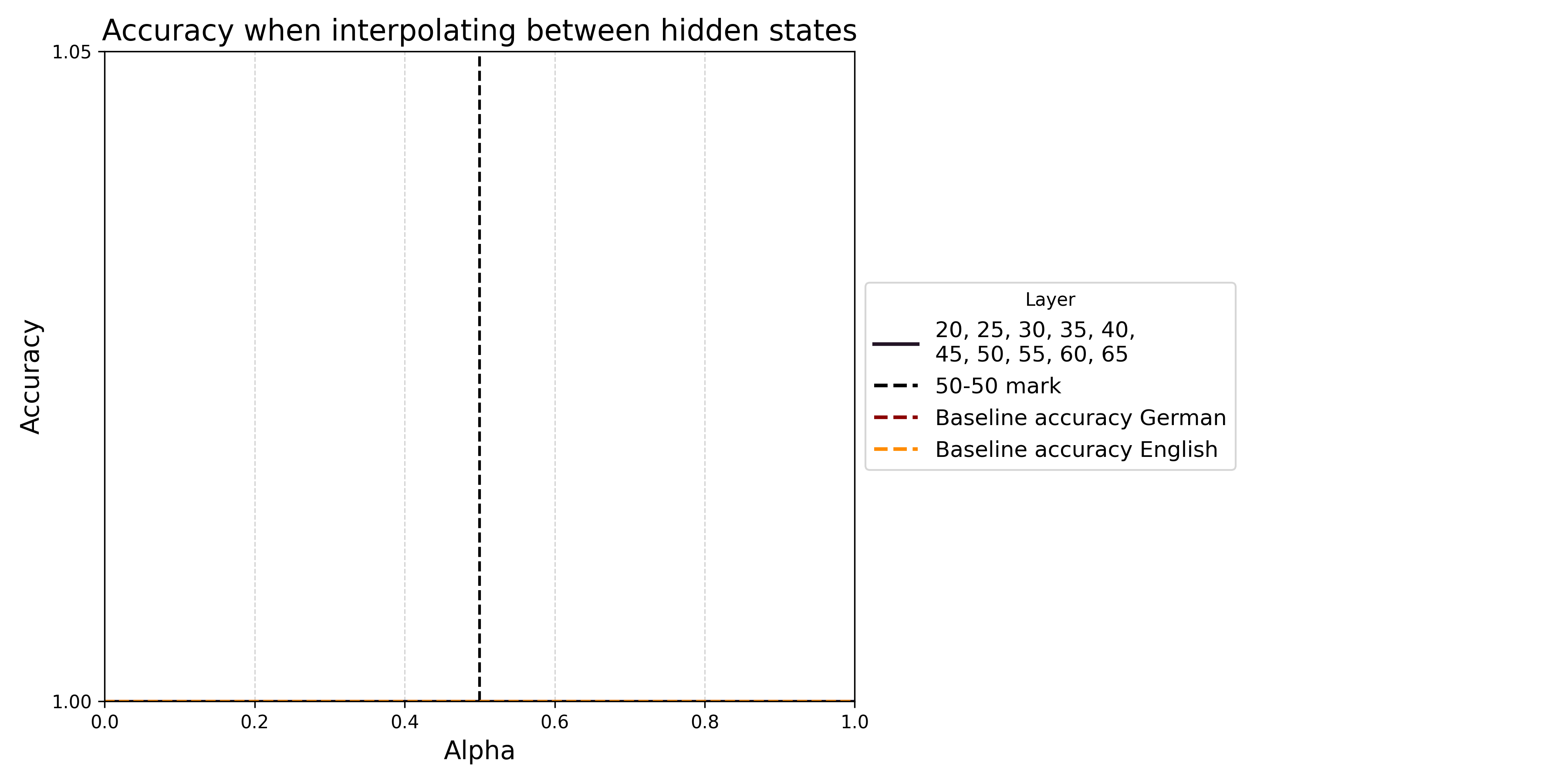} 
\end{minipage}
\begin{minipage}{0.49\textwidth}
    \centering
    \includegraphics[trim={1.5cm 0.5cm 1.5cm 1.5cm},clip,width=\textwidth]{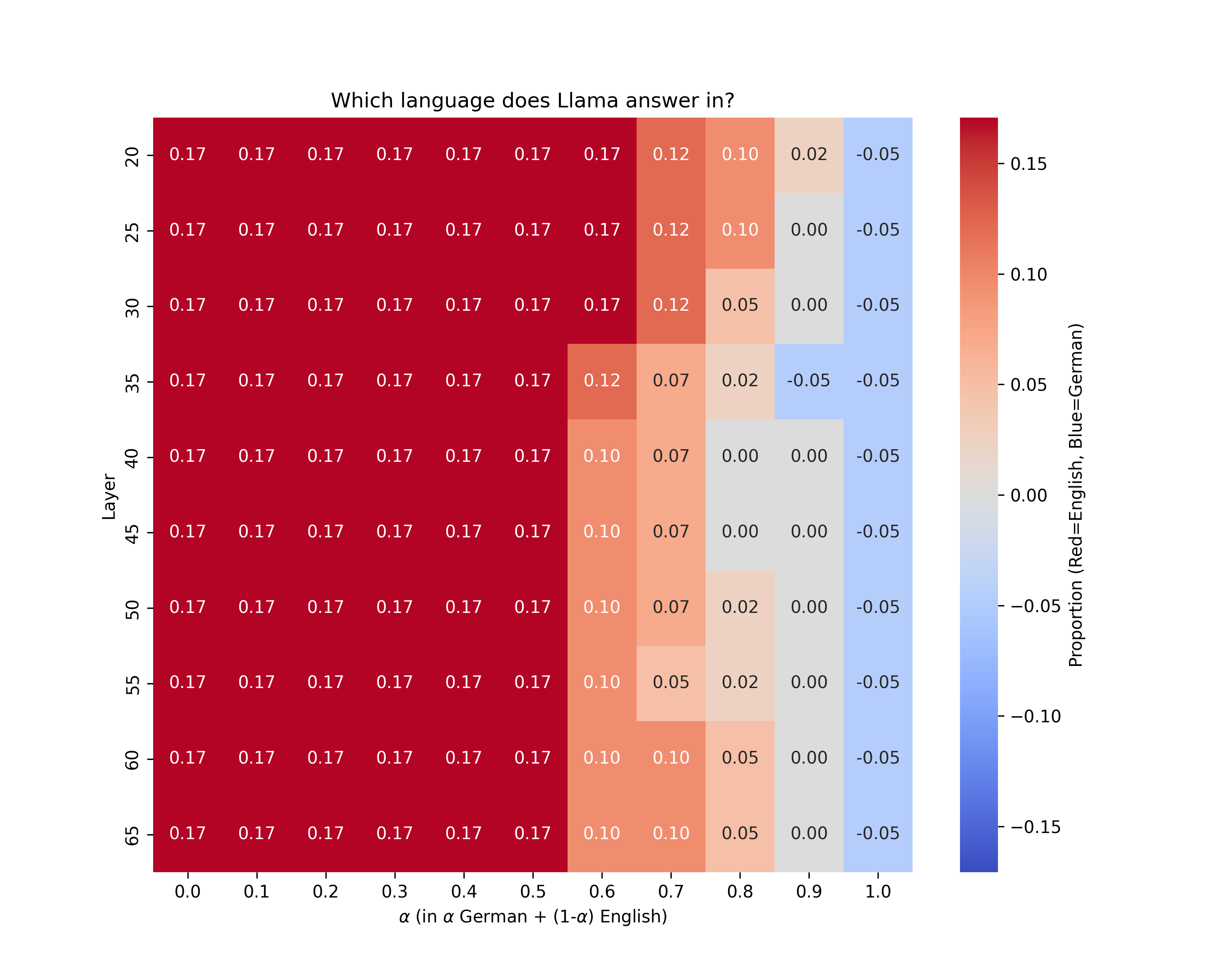} 
\end{minipage}
\caption{Hidden state interpolation between English prompts, and German prompts in \llama. Left shows the accuracy (i.e., the proportion of times the model correctly outputs city in either language). Right shows the propensity of the model to answer in English (red) and German (blue). }
\end{figure}

\begin{figure}[h]
\begin{minipage}{0.49\textwidth}
    \centering
    \includegraphics[trim={0 0 5cm 0},clip, width=\textwidth]{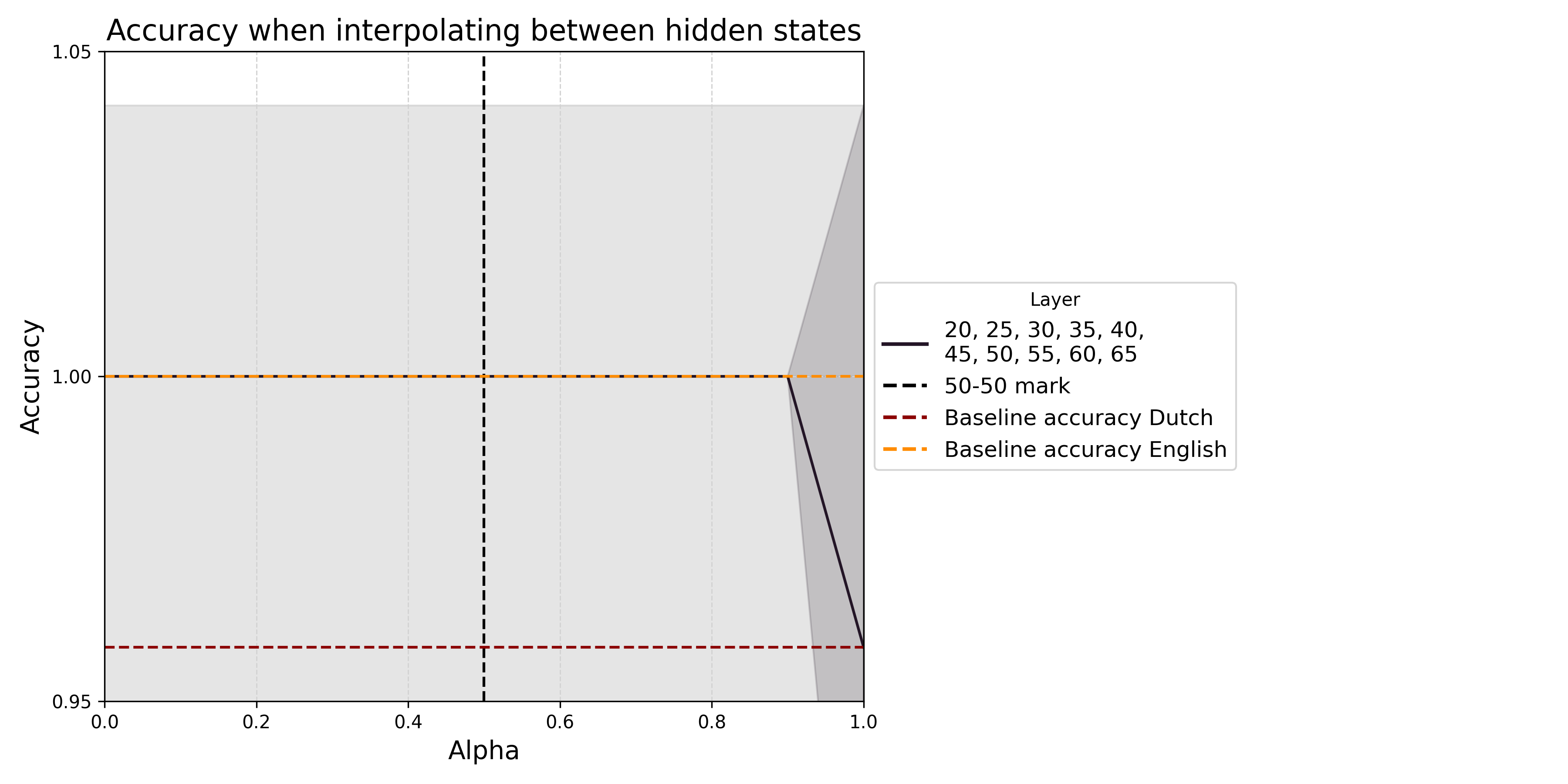} 
\end{minipage}
\begin{minipage}{0.49\textwidth}
    \centering
    \includegraphics[trim={1.5cm 0.5cm 1.5cm 1.5cm},clip,width=\textwidth]{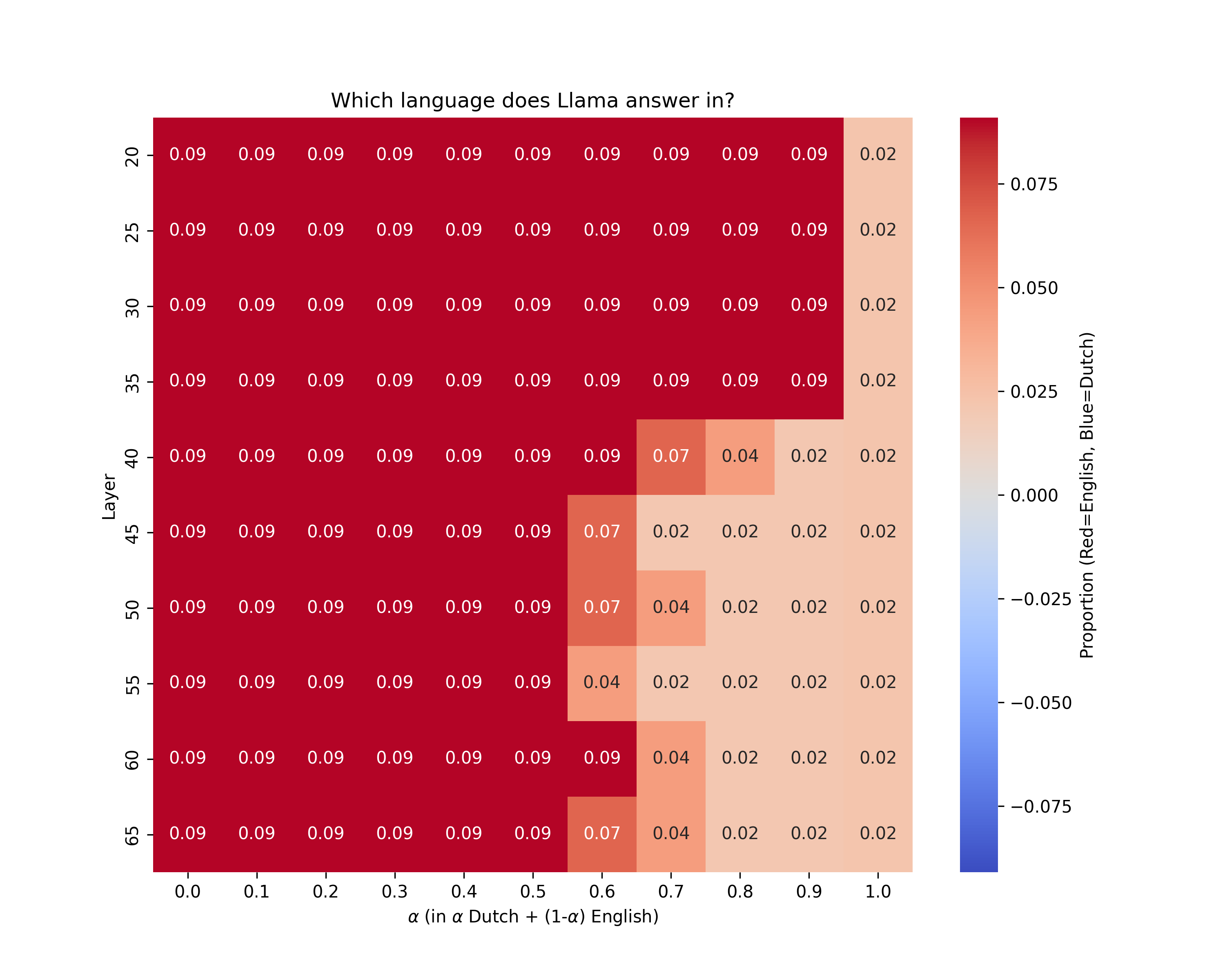} 
\end{minipage}
\caption{Hidden state interpolation between English prompts, and Dutch prompts in \llama. Left shows the accuracy (i.e., the proportion of times the model correctly outputs city in either language). Right shows the propensity of the model to answer in English (red) and Dutch (blue). }
\end{figure}

\begin{figure}[h]
\begin{minipage}{0.49\textwidth}
    \centering
    \includegraphics[trim={0 0 5cm 0},clip, width=\textwidth]{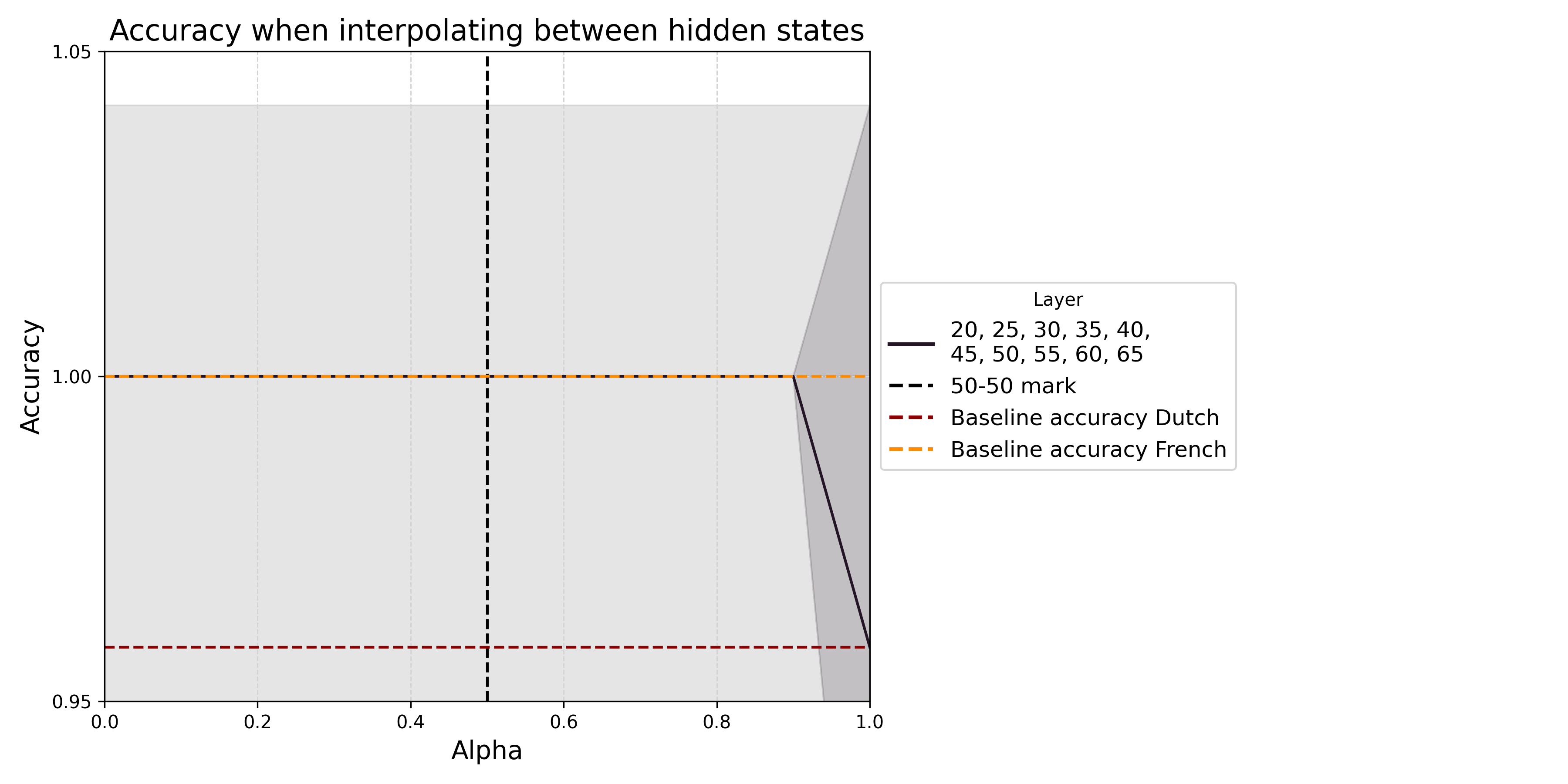} 
\end{minipage}
\begin{minipage}{0.49\textwidth}
    \centering
    \includegraphics[trim={1.5cm 0.5cm 1.5cm 1.5cm},clip,width=\textwidth]{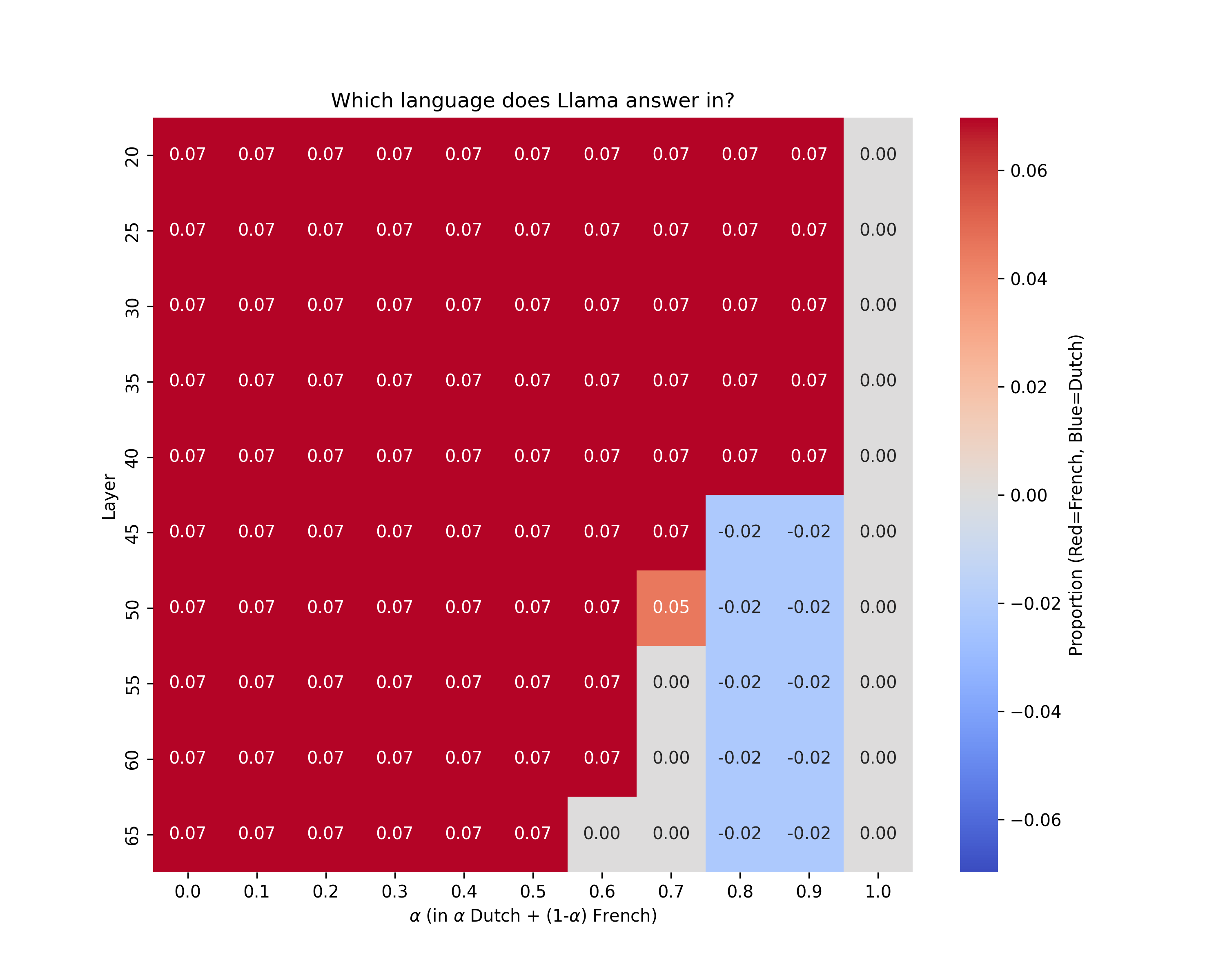} 
\end{minipage}
\caption{Hidden state interpolation between Dutch prompts, and French prompts in \llama. Left shows the accuracy (i.e., the proportion of times the model correctly outputs city in either language). Right shows the propensity of the model to answer in French (red) and Dutch (blue). }
\end{figure}

\begin{figure}[h]
\begin{minipage}{0.49\textwidth}
    \centering
    \includegraphics[trim={0 0 5cm 0},clip, width=\textwidth]{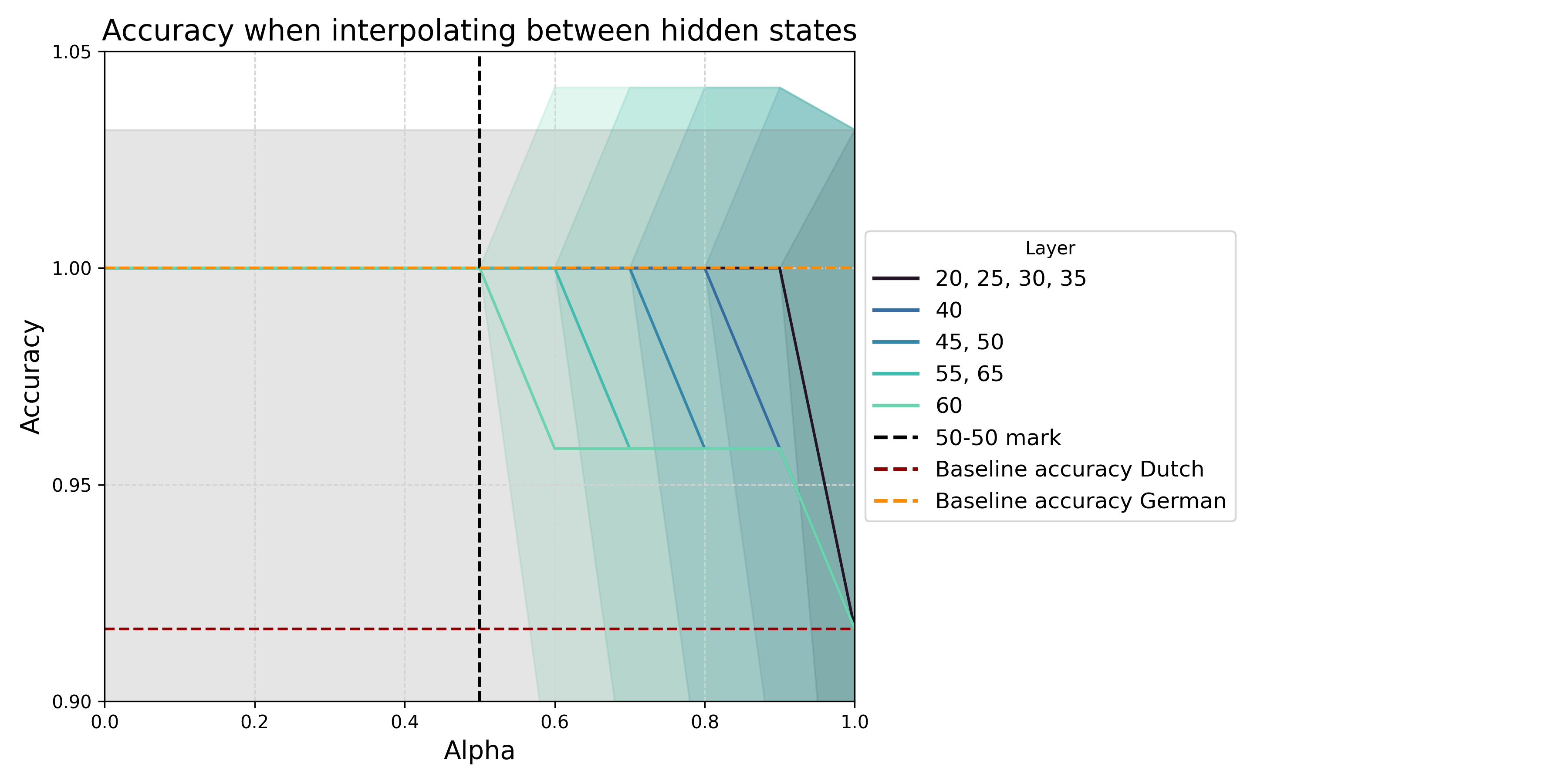} 
\end{minipage}
\begin{minipage}{0.49\textwidth}
    \centering
    \includegraphics[trim={1.5cm 0.5cm 1.5cm 1.5cm},clip,width=\textwidth]{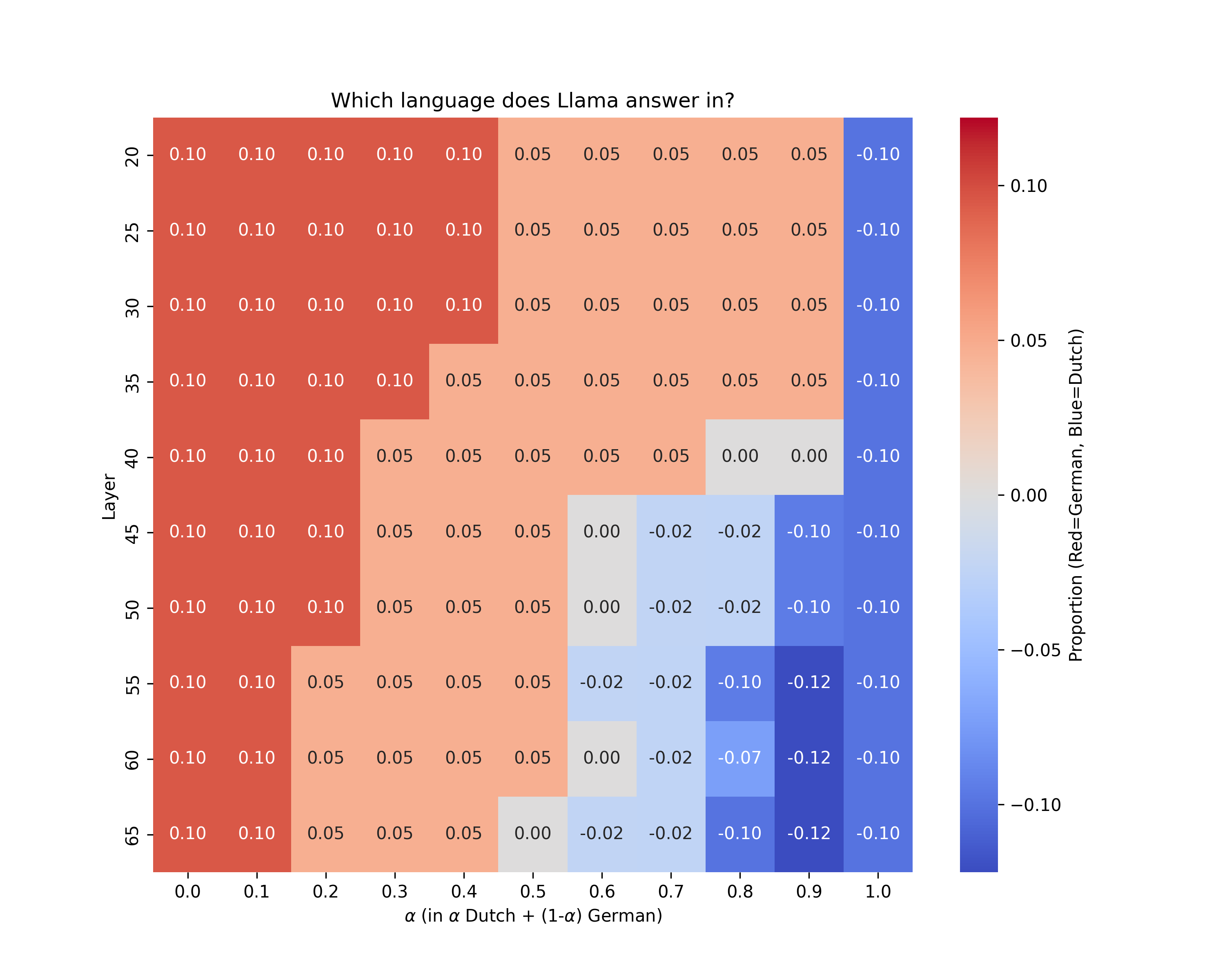} 
\end{minipage}
\caption{Hidden state interpolation between English prompts, and Dutch prompts in \llama. Left shows the accuracy (i.e., the proportion of times the model correctly outputs city in either language). Right shows the propensity of the model to answer in English (red) and Dutch (blue). }
\end{figure}

\begin{figure}[h]
\begin{minipage}{0.49\textwidth}
    \centering
    \includegraphics[trim={0 0 5cm 0},clip, width=\textwidth]{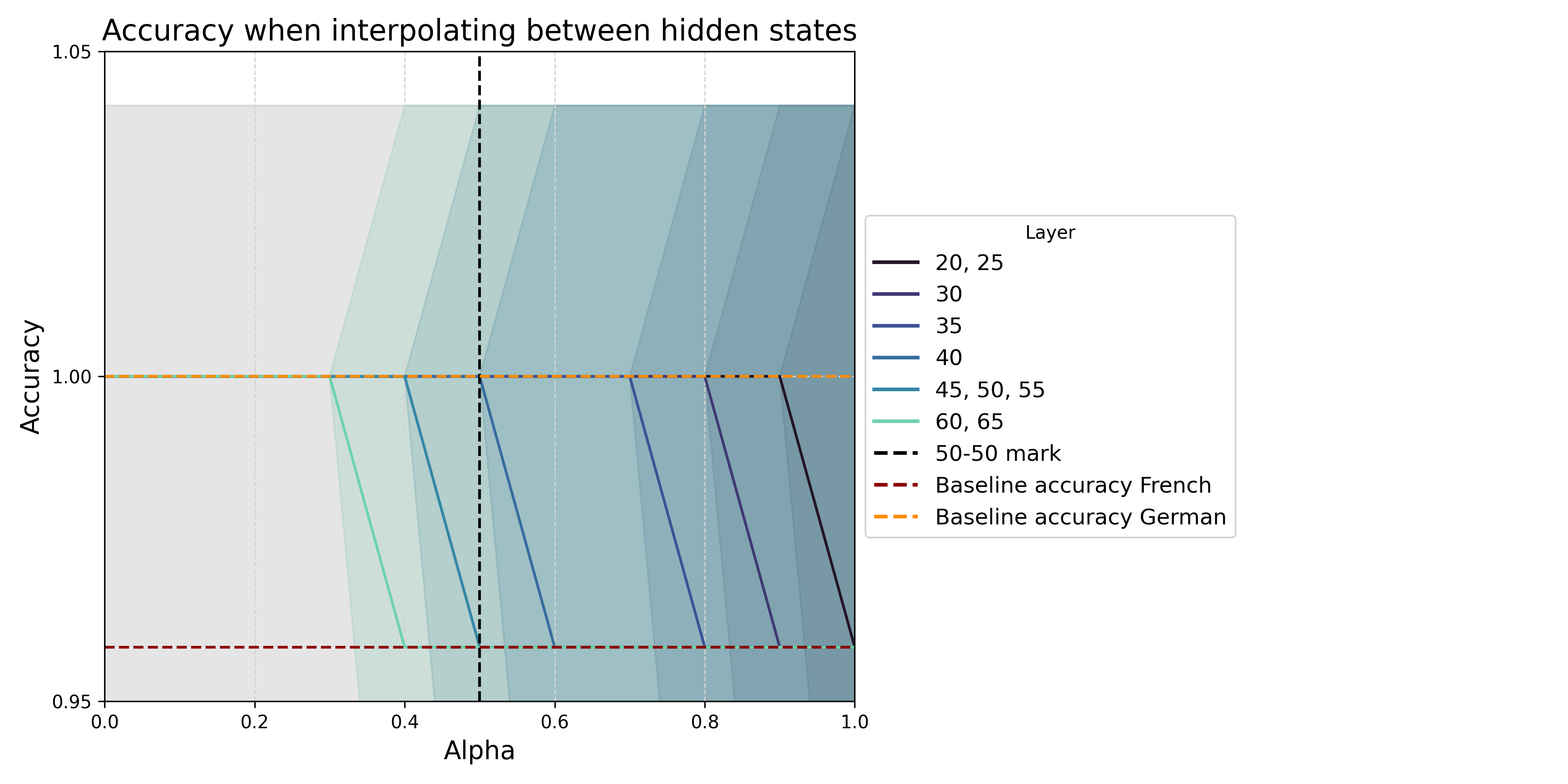} 
\end{minipage}
\begin{minipage}{0.49\textwidth}
    \centering
    \includegraphics[trim={1.5cm 0.5cm 1.5cm 1.5cm},clip,width=\textwidth]{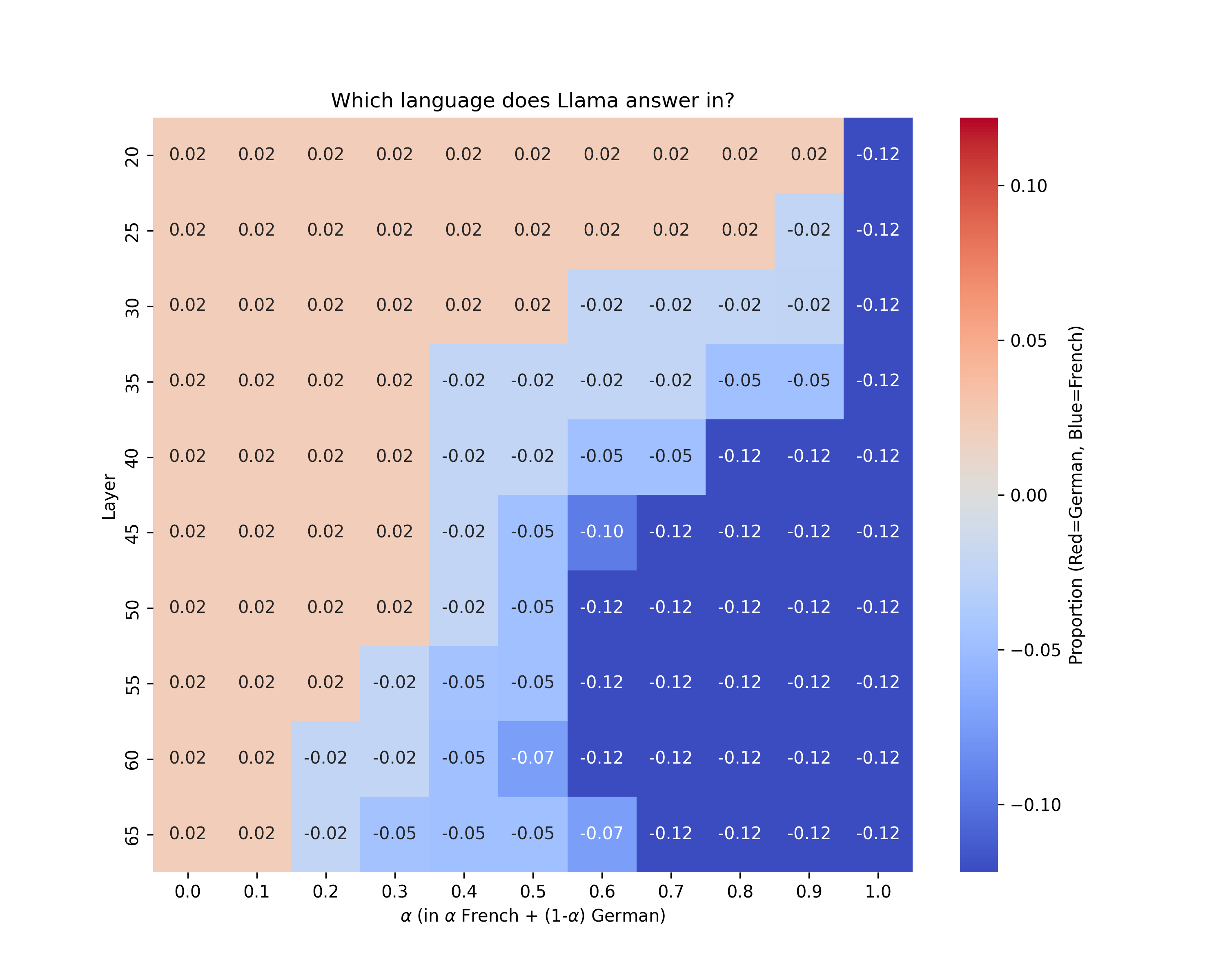} 
\end{minipage}
\caption{Hidden state interpolation between German prompts, and French prompts in \llama. Left shows the accuracy (i.e., the proportion of times the model correctly outputs city in either language). Right shows the propensity of the model to answer in German (red) and French (blue). }
\end{figure}

\begin{figure}[h]
\begin{minipage}{0.49\textwidth}
    \centering
    \includegraphics[trim={0 0 5cm 0},clip, width=\textwidth]{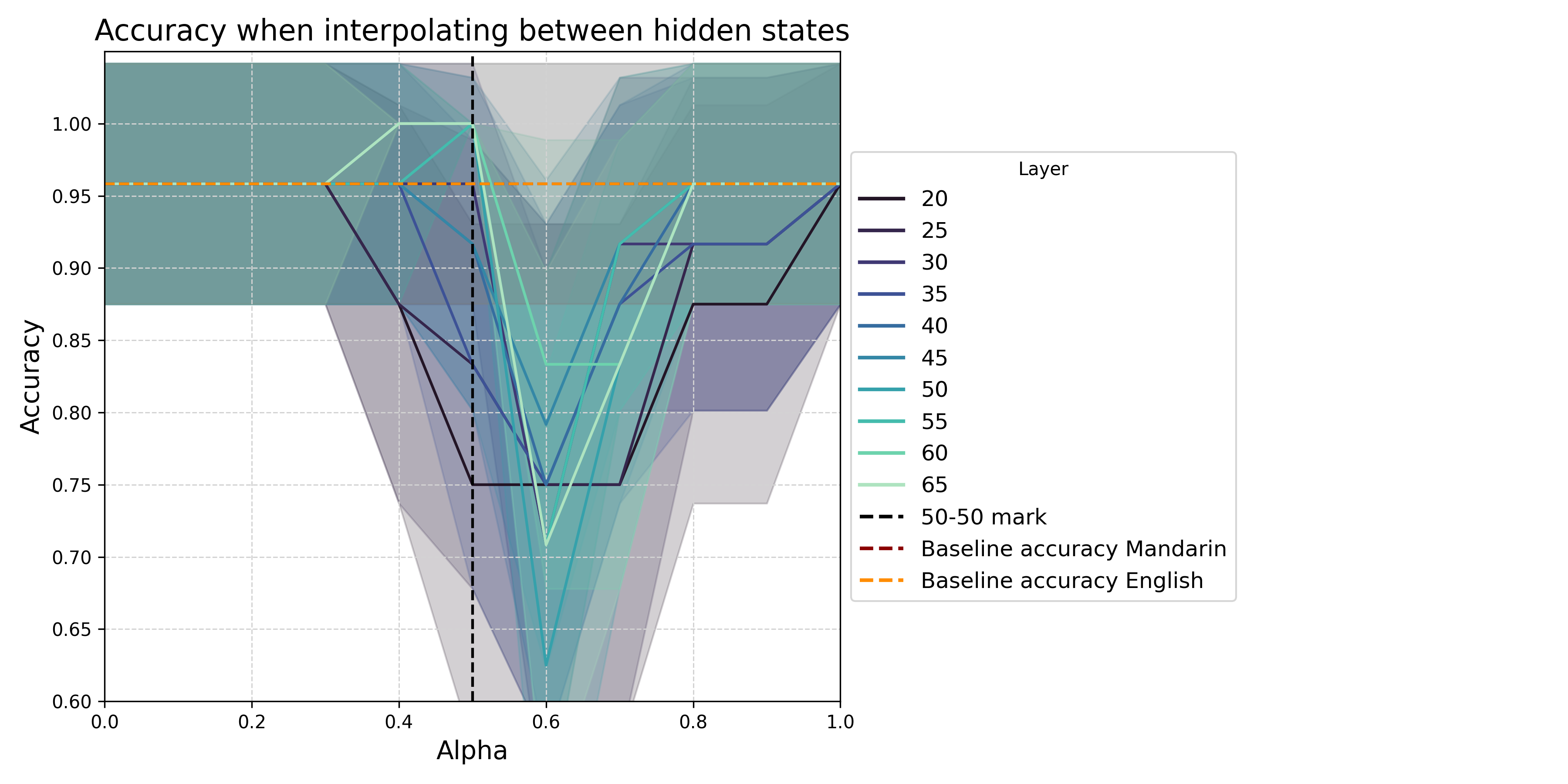}
\end{minipage}
\begin{minipage}{0.49\textwidth}
    \centering
    \includegraphics[trim={1.5cm 0.5cm 1.5cm 1.5cm},clip,width=\textwidth]{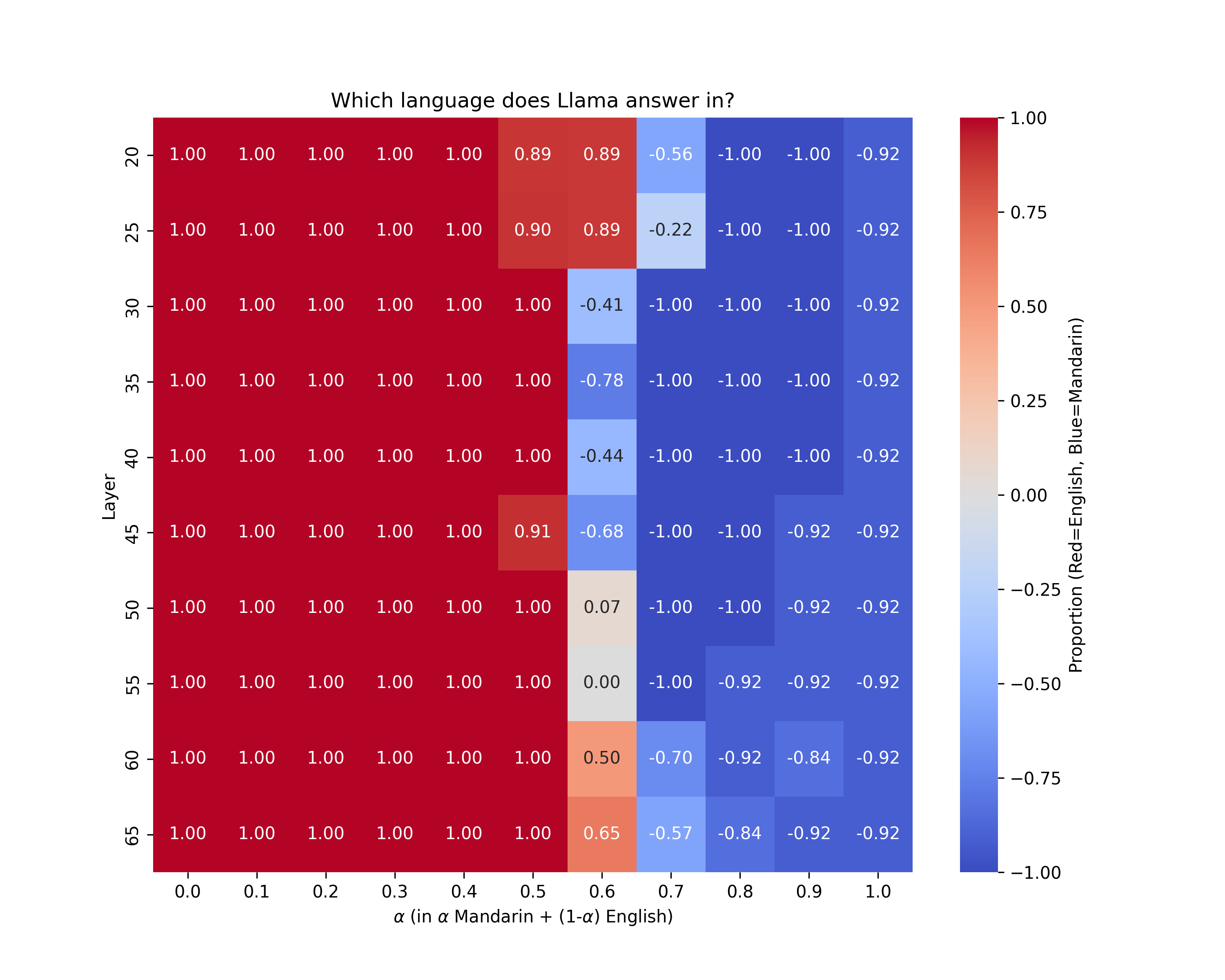} 
\end{minipage}
\caption{Hidden state interpolation between English prompts, and Mandarin prompts in \llama. Left shows the accuracy (i.e., the proportion of times the model correctly outputs city in either language). Right shows the propensity of the model to answer in English (red) and Mandarin (blue). }
\end{figure}

\begin{figure}[h]
\begin{minipage}{0.49\textwidth}
    \centering
    \includegraphics[trim={0 0 5cm 0},clip, width=\textwidth]{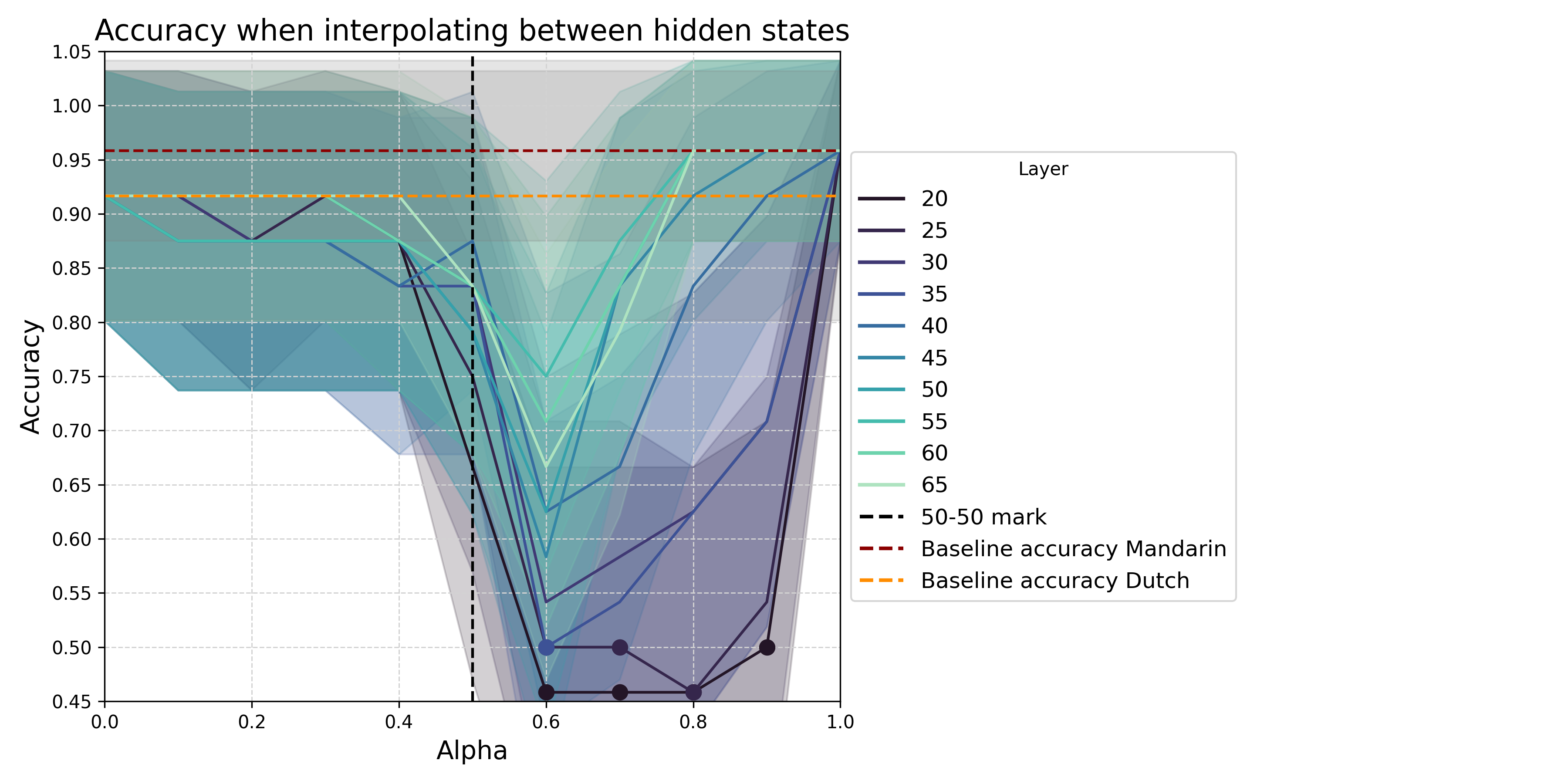} 
\end{minipage}
\begin{minipage}{0.49\textwidth}
    \centering
    \includegraphics[trim={1.5cm 0.5cm 1.5cm 1.5cm},clip,width=\textwidth]{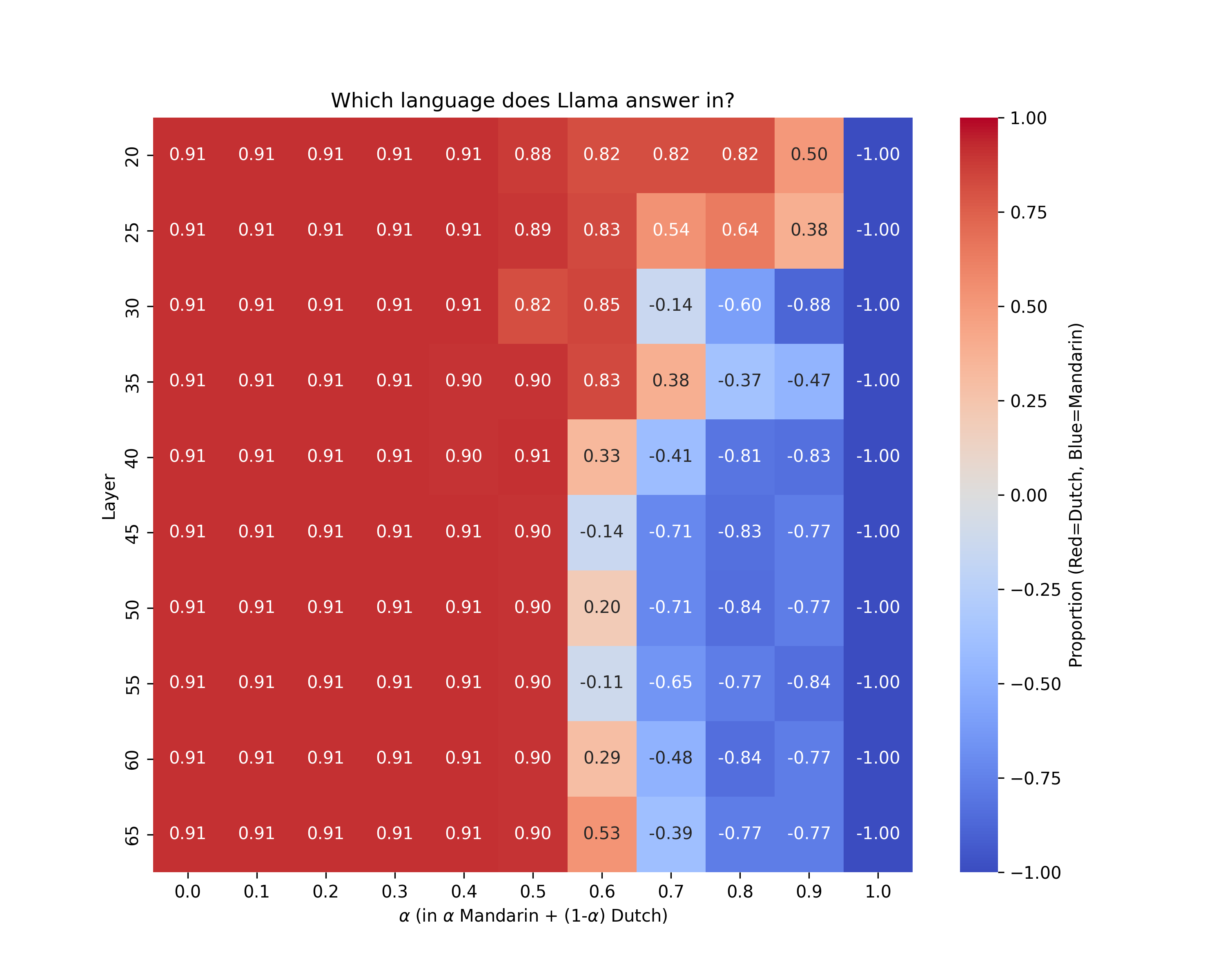} 
\end{minipage}
\caption{Hidden state interpolation between Dutch prompts, and Mandarin prompts in \llama. Left shows the accuracy (i.e., the proportion of times the model correctly outputs city in either language). Right shows the propensity of the model to answer in Dutch (red) and Mandarin (blue). }
\end{figure}

\begin{figure}[h]
\begin{minipage}{0.49\textwidth}
    \centering
    \includegraphics[trim={0 0 5cm 0},clip, width=\textwidth]{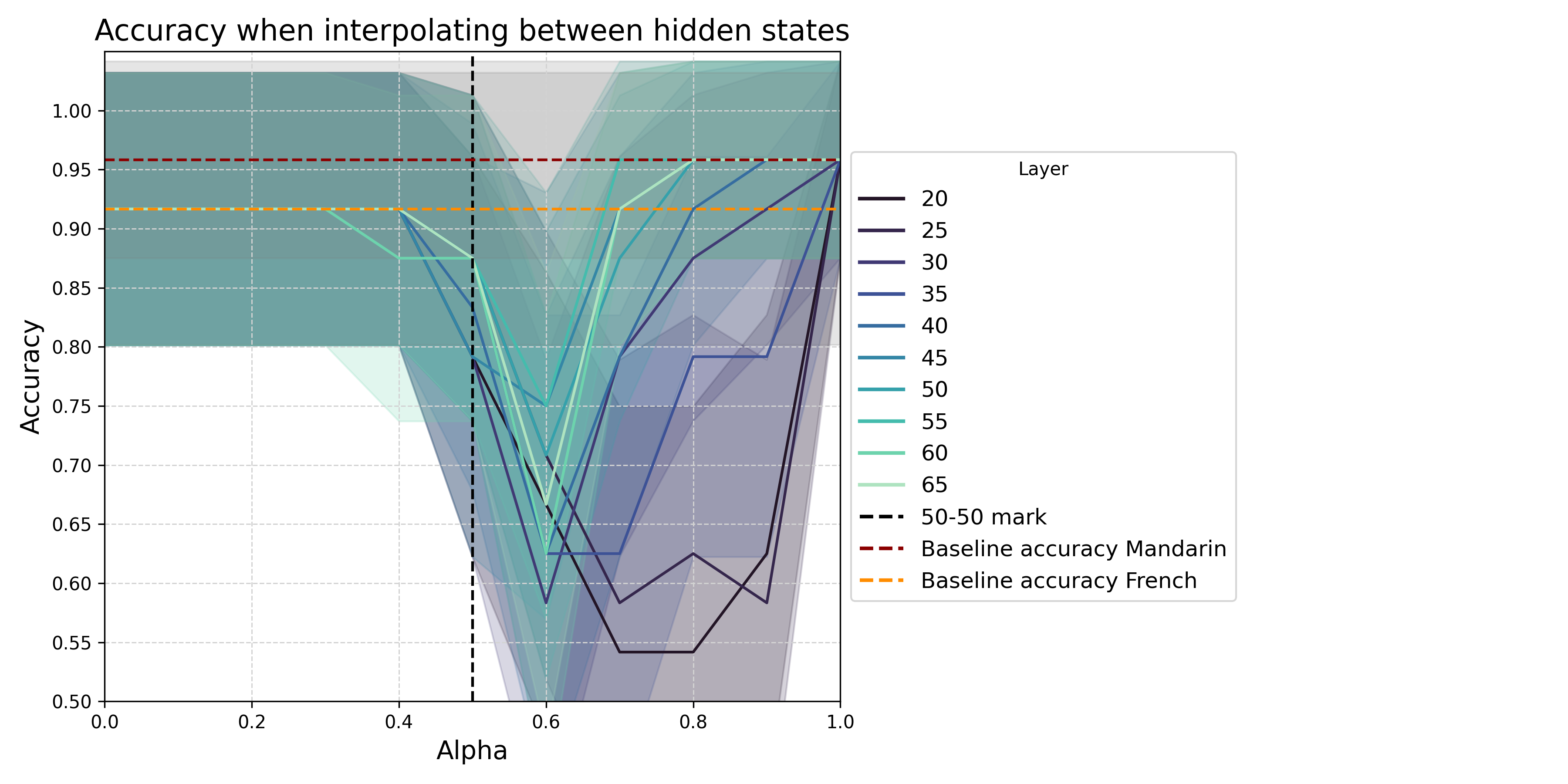} 
\end{minipage}
\begin{minipage}{0.49\textwidth}
    \centering
    \includegraphics[trim={1.5cm 0.5cm 1.5cm 1.5cm},clip,width=\textwidth]{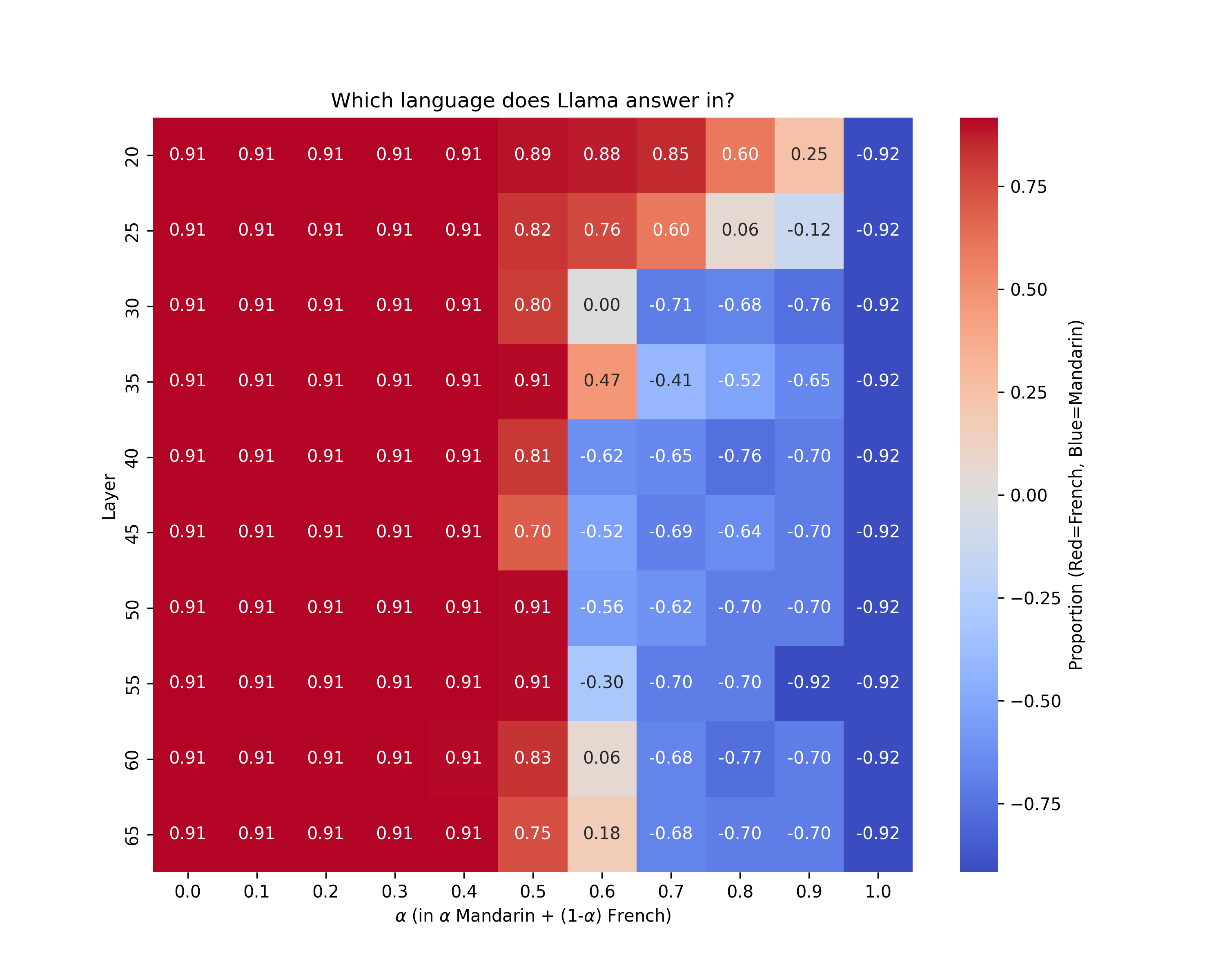} 
\end{minipage}
\caption{Hidden state interpolation between French prompts, and Mandarin prompts in \llama. Left shows the accuracy (i.e., the proportion of times the model correctly outputs city in either language). Right shows the propensity of the model to answer in French (red) and Mandarin (blue). }
\end{figure}

\begin{figure}[h]
\begin{minipage}{0.49\textwidth}
    \centering
    \includegraphics[trim={0 0 5cm 0},clip, width=\textwidth]{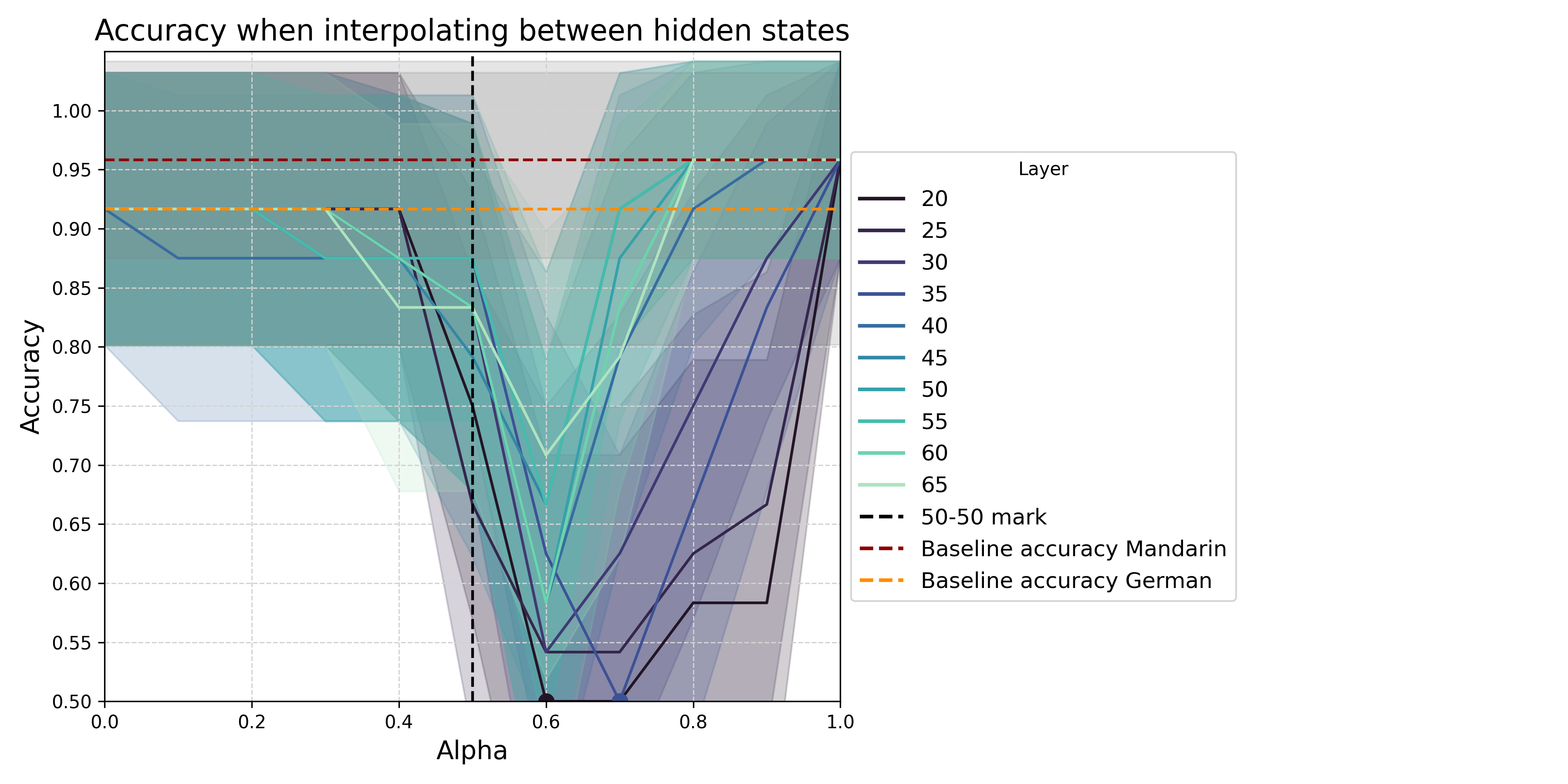} 
\end{minipage}
\begin{minipage}{0.49\textwidth}
    \centering
    \includegraphics[trim={1.5cm 0.5cm 1.5cm 1.5cm},clip,width=\textwidth]{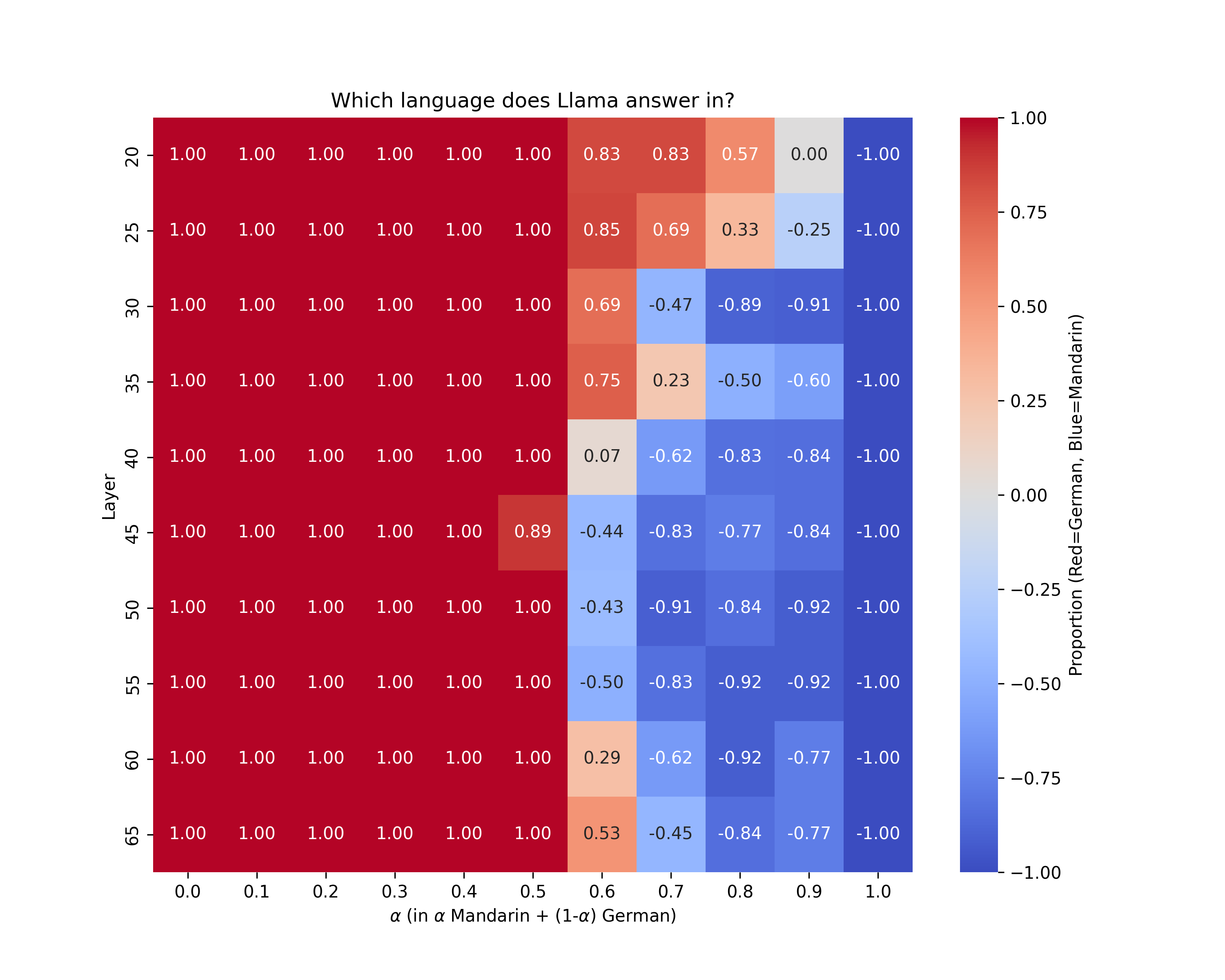} 
\end{minipage}
\caption{Hidden state interpolation between German prompts, and Mandarin prompts in \llama. Left shows the accuracy (i.e., the proportion of times the model correctly outputs city in either language). Right shows the propensity of the model to answer in German (red) and Mandarin (blue). }
\end{figure}

\FloatBarrier
\newpage 

\subsubsection{Mistral}
\mistral \ is most likely to answer in English and least likely to answer in Dutch. It is roughly equally likely to answer in German and French.

\begin{figure}[h]
\begin{minipage}{0.49\textwidth}
    \centering
    \includegraphics[trim={0 0 5cm 0},clip, width=\textwidth]{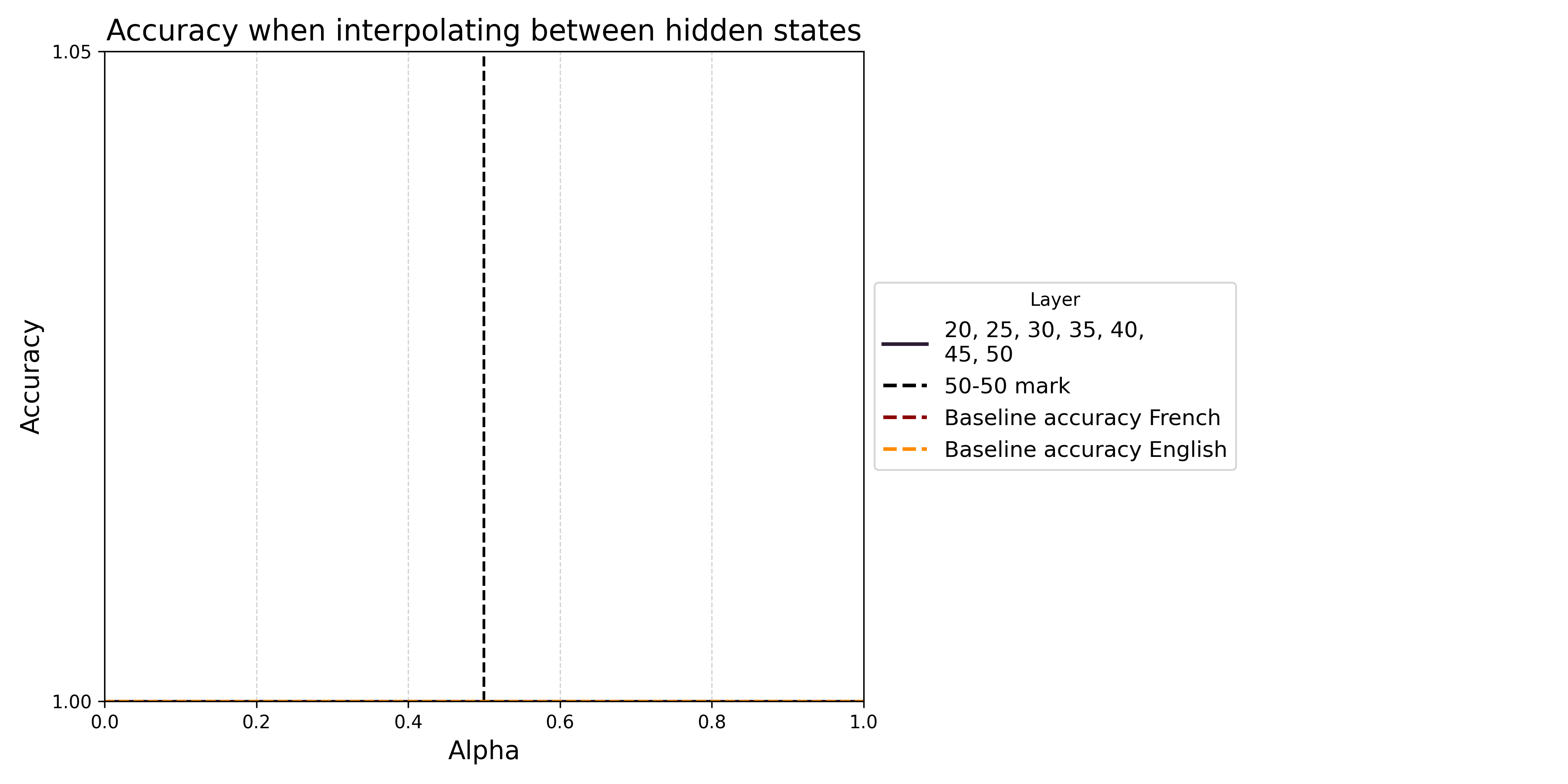} 
\end{minipage}
\begin{minipage}{0.49\textwidth}
    \centering
    \includegraphics[trim={1.5cm 0.5cm 1.5cm 1.5cm},clip,width=\textwidth]{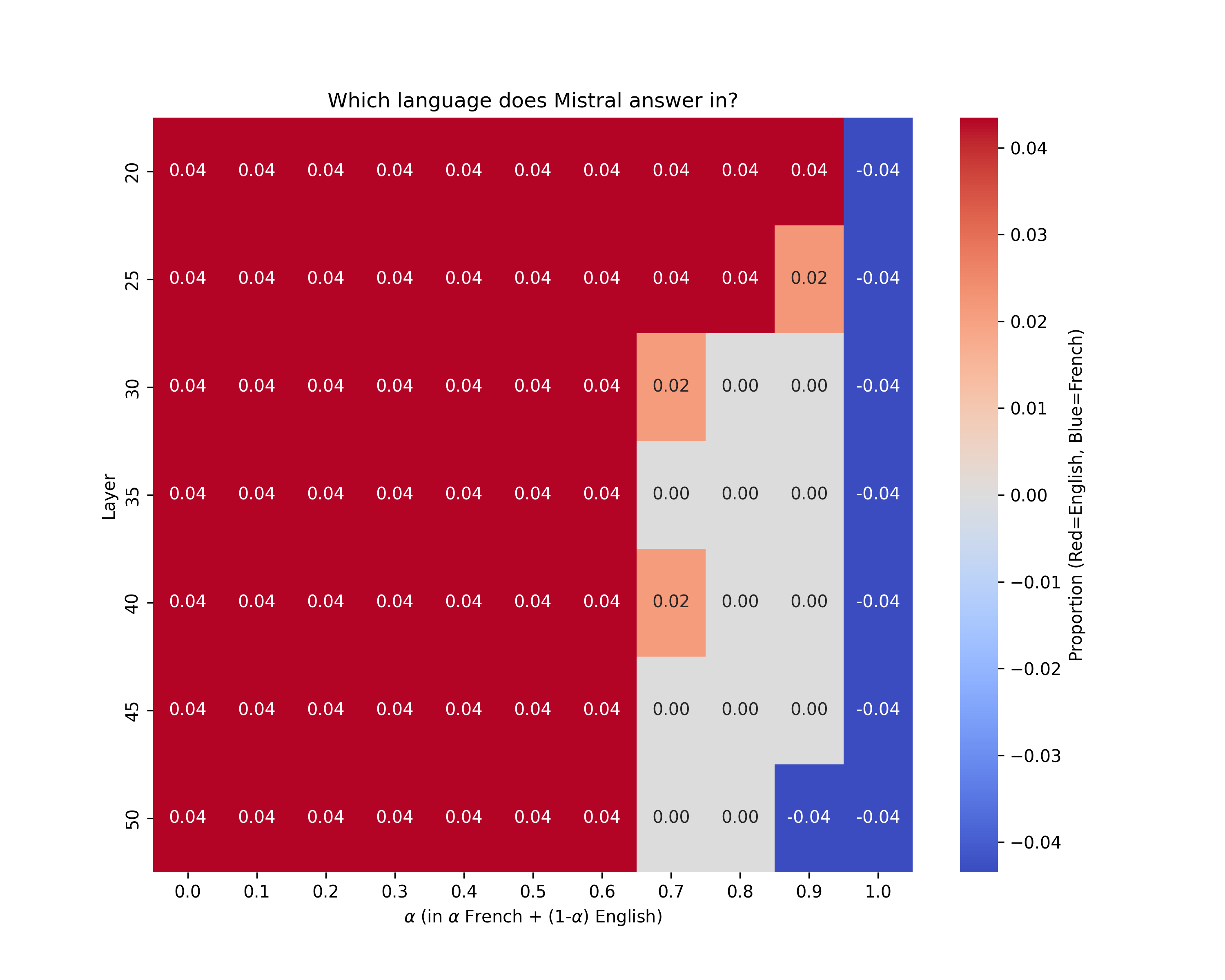} 
\end{minipage}
\caption{Hidden state interpolation between English prompts, and French prompts in \mistral. Left shows the accuracy (i.e., the proportion of times the model correctly outputs city in either language). Right shows the propensity of the model to answer in English (red) and French (blue). }
\end{figure}

\begin{figure}[h]
\begin{minipage}{0.49\textwidth}
    \centering
    \includegraphics[trim={0 0 5cm 0},clip, width=\textwidth]{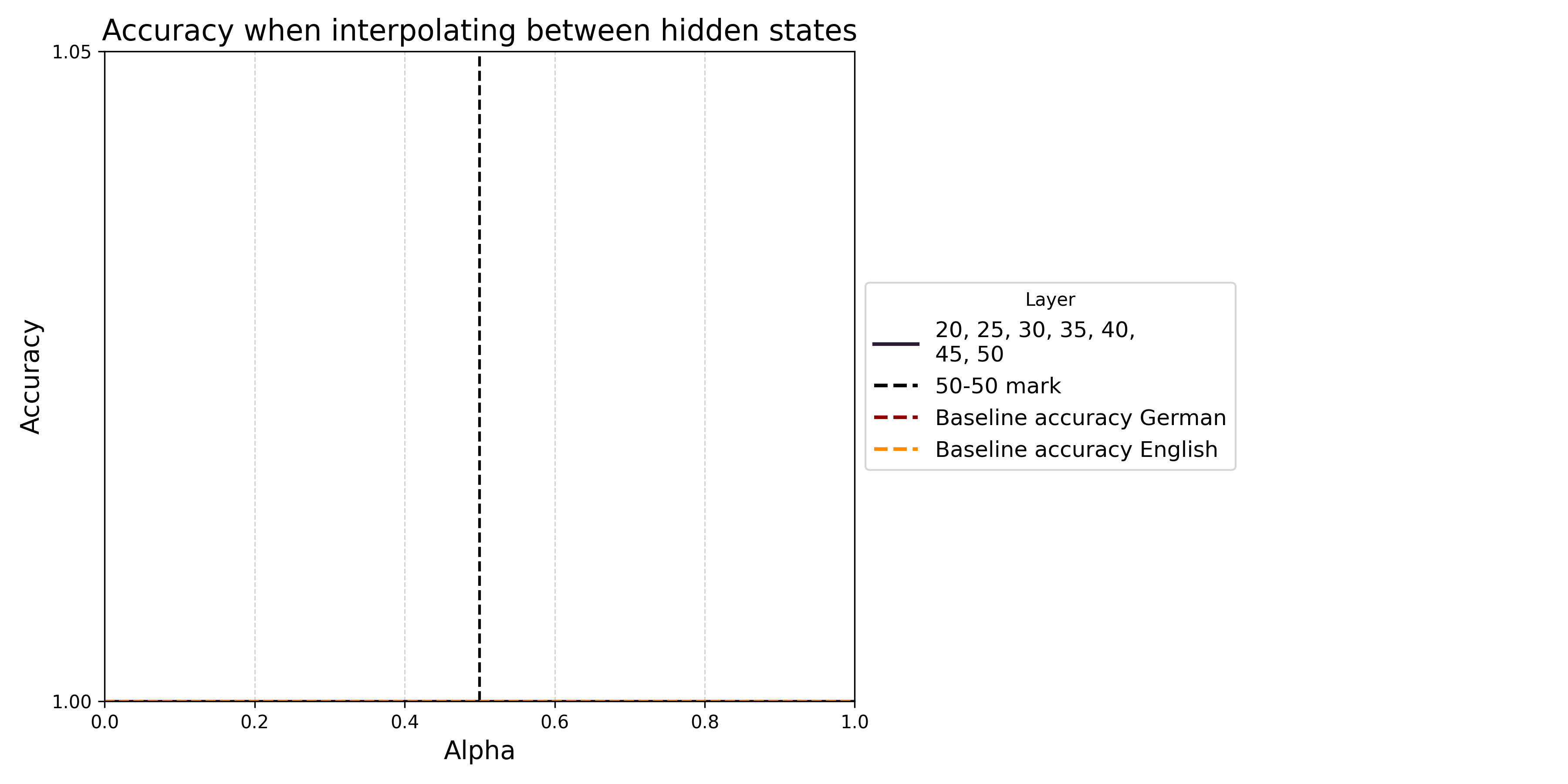} 
\end{minipage}
\begin{minipage}{0.49\textwidth}
    \centering
    \includegraphics[trim={1.5cm 0.5cm 1.5cm 1.5cm},clip,width=\textwidth]{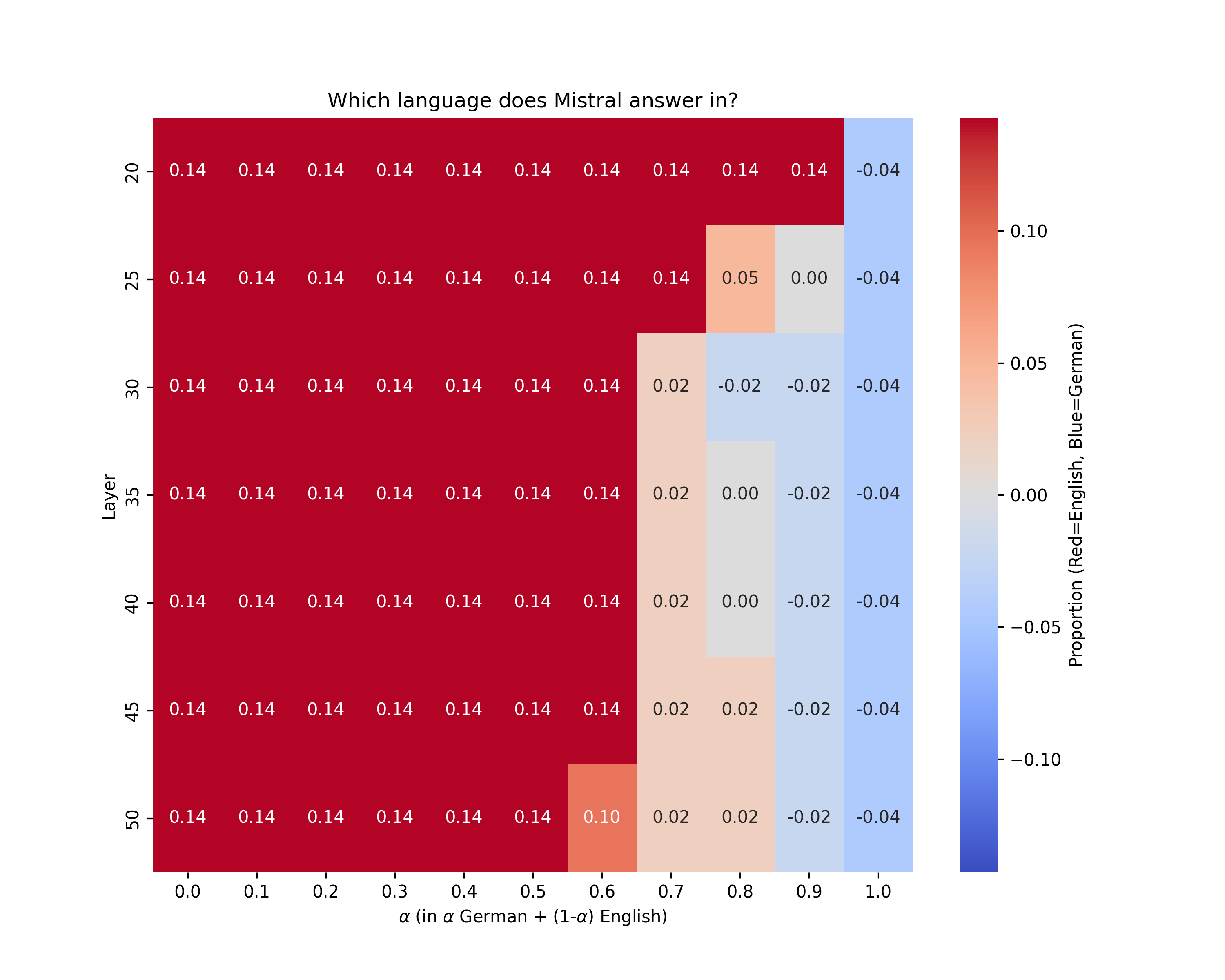} 
\end{minipage}
\caption{Hidden state interpolation between English prompts, and German prompts in \mistral. Left shows the accuracy (i.e., the proportion of times the model correctly outputs city in either language). Right shows the propensity of the model to answer in English (red) and German (blue). }
\end{figure}

\begin{figure}[h]
\begin{minipage}{0.49\textwidth}
    \centering
    \includegraphics[trim={0 0 5cm 0},clip, width=\textwidth]{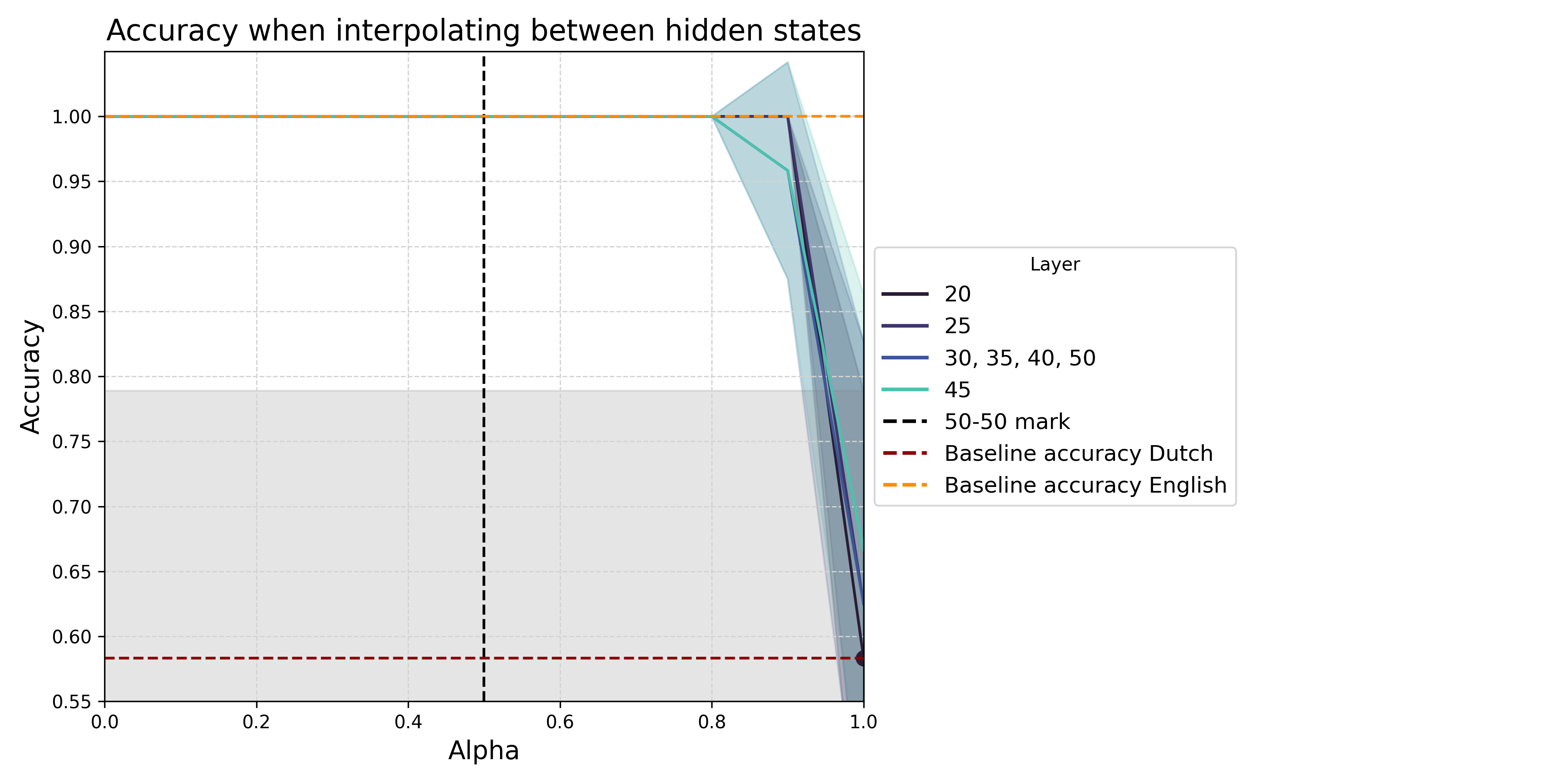} 
\end{minipage}
\begin{minipage}{0.49\textwidth}
    \centering
    \includegraphics[trim={1.5cm 0.5cm 1.5cm 1.5cm},clip,width=\textwidth]{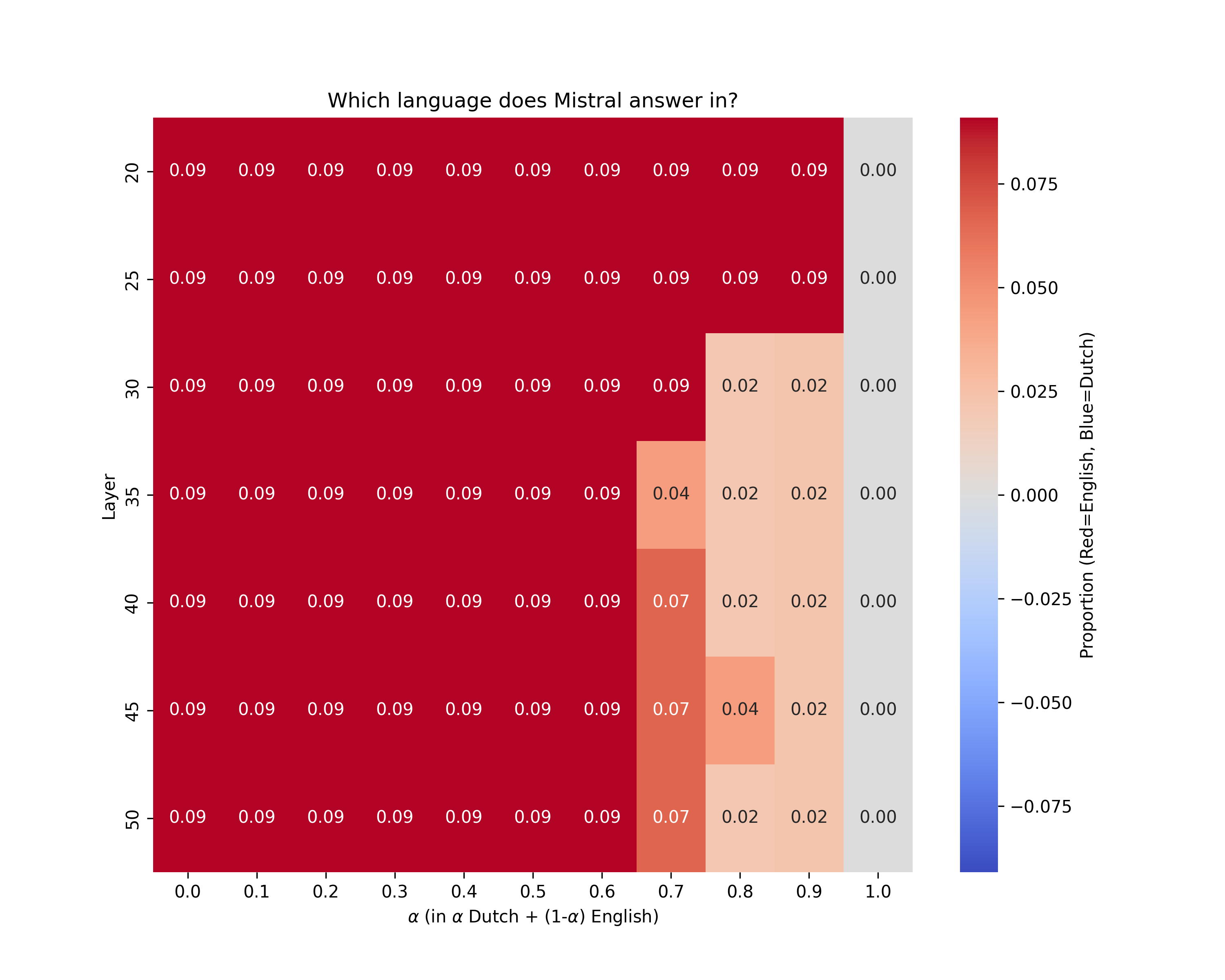} 
\end{minipage}
\caption{Hidden state interpolation between English prompts, and Dutch prompts in \mistral. Left shows the accuracy (i.e., the proportion of times the model correctly outputs city in either language). Right shows the propensity of the model to answer in English (red) and Dutch (blue). }
\end{figure}

\begin{figure}[h]
\begin{minipage}{0.49\textwidth}
    \centering
    \includegraphics[trim={0 0 5cm 0},clip, width=\textwidth]{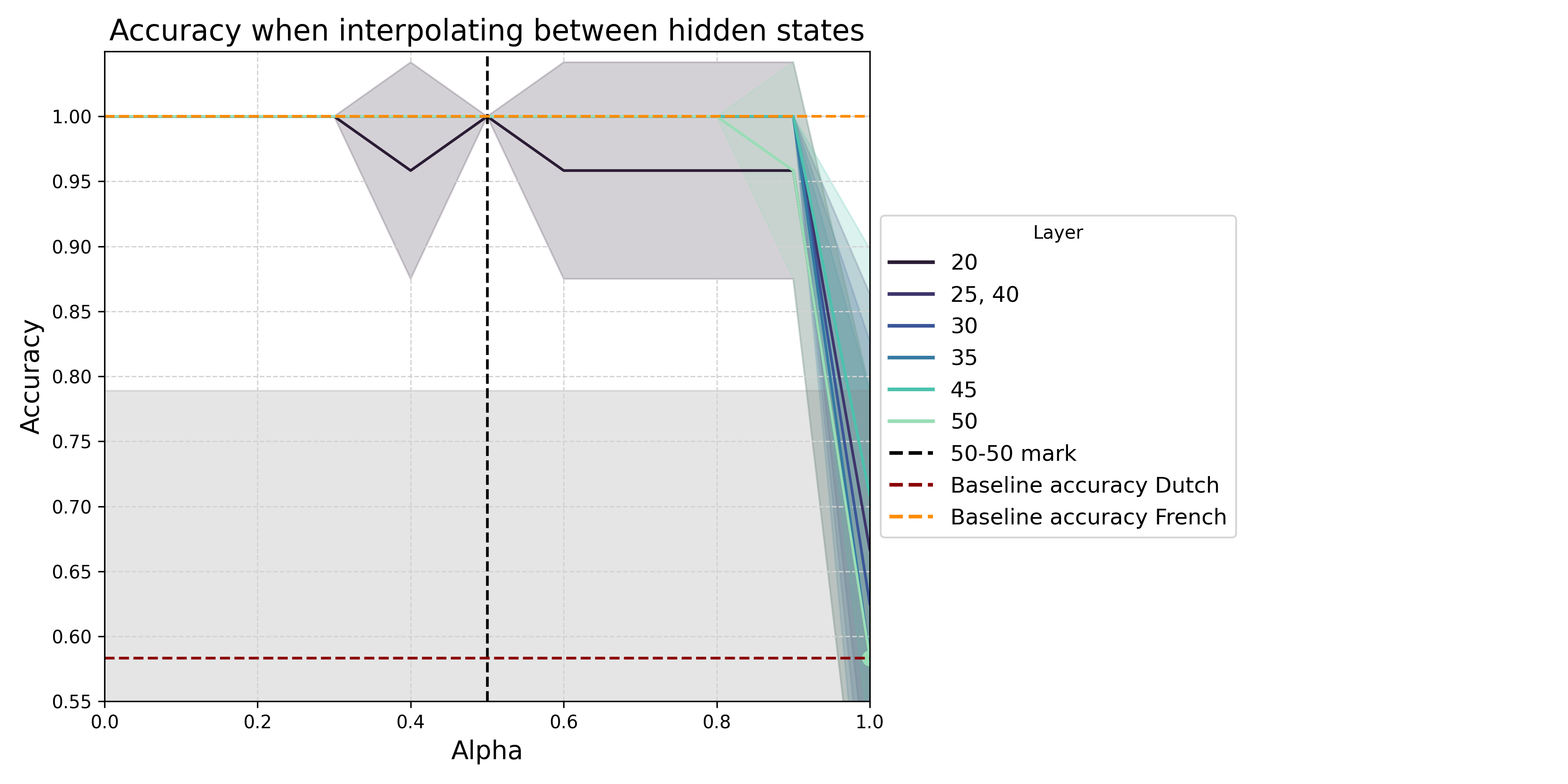} 
\end{minipage}
\begin{minipage}{0.49\textwidth}
    \centering
    \includegraphics[trim={1.5cm 0.5cm 1.5cm 1.5cm},clip,width=\textwidth]{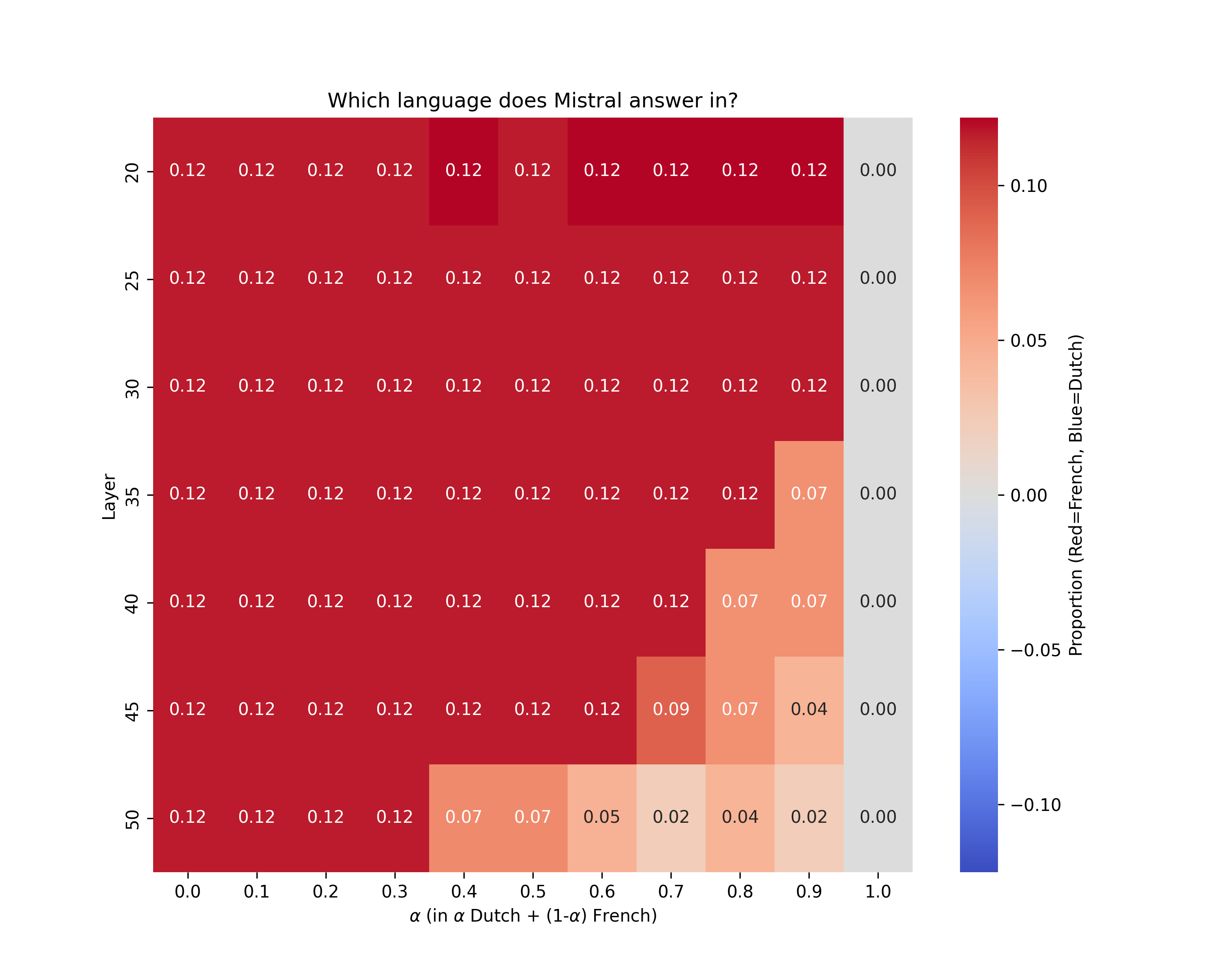} 
\end{minipage}
\caption{Hidden state interpolation between Dutch prompts, and French prompts in \mistral. Left shows the accuracy (i.e., the proportion of times the model correctly outputs city in either language). Right shows the propensity of the model to answer in French (red) and Dutch (blue). }
\end{figure}

\begin{figure}[h]
\begin{minipage}{0.49\textwidth}
    \centering
    \includegraphics[trim={0 0 5cm 0},clip, width=\textwidth]{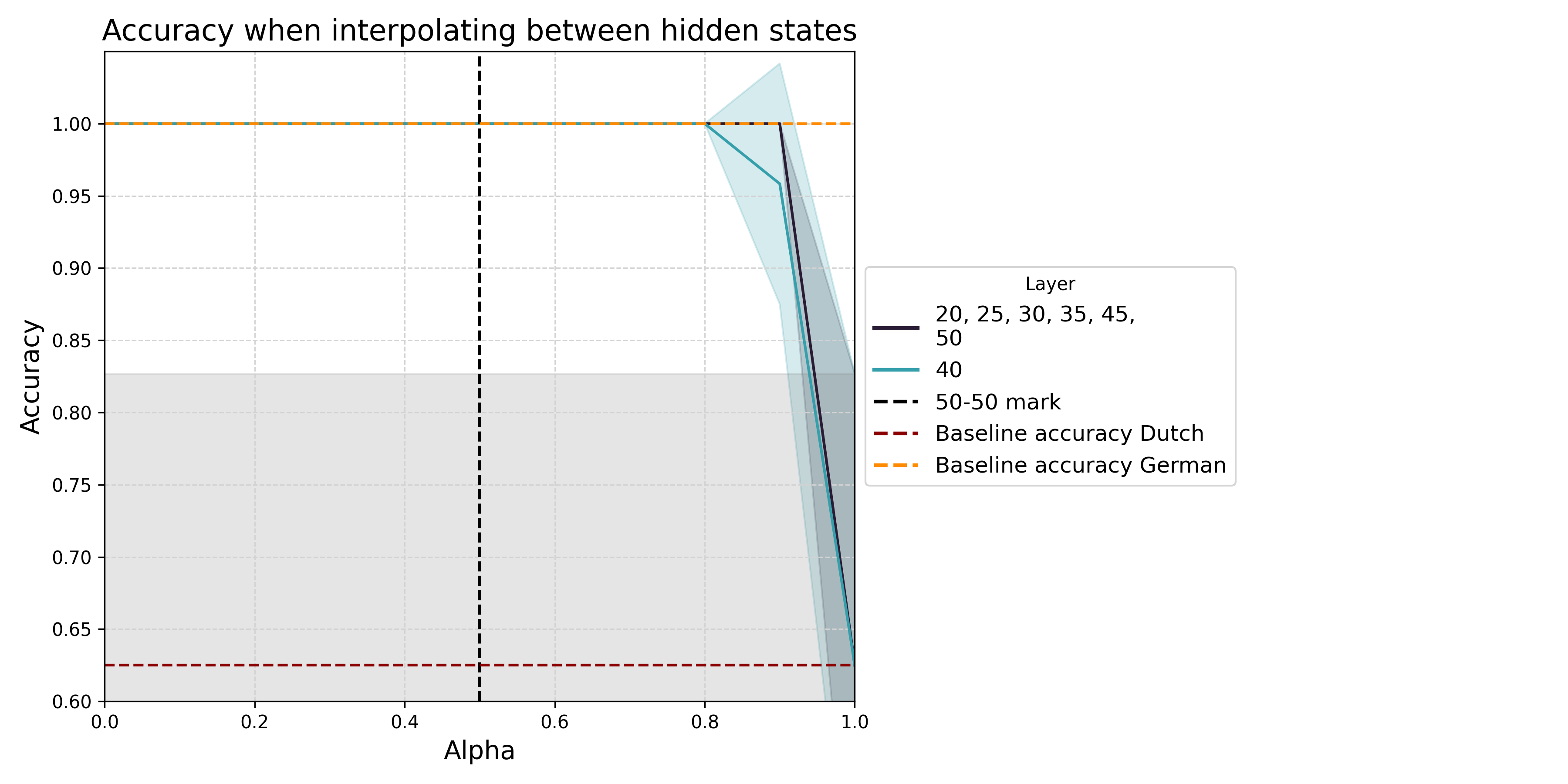} 
\end{minipage}
\begin{minipage}{0.49\textwidth}
    \centering
    \includegraphics[trim={1.5cm 0.5cm 1.5cm 1.5cm},clip,width=\textwidth]{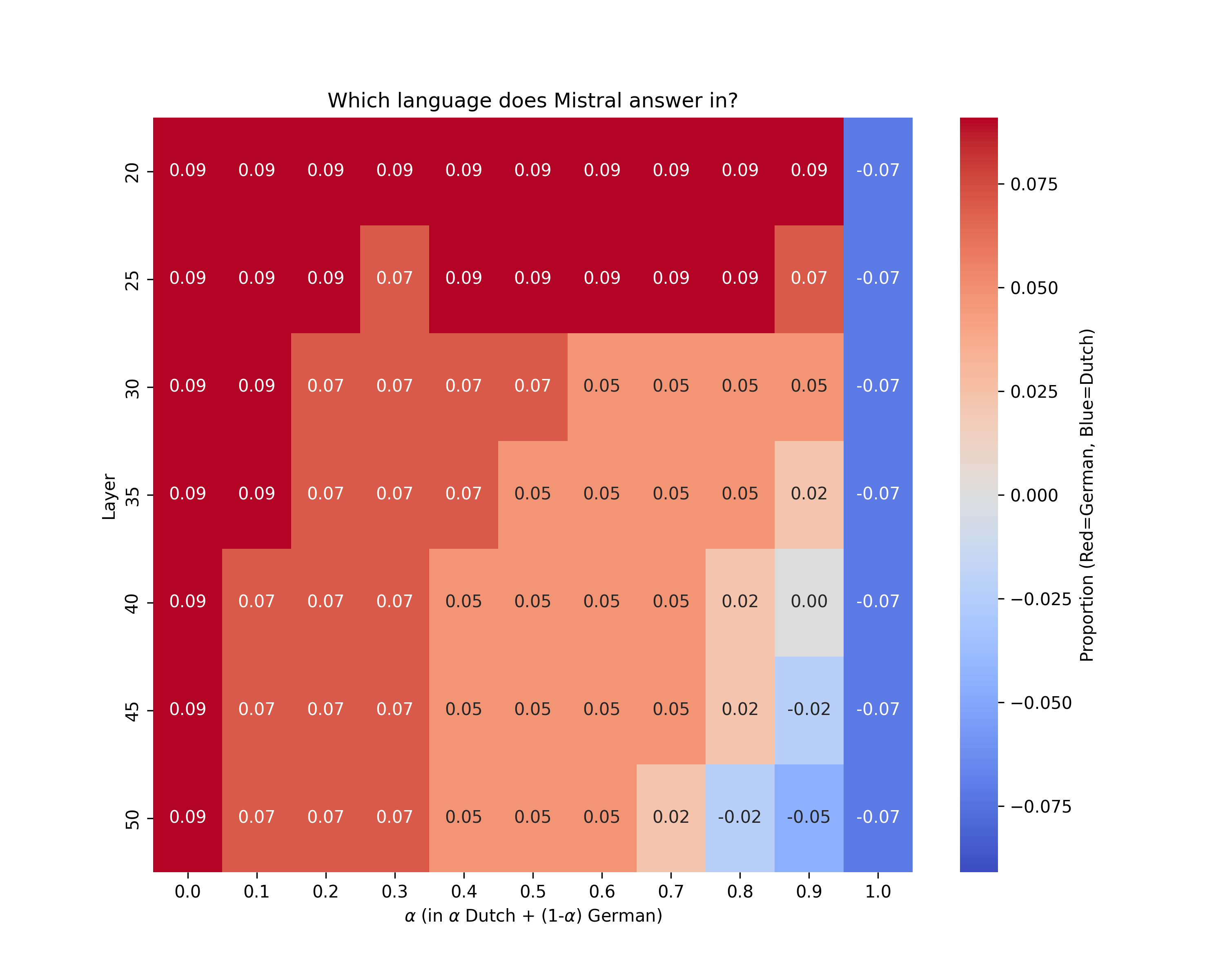} 
\end{minipage}
\caption{Hidden state interpolation between Dutch prompts, and German prompts in \mistral. Left shows the accuracy (i.e., the proportion of times the model correctly outputs city in either language). Right shows the propensity of the model to answer in German (red) and Dutch (blue). }
\end{figure}

\begin{figure}[h]
\begin{minipage}{0.49\textwidth}
    \centering
    \includegraphics[trim={0 0 5cm 0},clip, width=\textwidth]{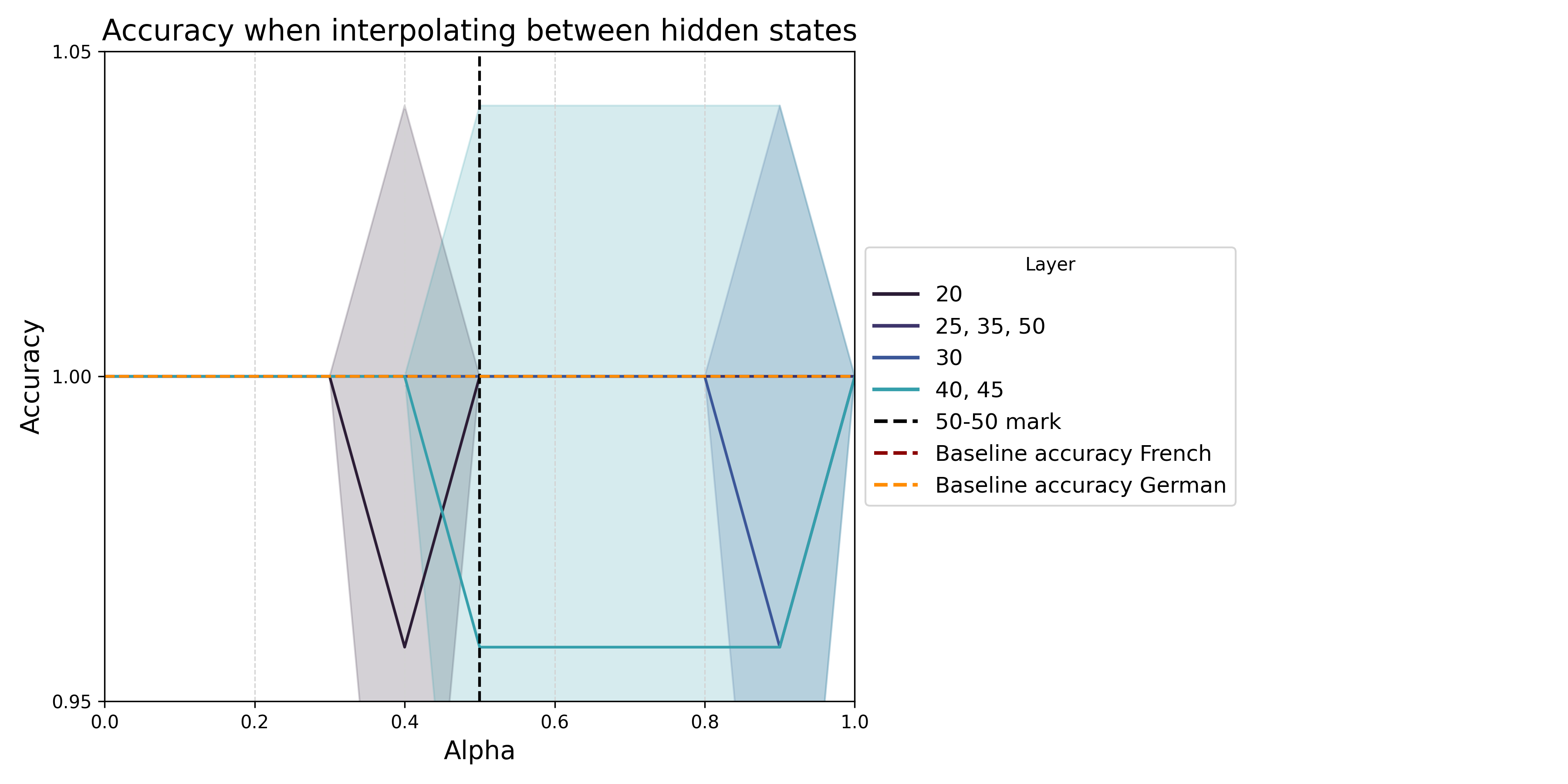} 
\end{minipage}
\begin{minipage}{0.49\textwidth}
    \centering
    \includegraphics[trim={1.5cm 0.5cm 1.5cm 1.5cm},clip,width=\textwidth]{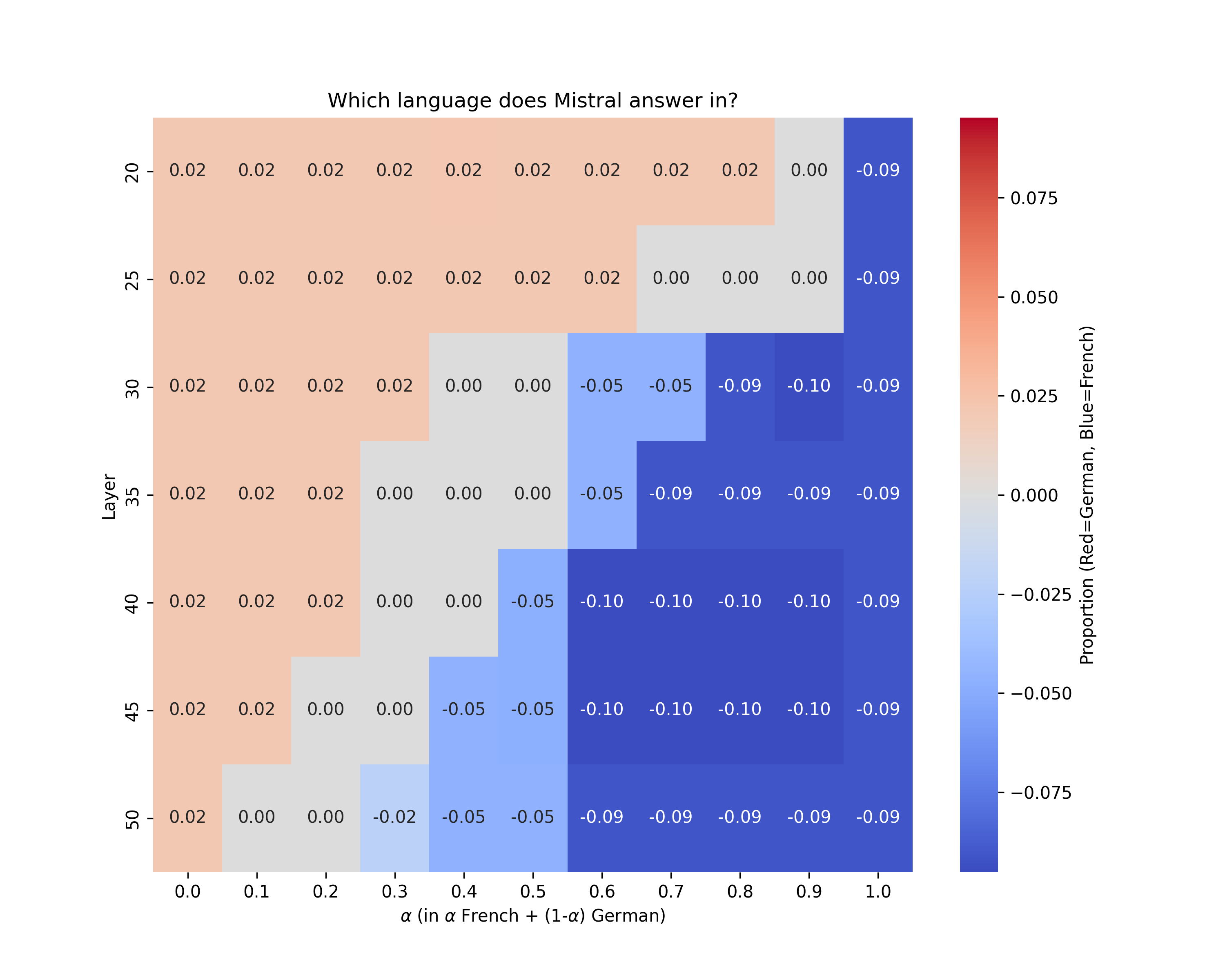} 
\end{minipage}
\caption{Hidden state interpolation between French prompts, and German prompts in \mistral. Left shows the accuracy (i.e., the proportion of times the model correctly outputs city in either language). Right shows the propensity of the model to answer in German (red) and French (blue). }
\end{figure}

\begin{figure}[h]
\begin{minipage}{0.49\textwidth}
    \centering
    \includegraphics[trim={0 0 5cm 0},clip, width=\textwidth]{figures/interpolate/FR_GER_mistral__interpolate_results.png} 
\end{minipage}
\begin{minipage}{0.49\textwidth}
    \centering
    \includegraphics[trim={1.5cm 0.5cm 1.5cm 1.5cm},clip,width=\textwidth]{figures/interpolate/FR_GER_mistral_heatmap_interpolate_results.png} 
\end{minipage}
\caption{Hidden state interpolation between French prompts, and German prompts in \mistral. Left shows the accuracy (i.e., the proportion of times the model correctly outputs city in either language). Right shows the propensity of the model to answer in German (red) and French (blue). }
\end{figure}

\begin{figure}[h]
\begin{minipage}{0.49\textwidth}
    \centering
    \includegraphics[trim={0 0 5cm 0},clip, width=\textwidth]{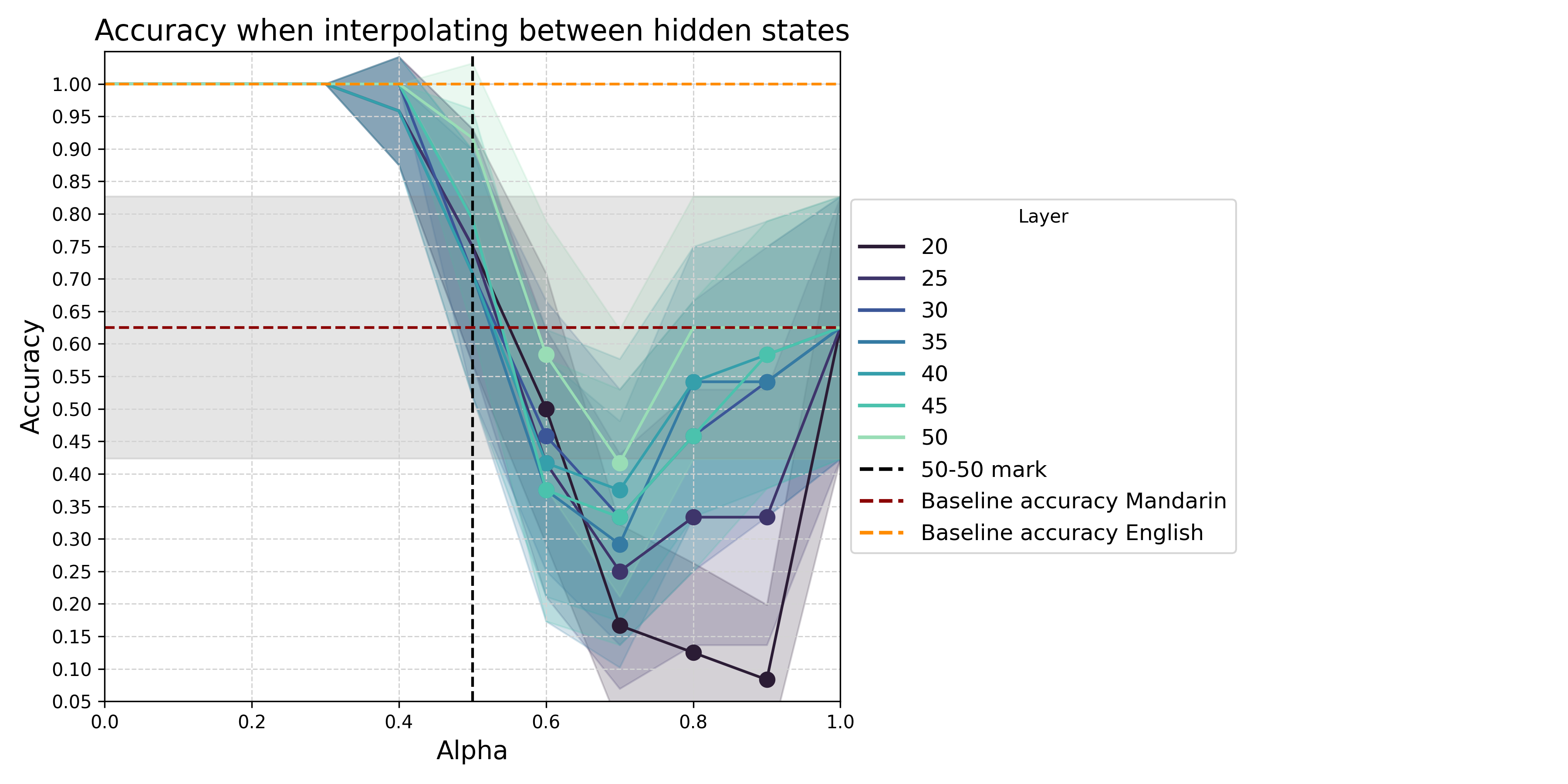} 
\end{minipage}
\begin{minipage}{0.49\textwidth}
    \centering
    \includegraphics[trim={1.5cm 0.5cm 1.5cm 1.5cm},clip,width=\textwidth]{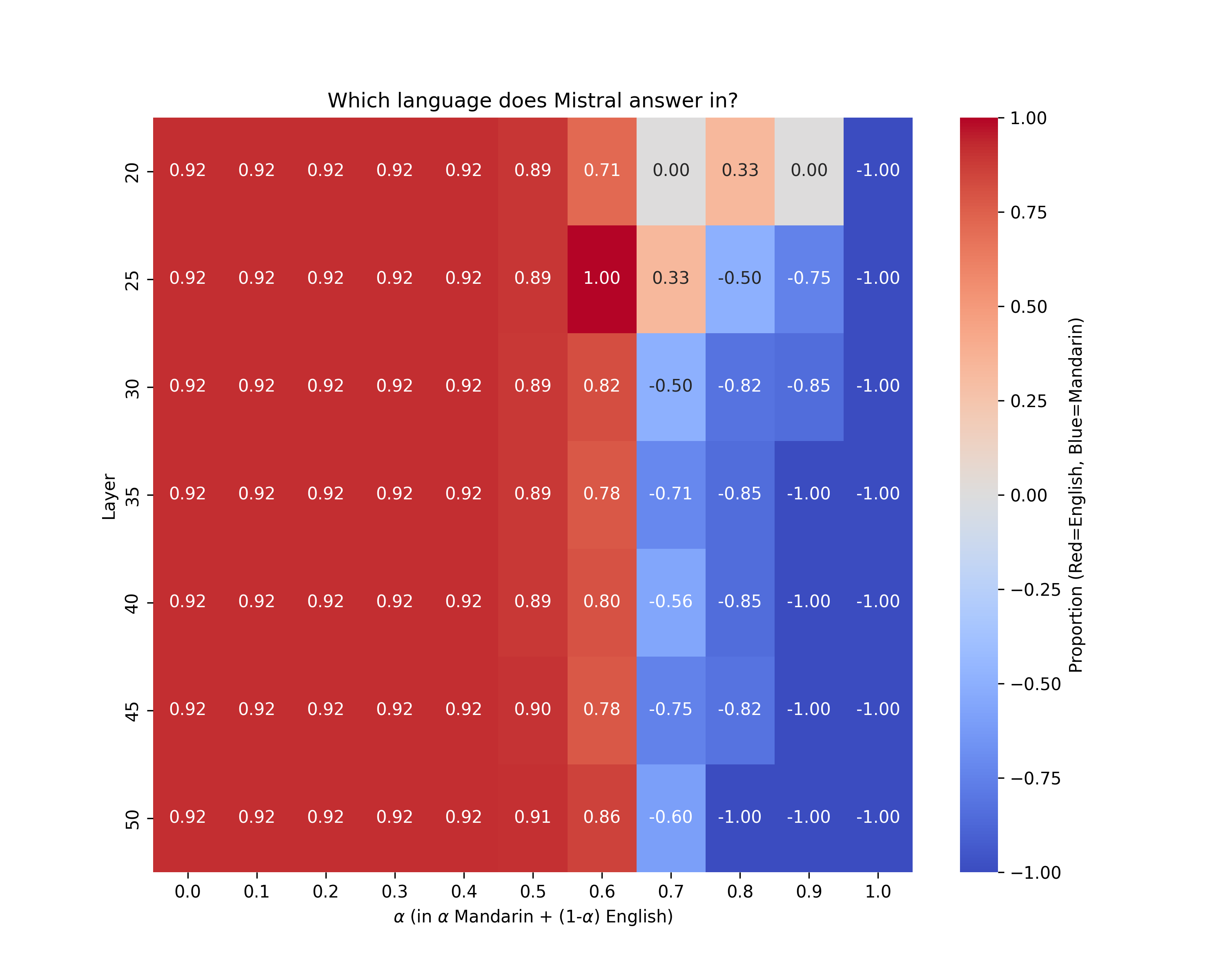} 
\end{minipage}
\caption{Hidden state interpolation between English prompts, and Mandarin prompts in \mistral. Left shows the accuracy (i.e., the proportion of times the model correctly outputs city in either language). Right shows the propensity of the model to answer in English (red) and Mandarin (blue). }
\end{figure}

\begin{figure}[h]
\begin{minipage}{0.49\textwidth}
    \centering
    \includegraphics[trim={0 0 5cm 0},clip, width=\textwidth]{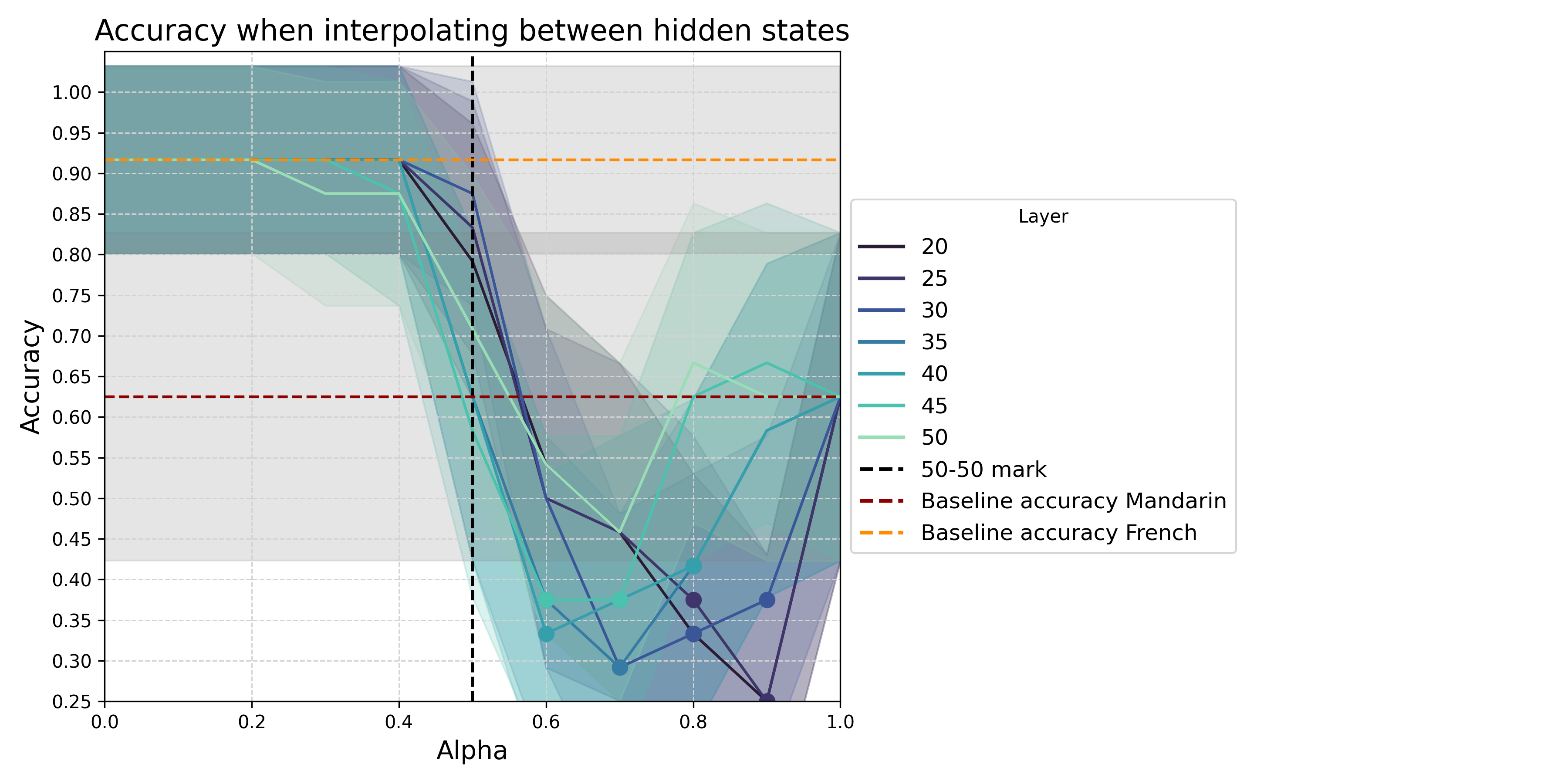} 
\end{minipage}
\begin{minipage}{0.49\textwidth}
    \centering
    \includegraphics[trim={1.5cm 0.5cm 1.5cm 1.5cm},clip,width=\textwidth]{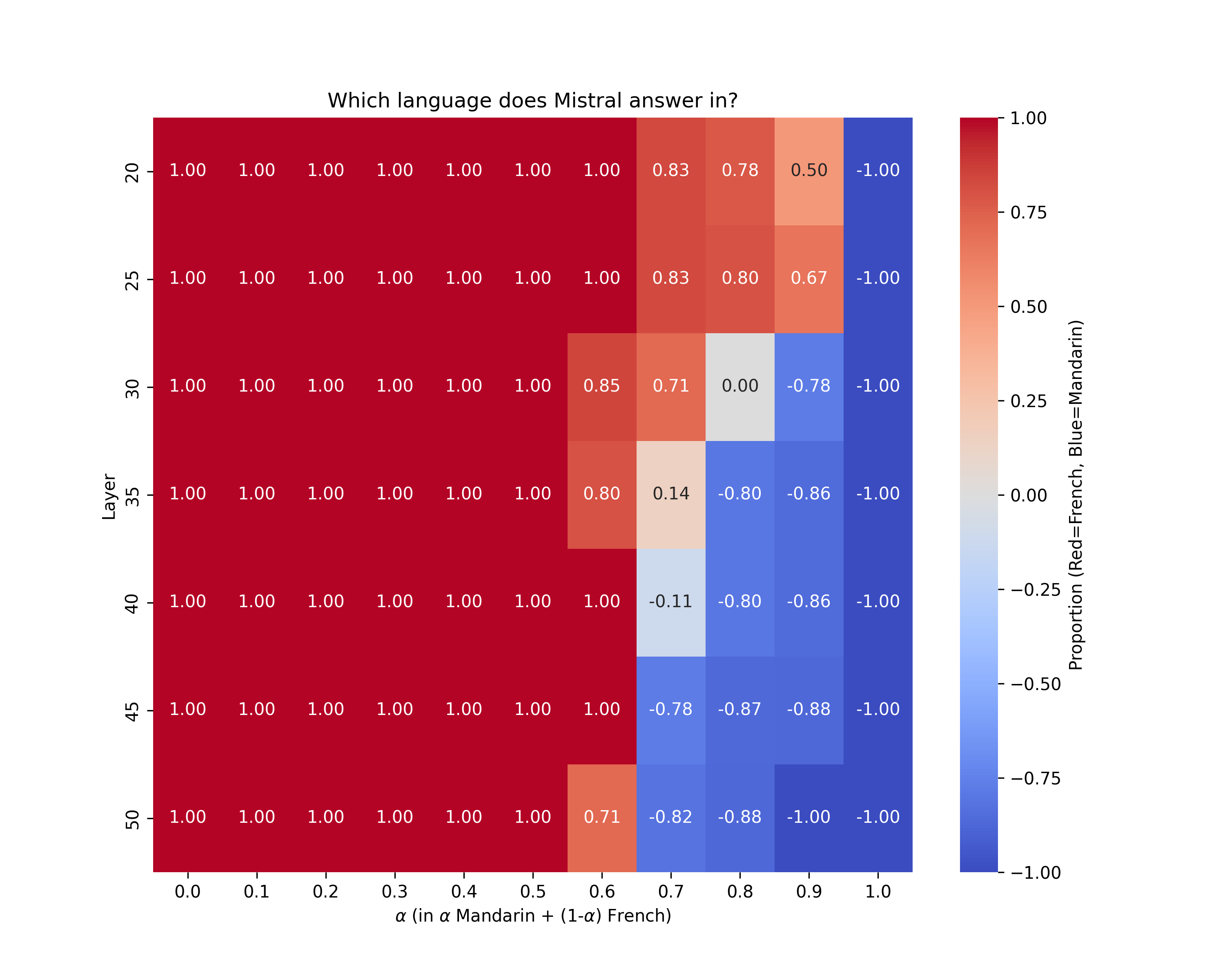} 
\end{minipage}
\caption{Hidden state interpolation between French prompts, and Mandarin prompts in \mistral. Left shows the accuracy (i.e., the proportion of times the model correctly outputs city in either language). Right shows the propensity of the model to answer in French (red) and Mandarin (blue). }
\end{figure}

\begin{figure}[h]
\begin{minipage}{0.49\textwidth}
    \centering
    \includegraphics[trim={0 0 5cm 0},clip, width=\textwidth]{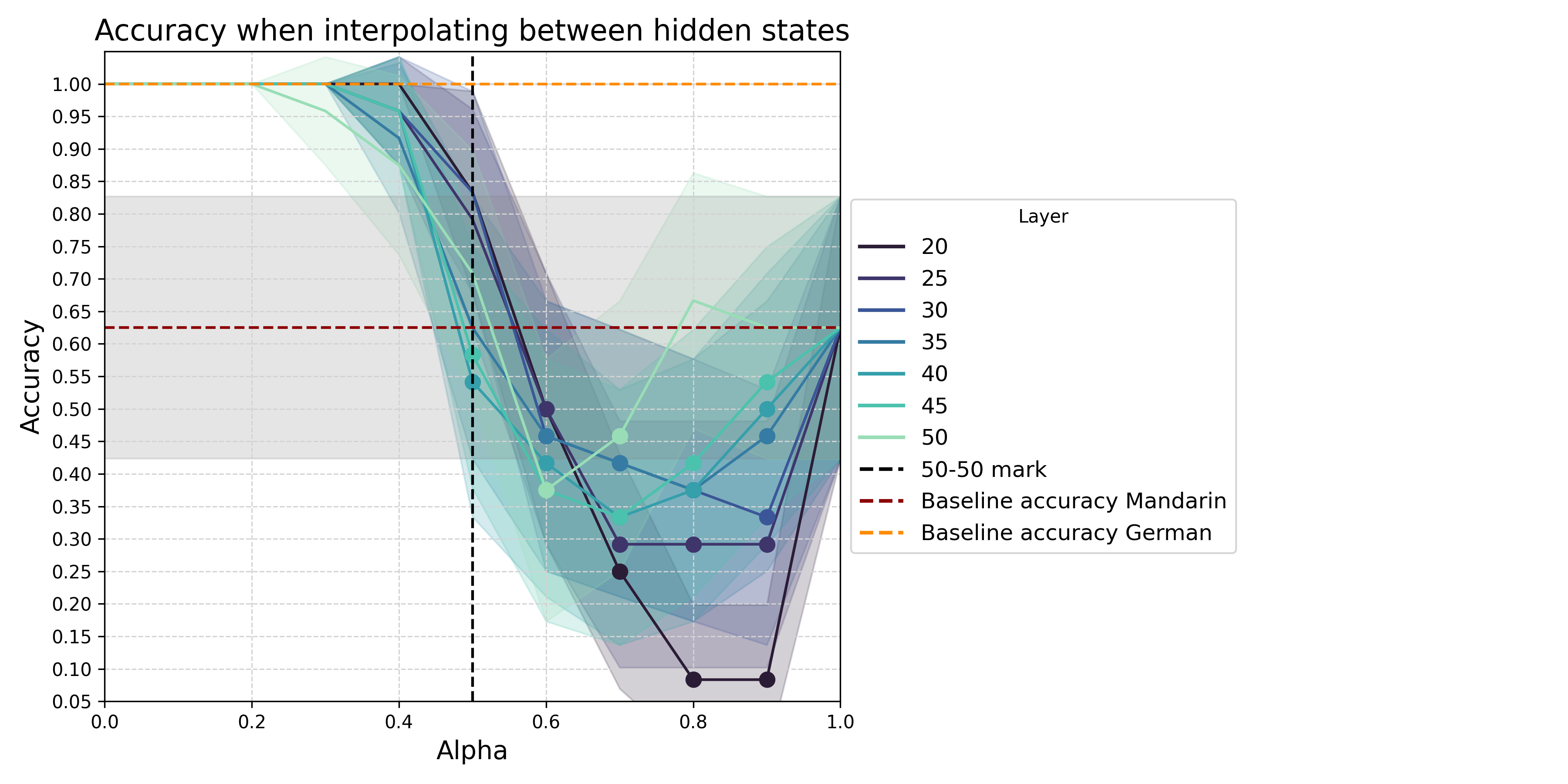} 
\end{minipage}
\begin{minipage}{0.49\textwidth}
    \centering
    \includegraphics[trim={1.5cm 0.5cm 1.5cm 1.5cm},clip,width=\textwidth]{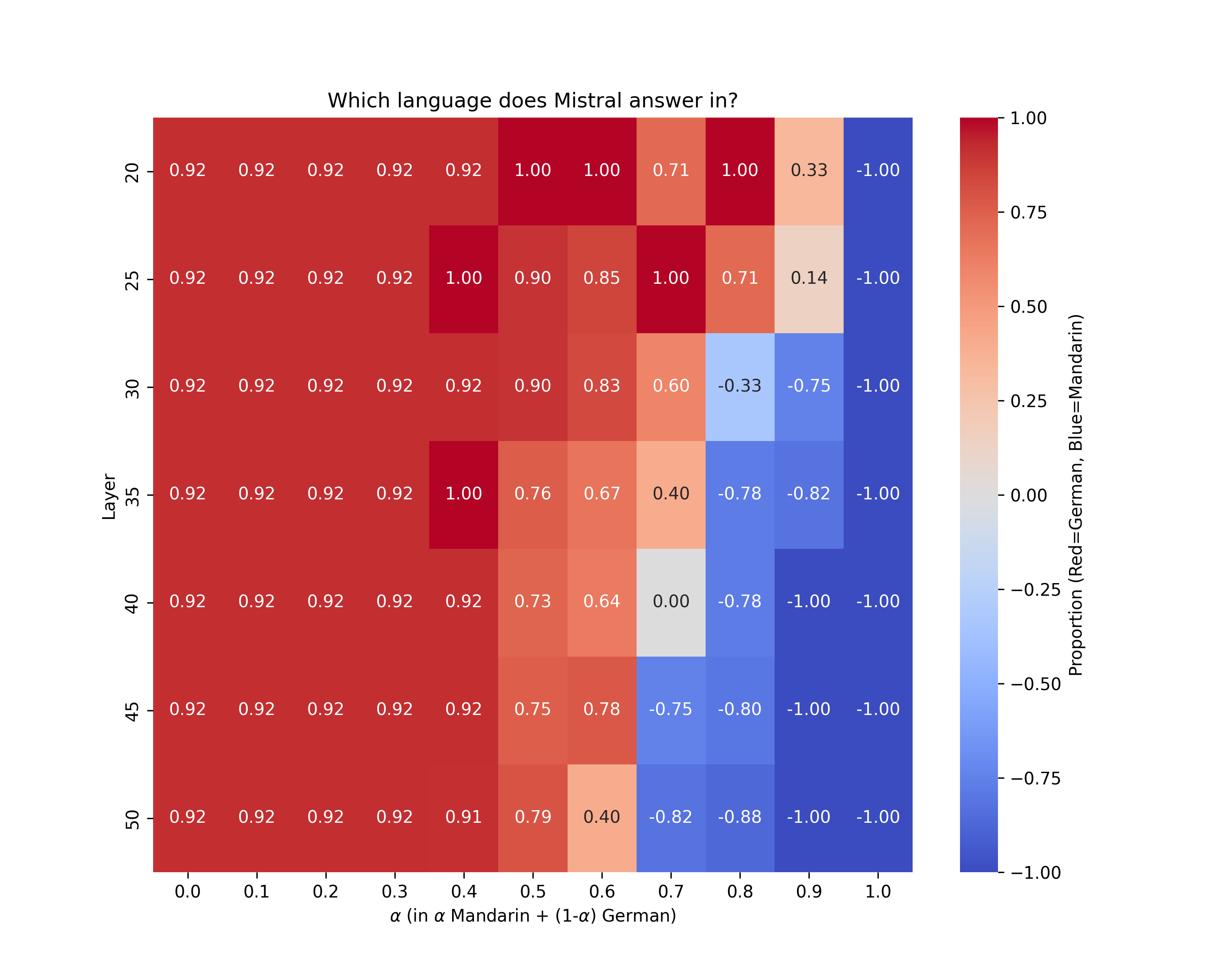} 
\end{minipage}
\caption{Hidden state interpolation between German prompts, and Mandarin prompts in \mistral. Left shows the accuracy (i.e., the proportion of times the model correctly outputs city in either language). Right shows the propensity of the model to answer in German (red) and Mandarin (blue). }
\end{figure}

\begin{figure}[h]
\begin{minipage}{0.49\textwidth}
    \centering
    \includegraphics[trim={0 0 5cm 0},clip, width=\textwidth]{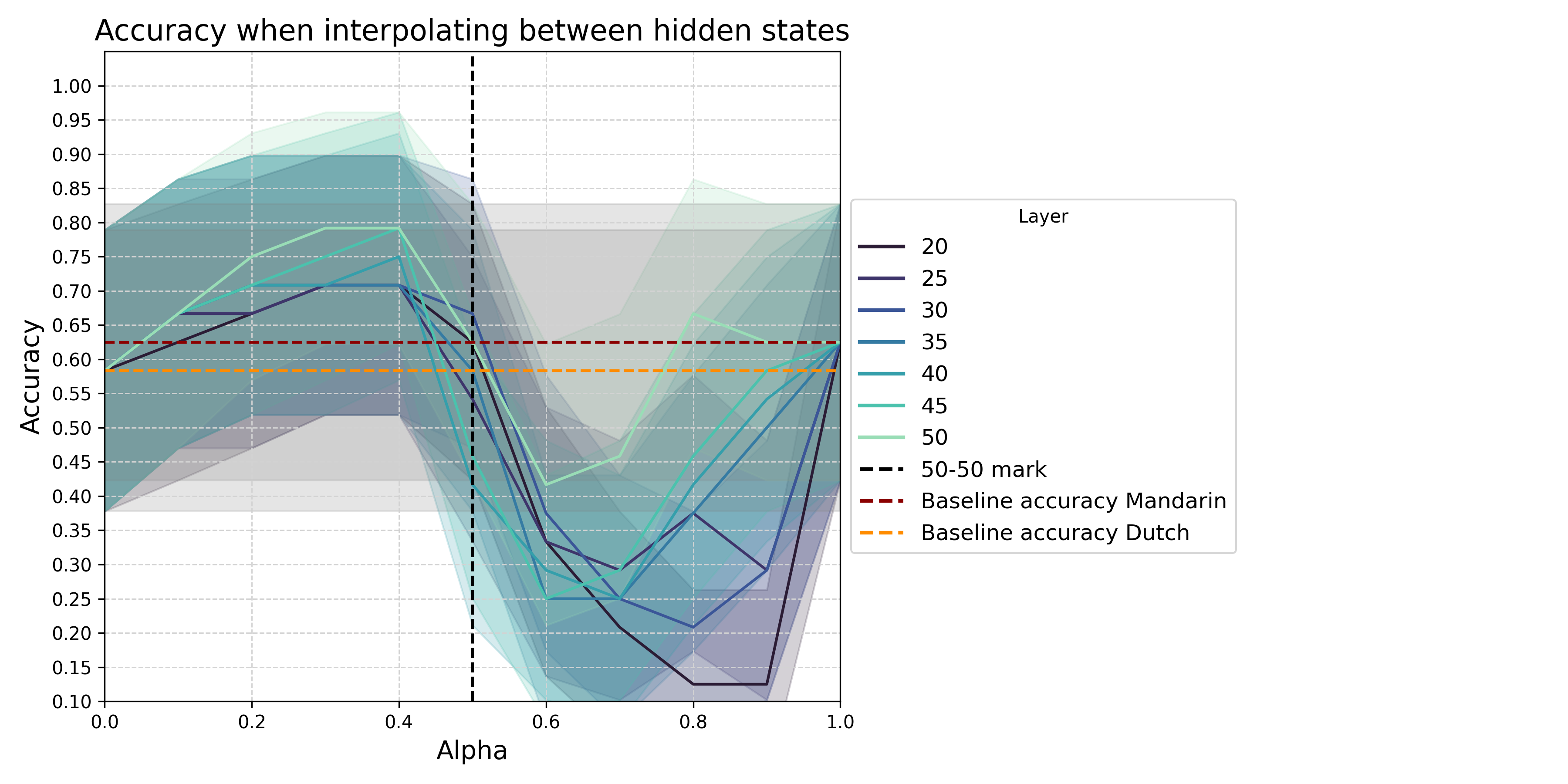} 
\end{minipage}
\begin{minipage}{0.49\textwidth}
    \centering
    \includegraphics[trim={1.5cm 0.5cm 1.5cm 1.5cm},clip,width=\textwidth]{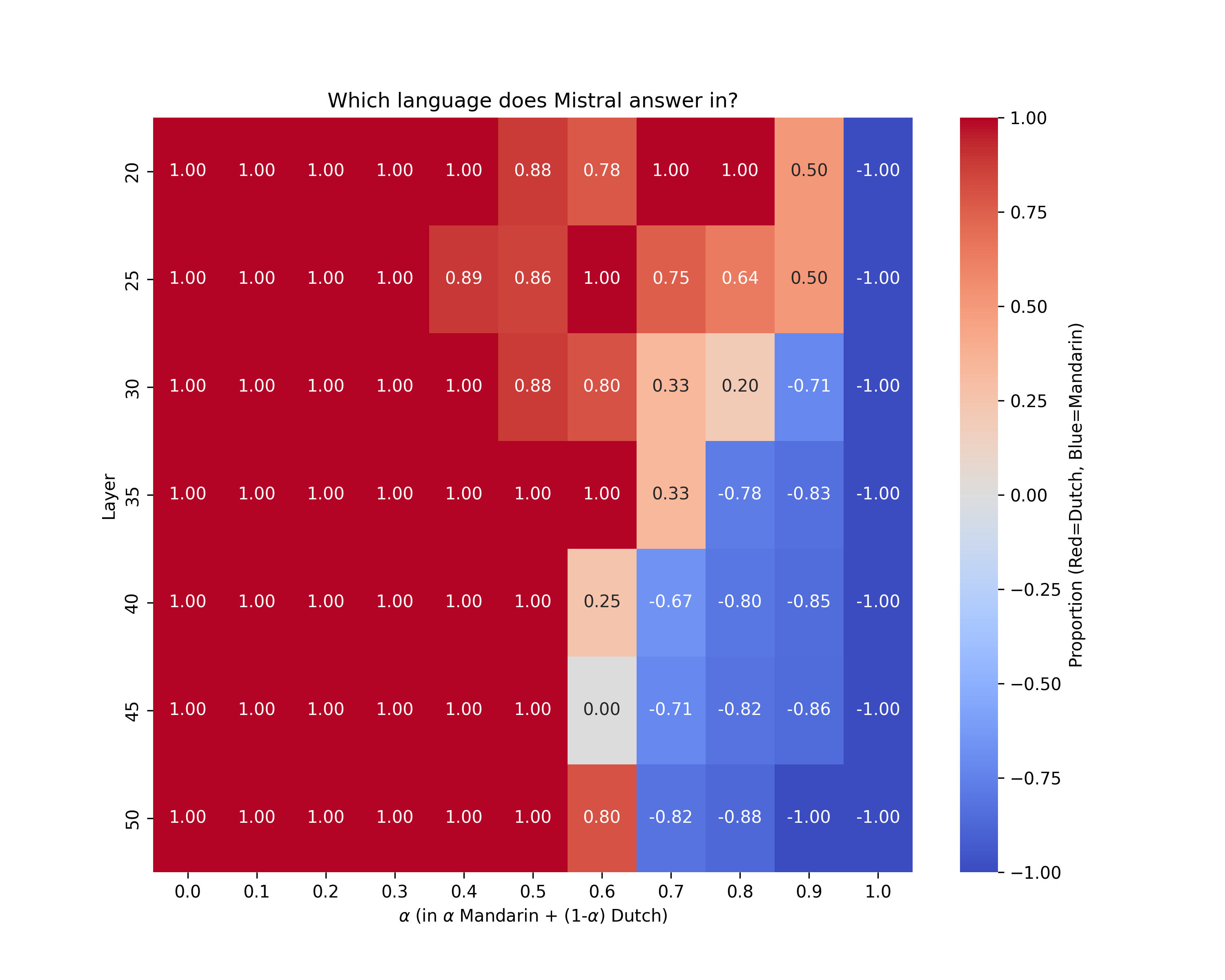} 
\end{minipage}
\caption{Hidden state interpolation between Dutch prompts, and Mandarin prompts in \mistral. Left shows the accuracy (i.e., the proportion of times the model correctly outputs city in either language). Right shows the propensity of the model to answer in Dutch (red) and Mandarin (blue). }
\end{figure}

\FloatBarrier
\newpage 
\subsubsection{Gemma} 

\gemma \ is most likely to answer in English, German, French and then Dutch.

\begin{figure}[h]
\begin{minipage}{0.49\textwidth}
    \centering
    \includegraphics[trim={0 0 5cm 0},clip, width=\textwidth]{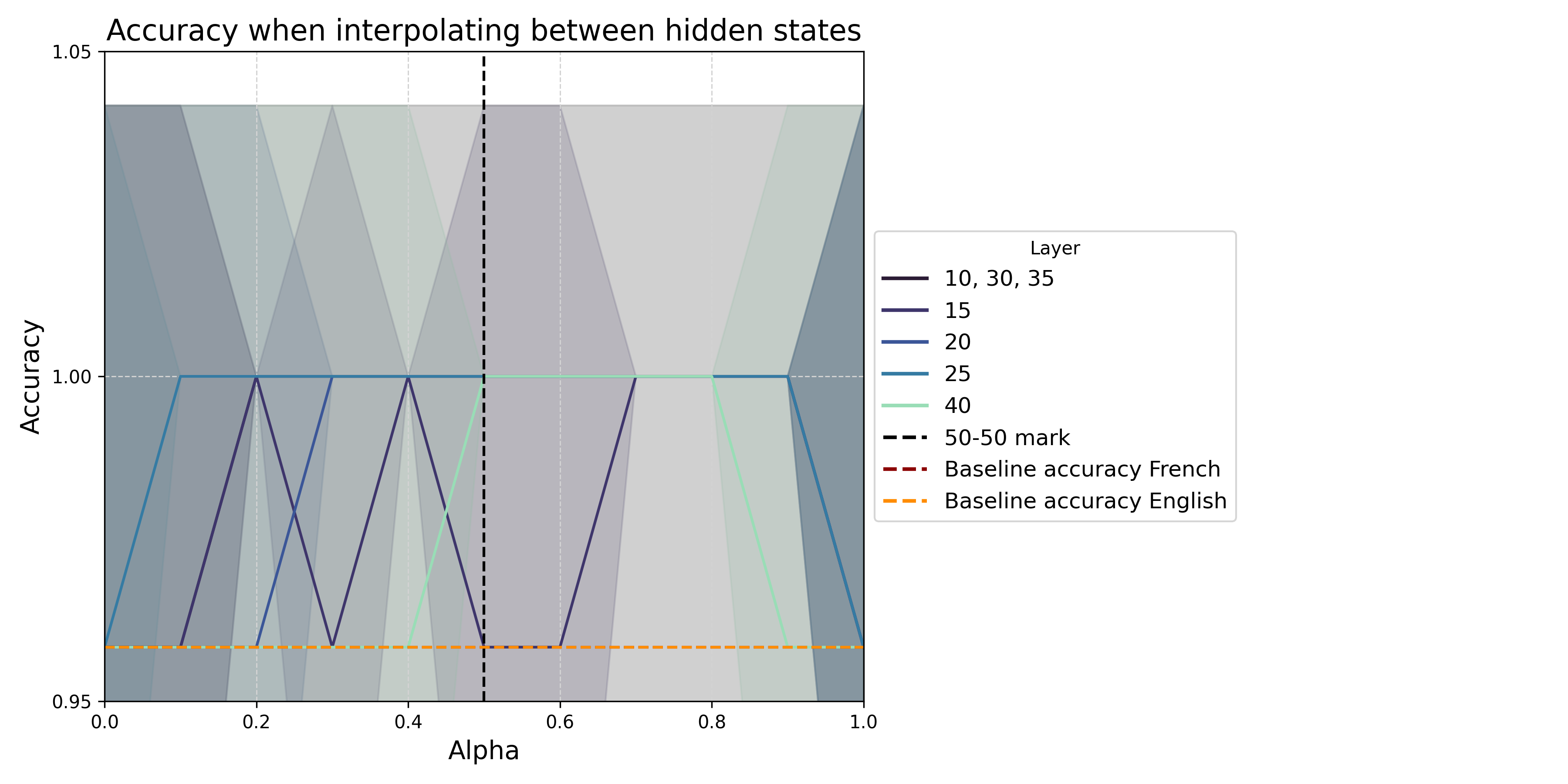} 
\end{minipage}
\begin{minipage}{0.49\textwidth}
    \centering
    \includegraphics[trim={1.5cm 0.5cm 1.5cm 1.5cm},clip,width=\textwidth]{figures/interpolate/FR_ENG_aya_heatmap_interpolate_results.png} 
\end{minipage}
\caption{Hidden state interpolation between French prompts, and English prompts in \gemma. Left shows the accuracy (i.e., the proportion of times the model correctly outputs city in either language). Right shows the propensity of the model to answer in English (red) and French (blue). }
\end{figure}

\begin{figure}[h]
\begin{minipage}{0.49\textwidth}
    \centering
    \includegraphics[trim={0 0 5cm 0},clip, width=\textwidth]{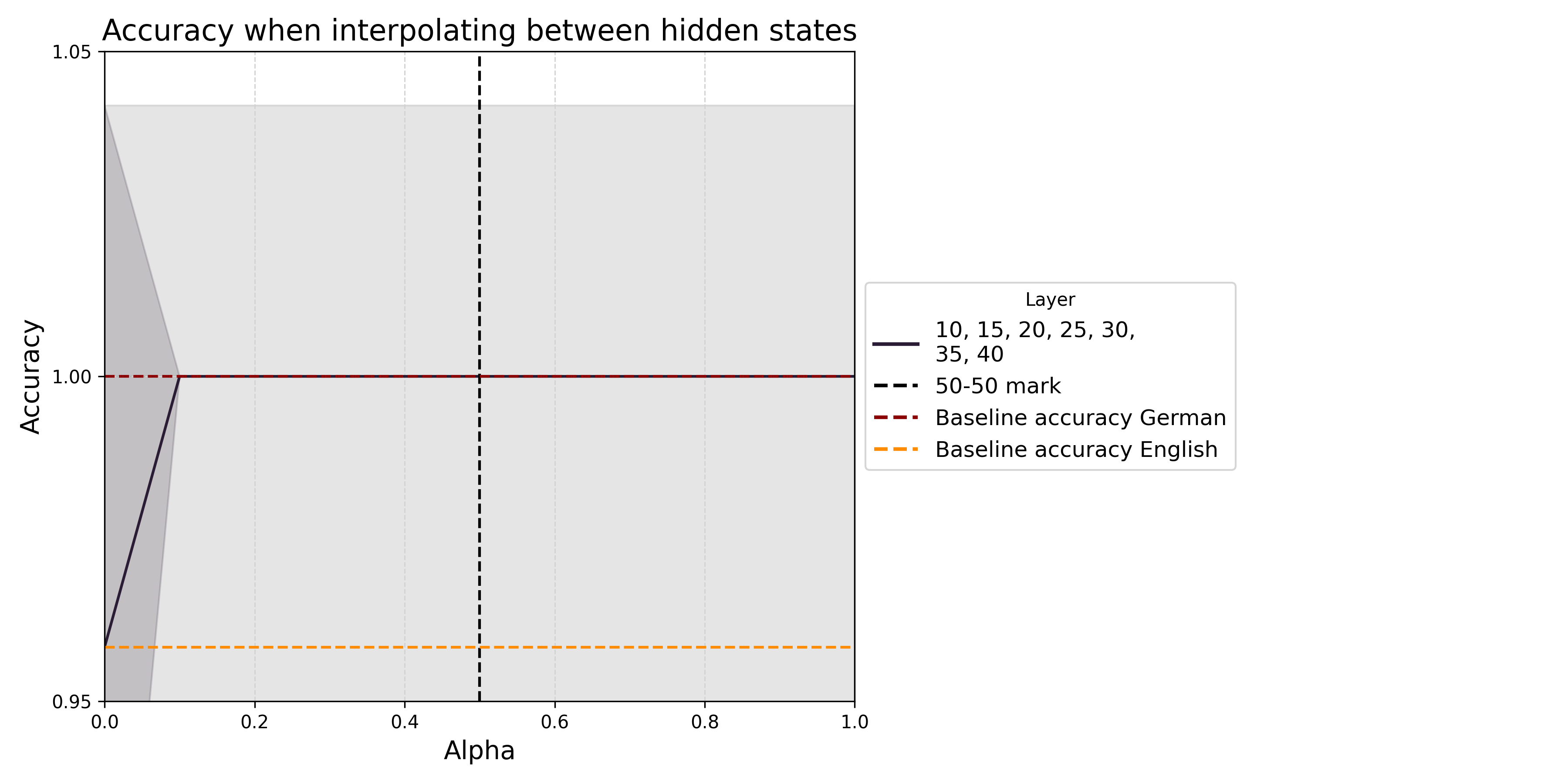} 
\end{minipage}
\begin{minipage}{0.49\textwidth}
    \centering
    \includegraphics[trim={1.5cm 0.5cm 1.5cm 1.5cm},clip,width=\textwidth]{figures/interpolate/GER_ENG_aya_heatmap_interpolate_results.png} 
\end{minipage}
\caption{Hidden state interpolation between German prompts, and English prompts in \gemma. Left shows the accuracy (i.e., the proportion of times the model correctly outputs city in either language). Right shows the propensity of the model to answer in English (red) and German (blue). }
\end{figure}

\begin{figure}[h]
\begin{minipage}{0.49\textwidth}
    \centering
    \includegraphics[trim={0 0 5cm 0},clip, width=\textwidth]{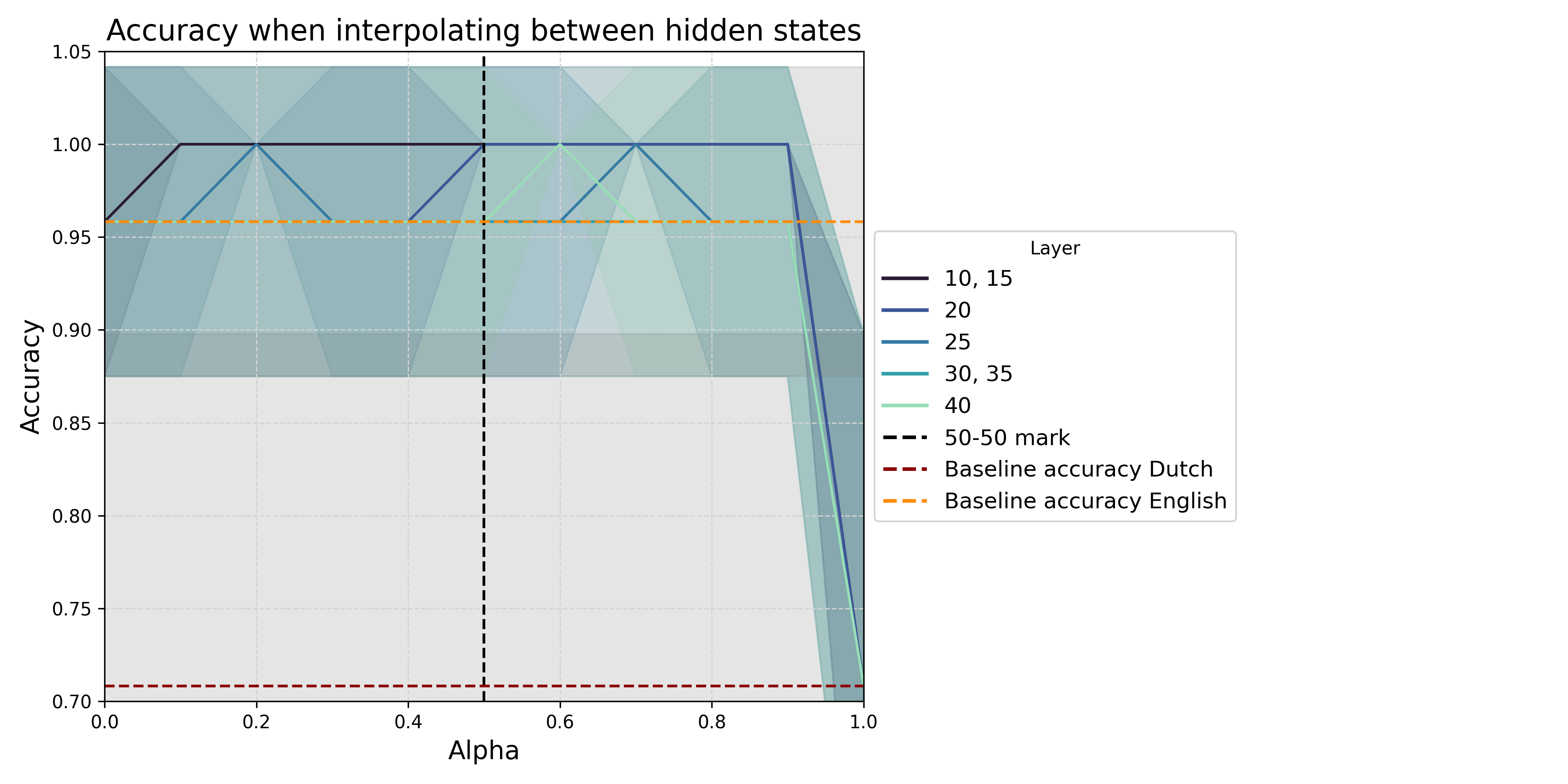} 
\end{minipage}
\begin{minipage}{0.49\textwidth}
    \centering
    \includegraphics[trim={1.5cm 0.5cm 1.5cm 1.5cm},clip,width=\textwidth]{figures/interpolate/NL_ENG_aya_heatmap_interpolate_results.png} 
\end{minipage}
\caption{Hidden state interpolation between Dutch prompts, and English prompts in \gemma. Left shows the accuracy (i.e., the proportion of times the model correctly outputs city in either language). Right shows the propensity of the model to answer in English (red) and Dutch (blue). }
\end{figure}

\begin{figure}[h]
\begin{minipage}{0.49\textwidth}
    \centering
    \includegraphics[trim={0 0 5cm 0},clip, width=\textwidth]{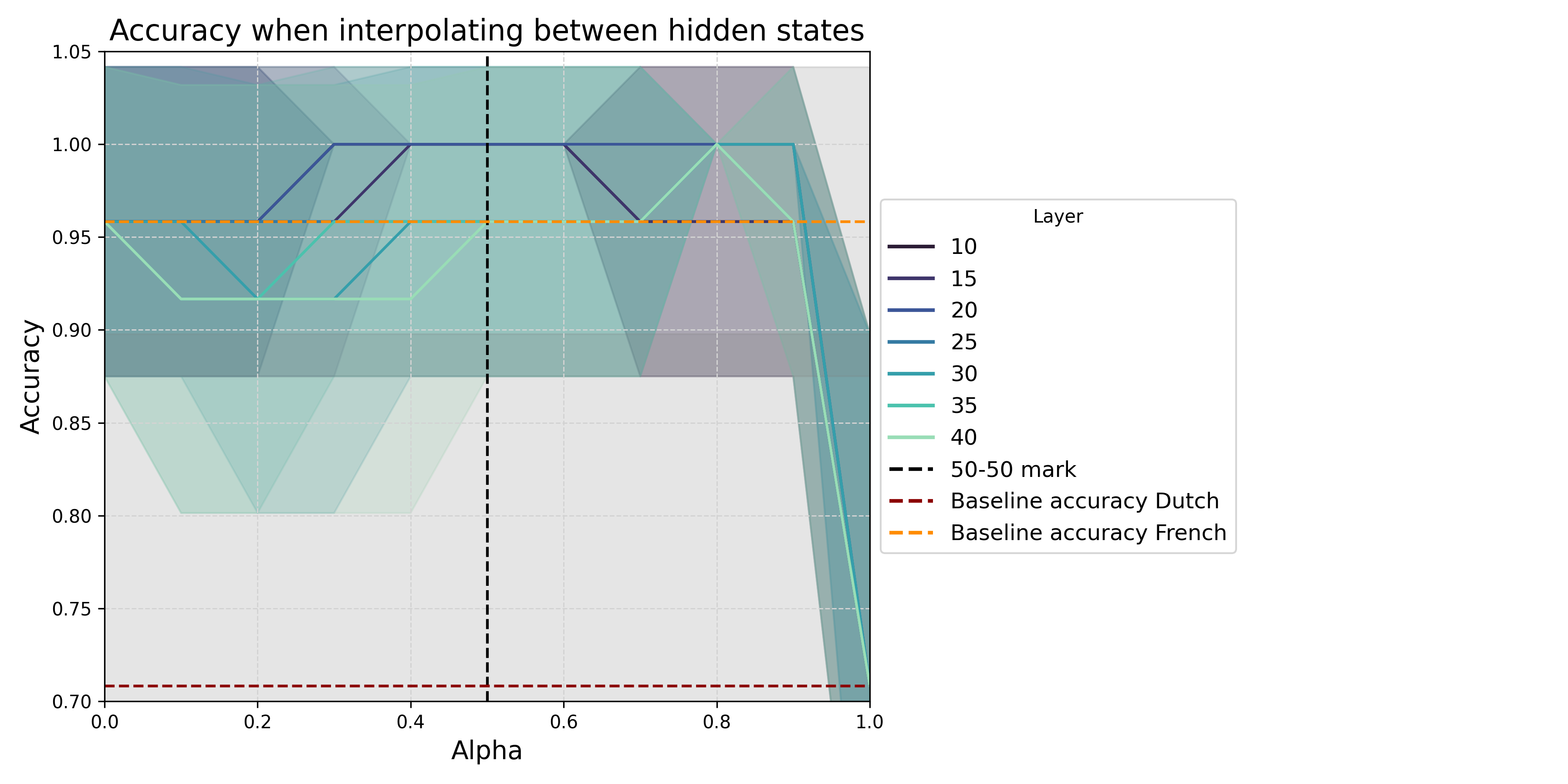} 
\end{minipage}
\begin{minipage}{0.49\textwidth}
    \centering
    \includegraphics[trim={1.5cm 0.5cm 1.5cm 1.5cm},clip,width=\textwidth]{figures/interpolate/NL_FR_aya_heatmap_interpolate_results.png} 
\end{minipage}
\caption{Hidden state interpolation between French prompts, and Dutch prompts in \gemma. Left shows the accuracy (i.e., the proportion of times the model correctly outputs city in either language). Right shows the propensity of the model to answer in French (red) and Dutch (blue). }
\end{figure}

\begin{figure}[h]
\begin{minipage}{0.49\textwidth}
    \centering
    \includegraphics[trim={0 0 5cm 0},clip, width=\textwidth]{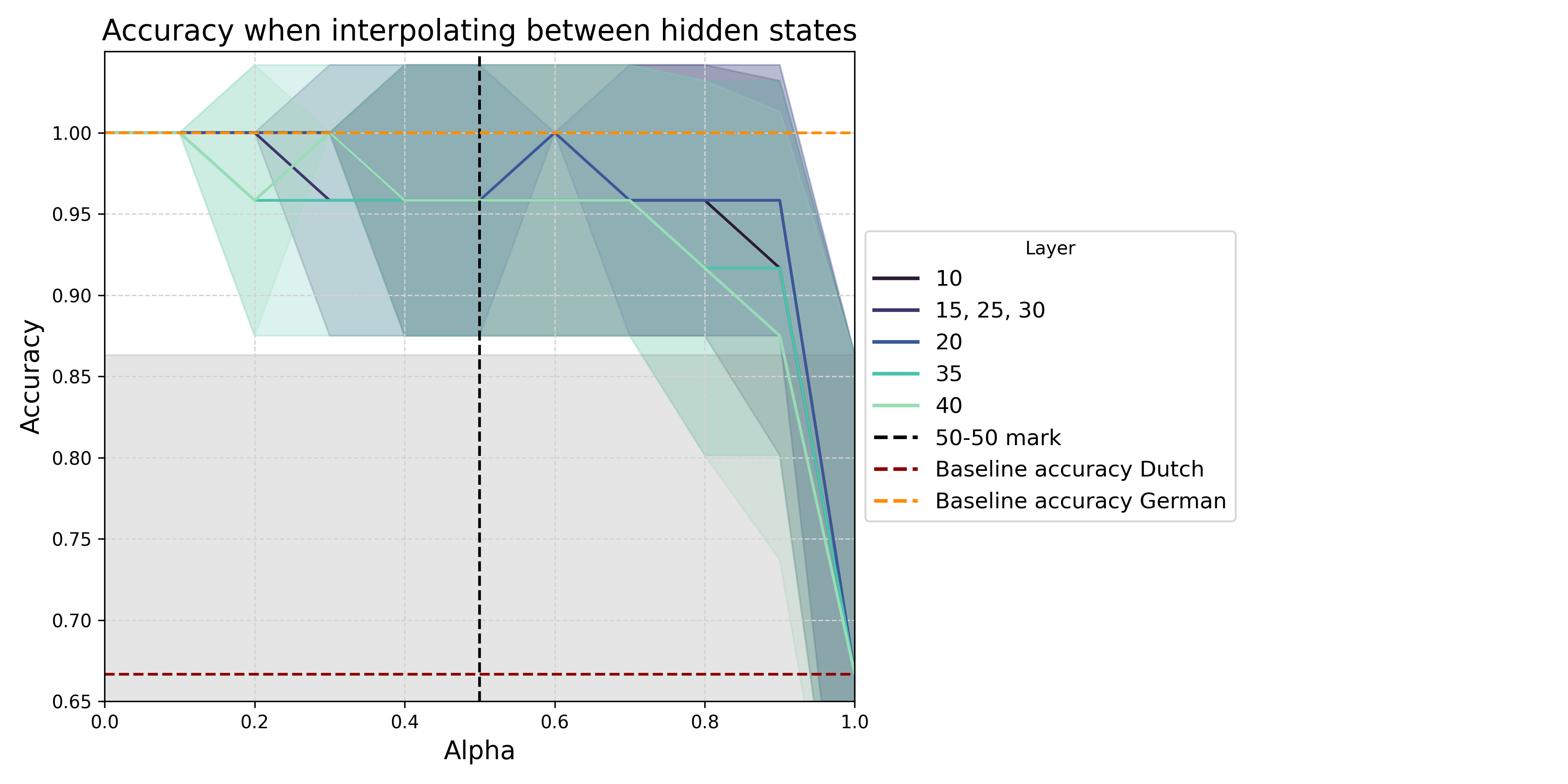} 
\end{minipage}
\begin{minipage}{0.49\textwidth}
    \centering
    \includegraphics[trim={1.5cm 0.5cm 1.5cm 1.5cm},clip,width=\textwidth]{figures/interpolate/NL_GER_aya_heatmap_interpolate_results.png} 
\end{minipage}
\caption{Hidden state interpolation between German prompts, and Dutch prompts in \gemma. Left shows the accuracy (i.e., the proportion of times the model correctly outputs city in either language). Right shows the propensity of the model to answer in German (red) and Dutch (blue). }
\end{figure}

\begin{figure}[h]
\begin{minipage}{0.49\textwidth}
    \centering
    \includegraphics[trim={0 0 5cm 0},clip, width=\textwidth]{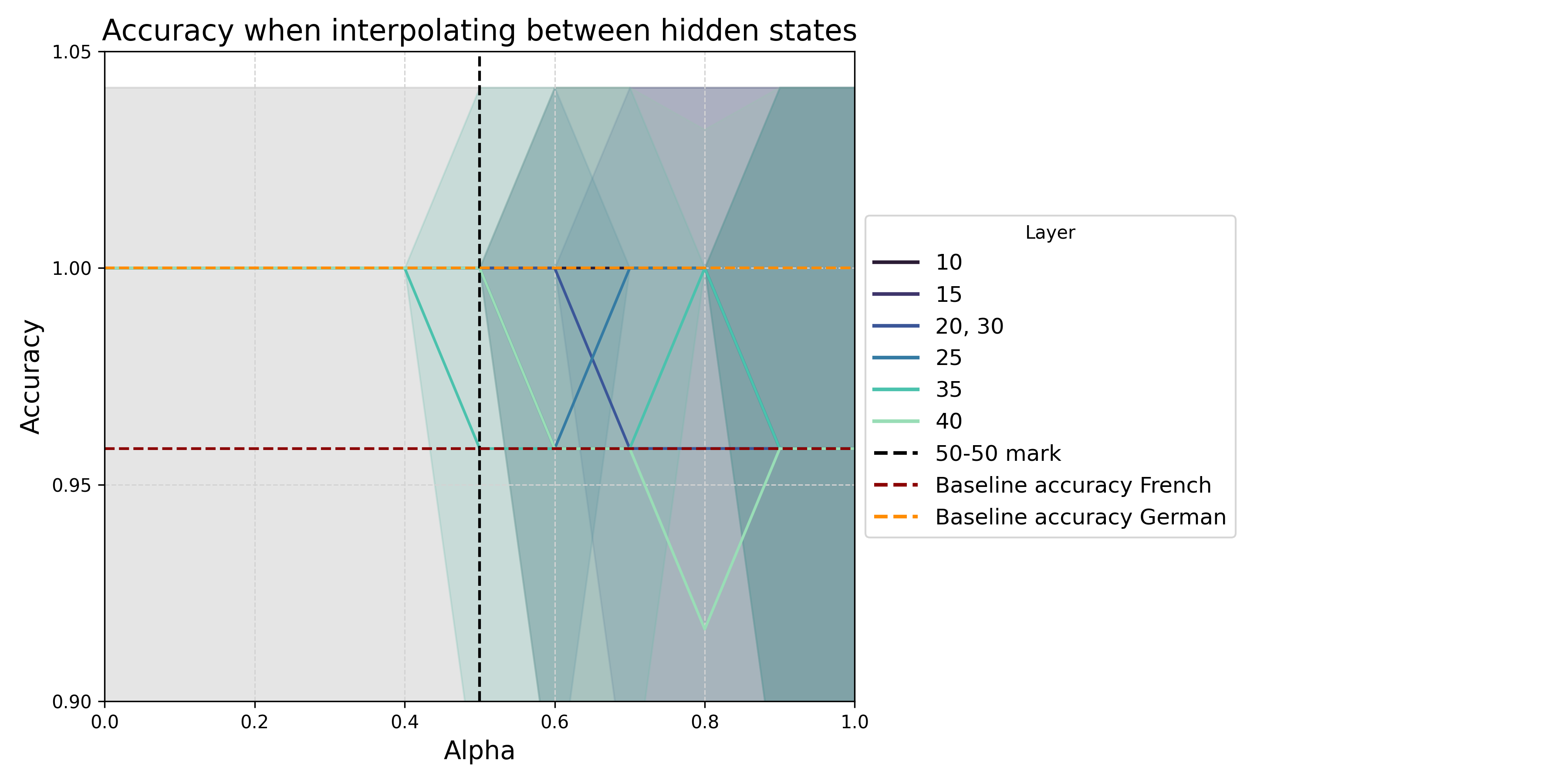} 
\end{minipage}
\begin{minipage}{0.49\textwidth}
    \centering
    \includegraphics[trim={1.5cm 0.5cm 1.5cm 1.5cm},clip,width=\textwidth]{figures/interpolate/FR_GER_aya_heatmap_interpolate_results.png} 
\end{minipage}
\caption{Hidden state interpolation between French prompts, and German prompts in \gemma. Left shows the accuracy (i.e., the proportion of times the model correctly outputs city in either language). Right shows the propensity of the model to answer in German (red) and French (blue). }
\end{figure}

\begin{figure}[h]
\begin{minipage}{0.49\textwidth}
    \centering
    \includegraphics[trim={0 0 5cm 0},clip, width=\textwidth]{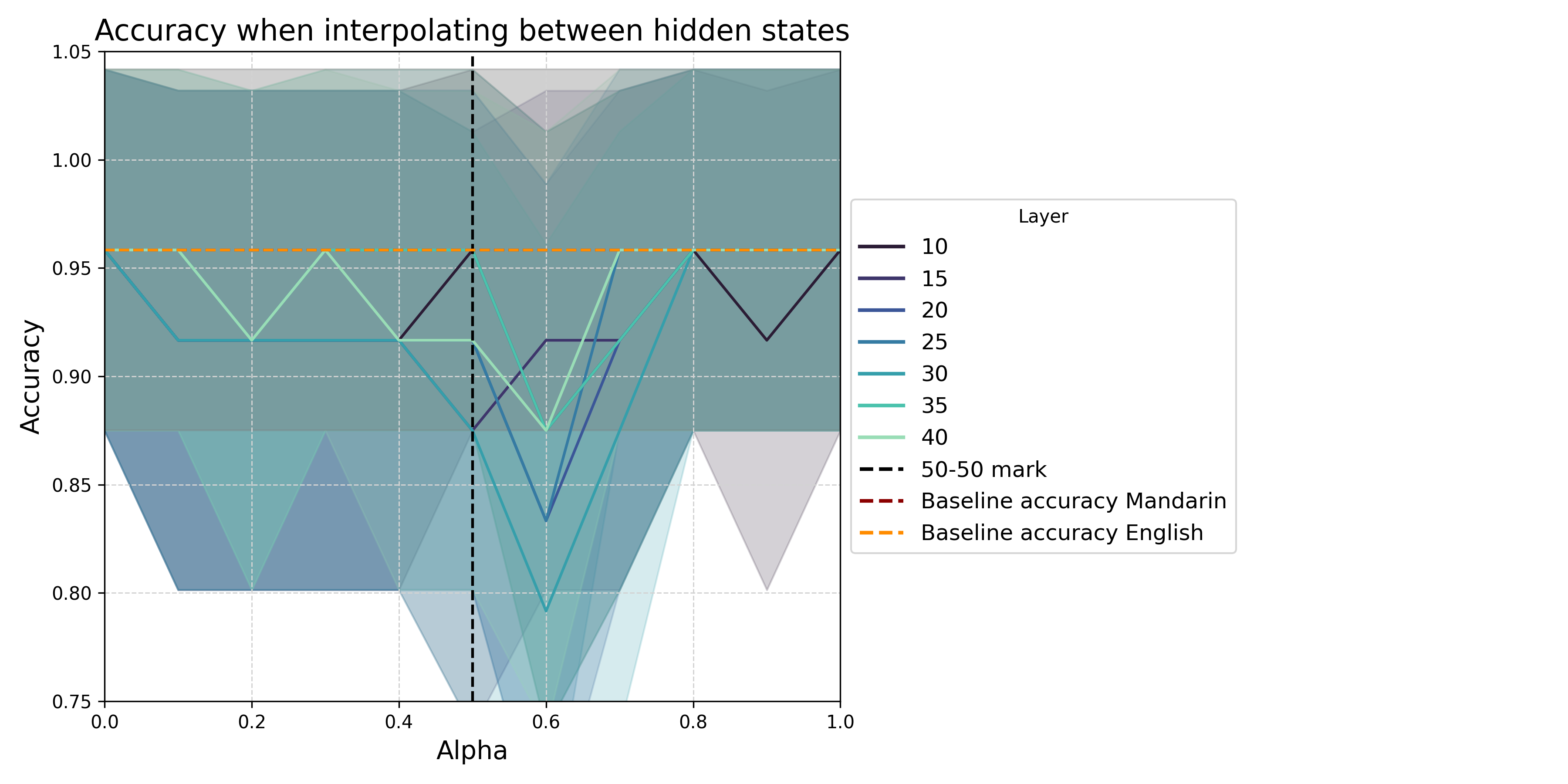} 
\end{minipage}
\begin{minipage}{0.49\textwidth}
    \centering
    \includegraphics[trim={1.5cm 0.5cm 1.5cm 1.5cm},clip,width=\textwidth]{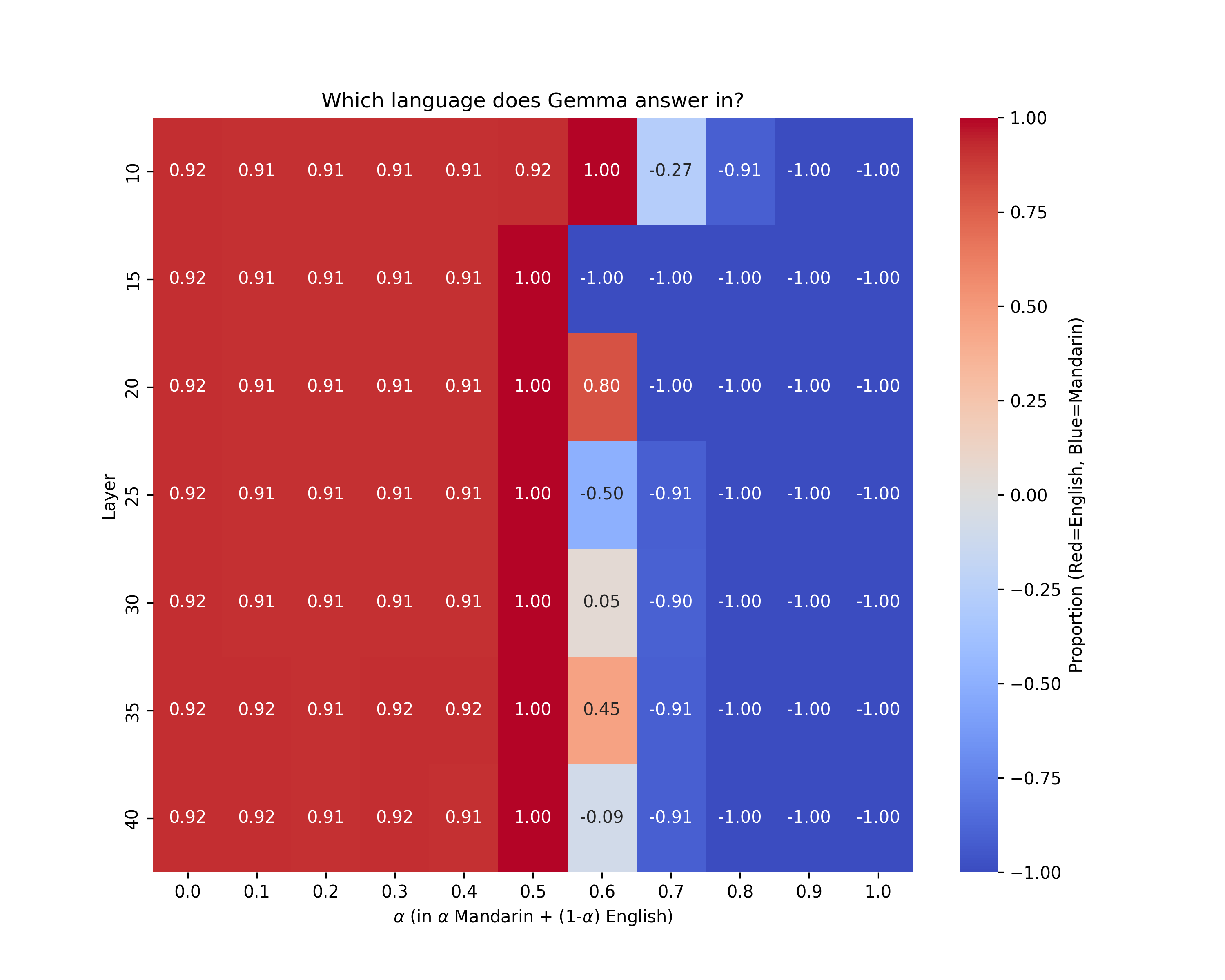} 
\end{minipage}
\caption{Hidden state interpolation between English prompts, and Mandarin prompts in \gemma. Left shows the accuracy (i.e., the proportion of times the model correctly outputs city in either language). Right shows the propensity of the model to answer in English (red) and Mandarin (blue). }
\end{figure}

\begin{figure}[h]
\begin{minipage}{0.49\textwidth}
    \centering
    \includegraphics[trim={0 0 5cm 0},clip, width=\textwidth]{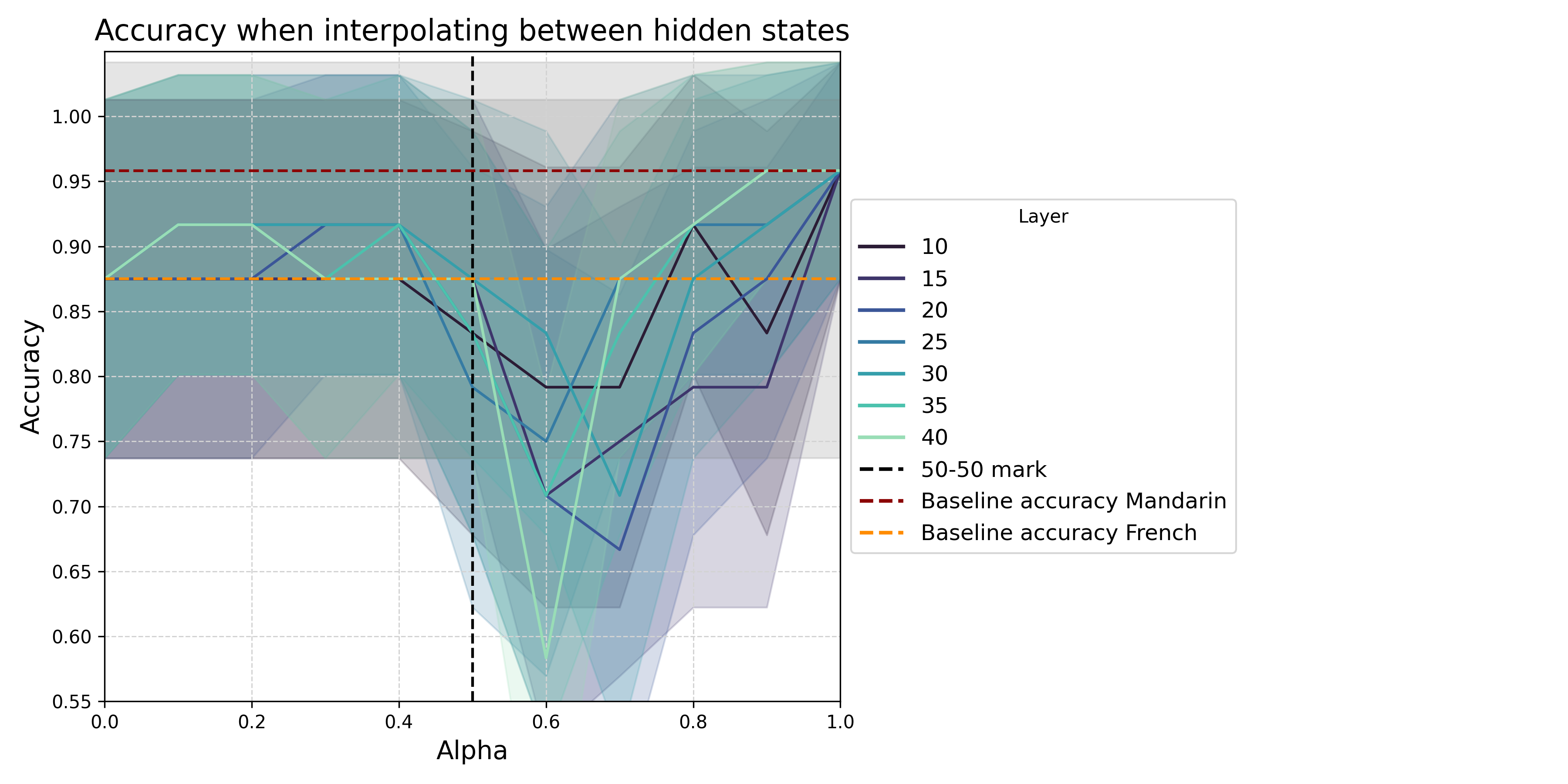} 
\end{minipage}
\begin{minipage}{0.49\textwidth}
    \centering
    \includegraphics[trim={1.5cm 0.5cm 1.5cm 1.5cm},clip,width=\textwidth]{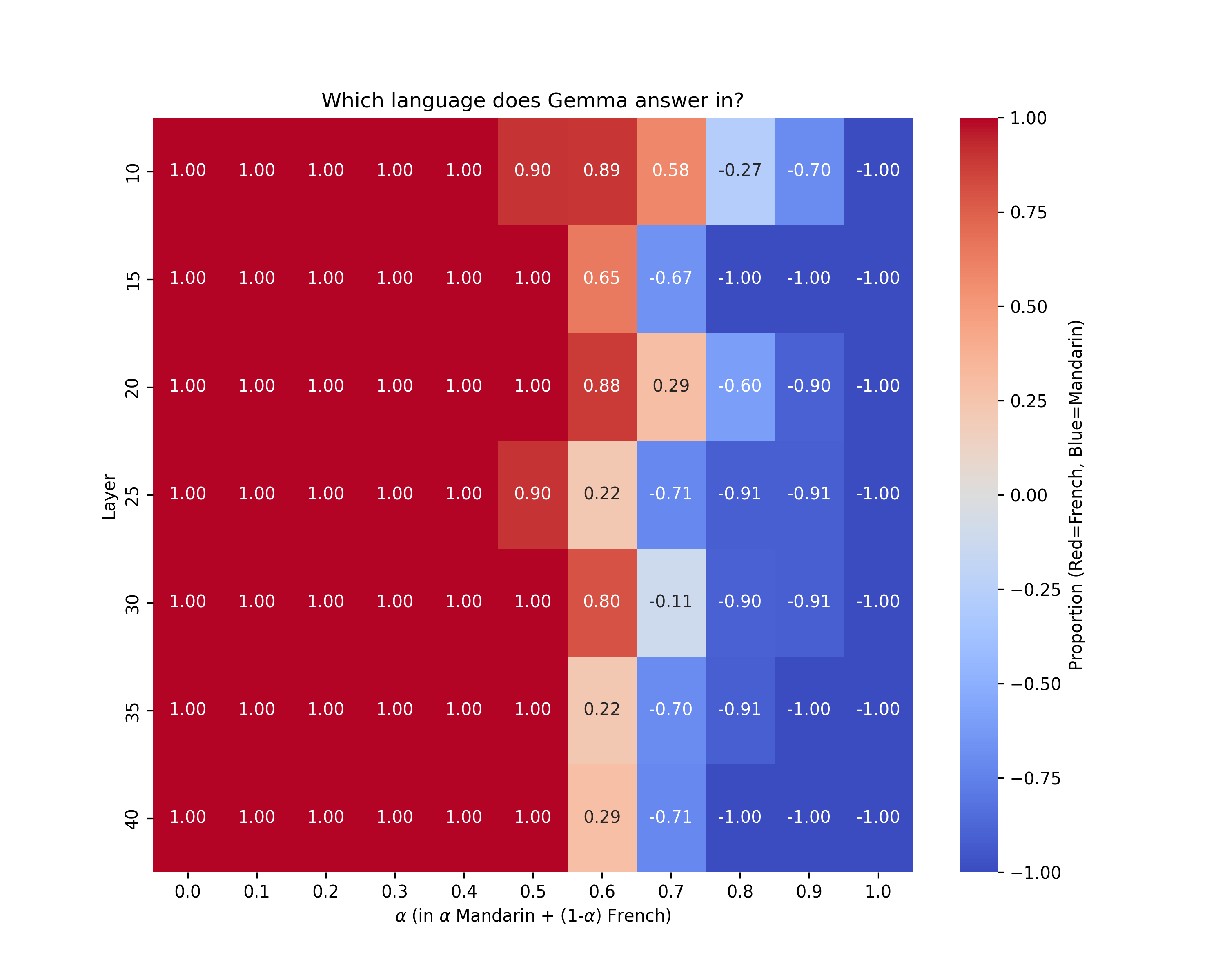} 
\end{minipage}
\caption{Hidden state interpolation between French prompts, and Mandarin prompts in \mistral. Left shows the accuracy (i.e., the proportion of times the model correctly outputs city in either language). Right shows the propensity of the model to answer in French (red) and Mandarin (blue). }
\end{figure}

\begin{figure}[h]
\begin{minipage}{0.49\textwidth}
    \centering
    \includegraphics[trim={0 0 5cm 0},clip, width=\textwidth]{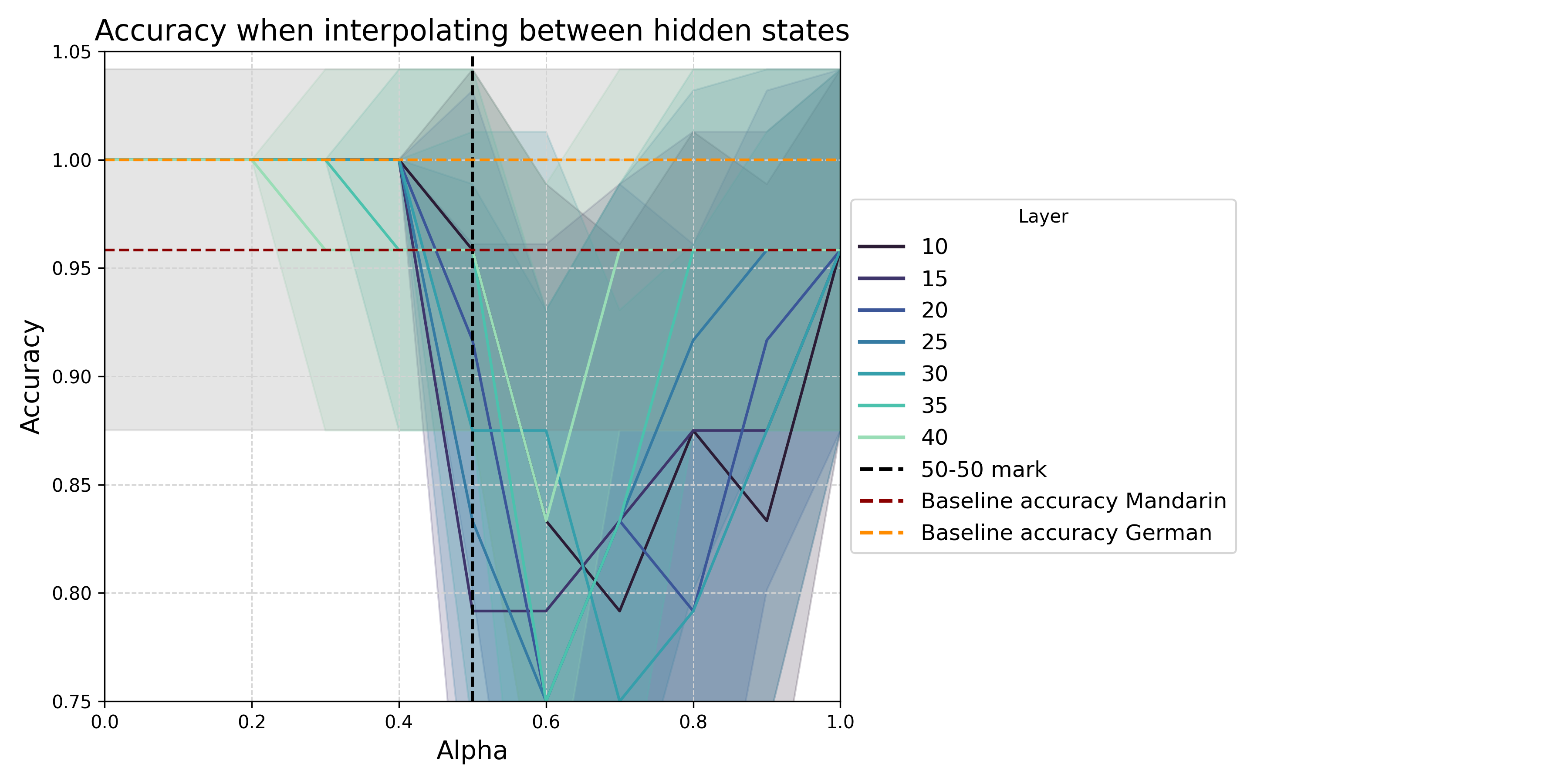} 
\end{minipage}
\begin{minipage}{0.49\textwidth}
    \centering
    \includegraphics[trim={1.5cm 0.5cm 1.5cm 1.5cm},clip,width=\textwidth]{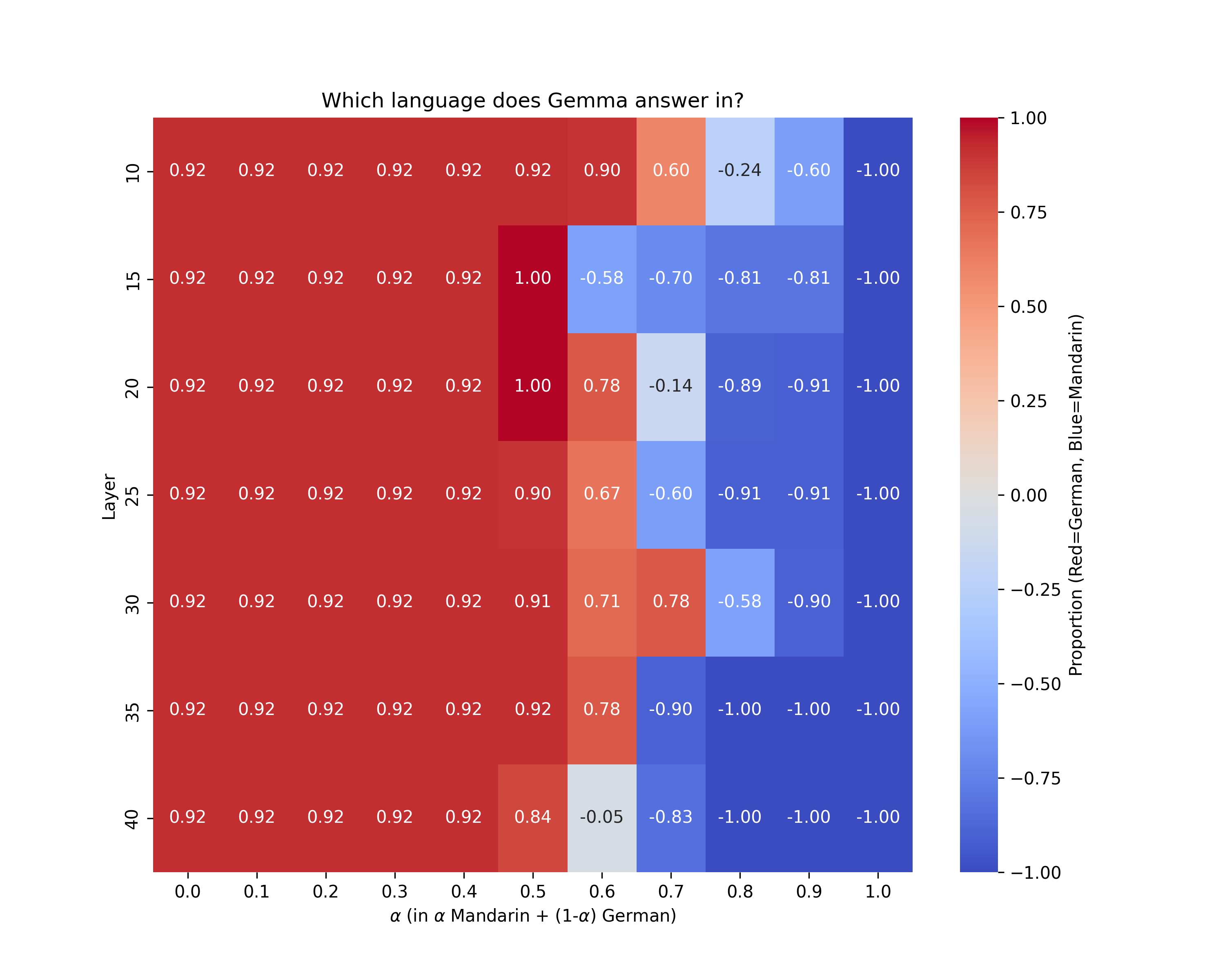} 
\end{minipage}
\caption{Hidden state interpolation between German prompts, and Mandarin prompts in \mistral. Left shows the accuracy (i.e., the proportion of times the model correctly outputs city in either language). Right shows the propensity of the model to answer in German (red) and Mandarin (blue). }
\end{figure}

\begin{figure}[h]
\begin{minipage}{0.49\textwidth}
    \centering
    \includegraphics[trim={0 0 5cm 0},clip, width=\textwidth]{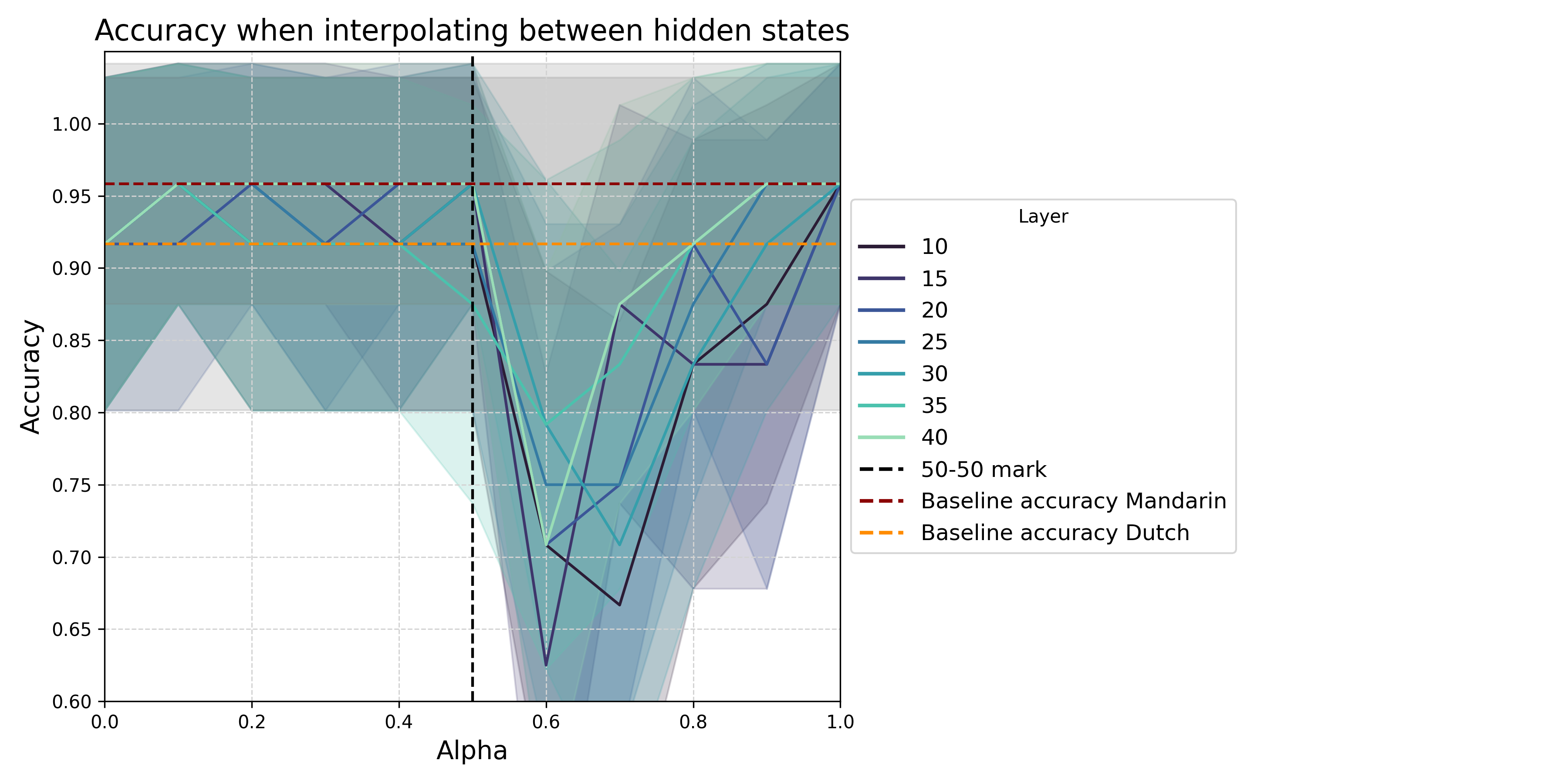} 
\end{minipage}
\begin{minipage}{0.49\textwidth}
    \centering
    \includegraphics[trim={1.5cm 0.5cm 1.5cm 1.5cm},clip,width=\textwidth]{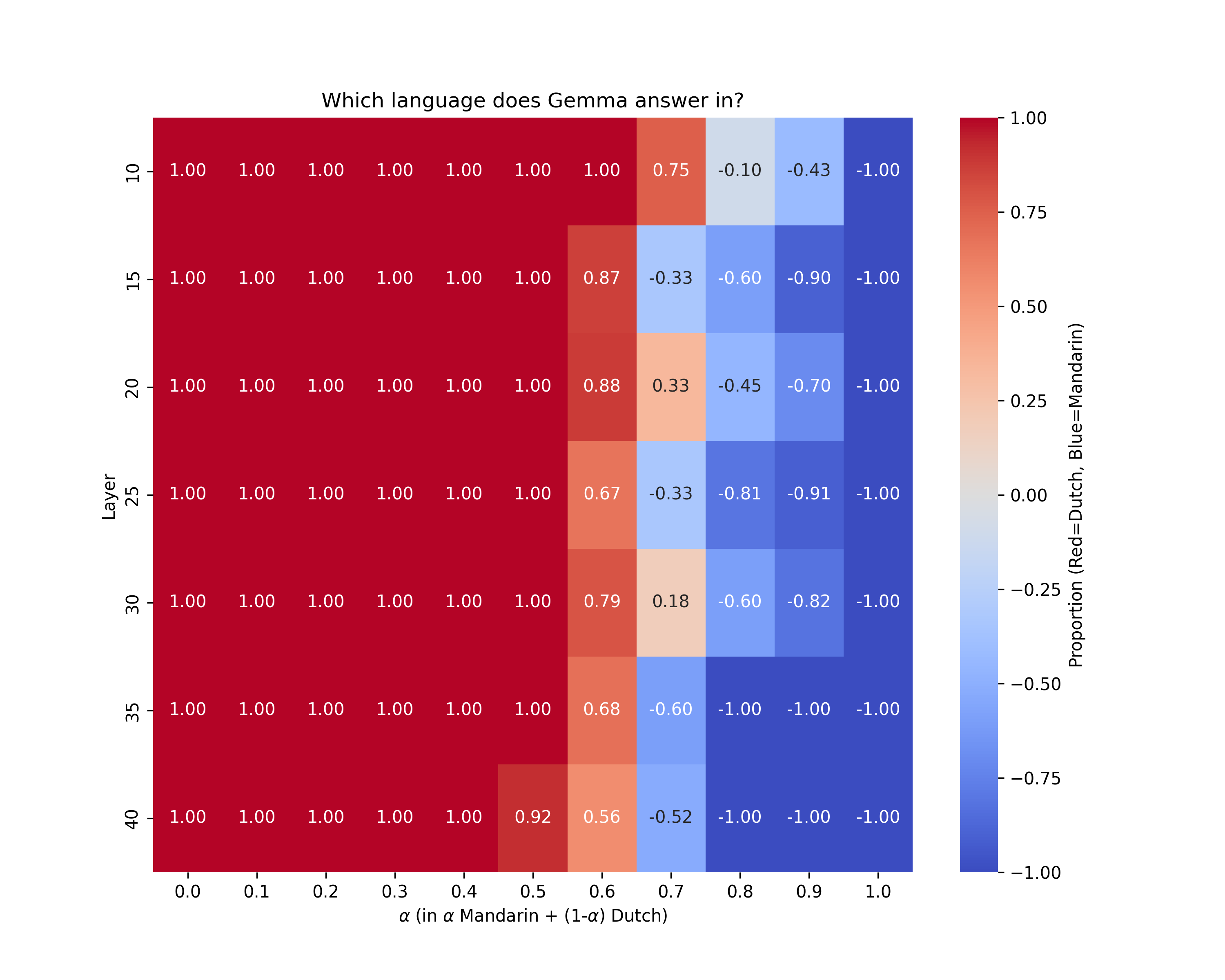} 
\end{minipage}
\caption{Hidden state interpolation between Dutch prompts, and Mandarin prompts in \mistral. Left shows the accuracy (i.e., the proportion of times the model correctly outputs city in either language). Right shows the propensity of the model to answer in Dutch (red) and Mandarin (blue). }
\end{figure}

\end{document}